\documentclass[final,3p]{elsarticle}

\usepackage{framed,multirow}
\usepackage{caption}
\usepackage{longtable}

\usepackage{amssymb}
\usepackage{latexsym}

\usepackage{lscape}
\usepackage{pdflscape}
\usepackage{afterpage}
\usepackage{capt-of}
\usepackage{adjustbox}
\usepackage{url}
\usepackage{xcolor}
\definecolor{newcolor}{rgb}{.8,.349,.1}

\usepackage{rotating}

\journal{Neurocomputing}

\begin{document}

\clearpage

\ifpreprint
  \setcounter{page}{1}
\else
  \setcounter{page}{1}
\fi

\begin{frontmatter}

\title{Object Recognition Datasets and Challenges: A Review}

\author[1]{Aria Salari}
\author[1]{Abtin Djavadifar}
\author[1]{Xiang Rui Liu}
\author[1]{Homayoun Najjaran\corref{cor1}} 
\cortext[cor1]{Corresponding author; 
  Tel.: +1-250-826-1621;}
\ead{homayoun.najjaran@ubc.ca}

\address[1]{School of Engineering, University of British Columbia, 1137 Alumni Ave, Kelowna, V1V 1V7, BC, Canada}

\begin{abstract}
Object recognition is among the fundamental tasks in the computer vision applications, paving the path for all other image understanding operations. In every stage of progress in object recognition research, efforts have been made to collect and annotate new datasets to match the capacity of the state-of-the-art algorithms. In recent years, the importance of the size and quality of datasets has been intensified as the utility of the emerging deep network techniques heavily relies on training data. Furthermore, datasets lay a fair benchmarking means for competitions and have proved instrumental to the advancements of object recognition research by providing quantifiable benchmarks for the developed models. Taking a closer look at the characteristics of commonly-used public datasets seems to be an important first step for data-driven and machine learning researchers.
In this survey, we provide a detailed analysis of datasets in the highly investigated object recognition areas. More than 160 datasets have been scrutinized through statistics and descriptions. Additionally, we present an overview of the prominent object recognition benchmarks and competitions, along with a description of the metrics widely adopted for evaluation purposes in the computer vision community. All introduced datasets and challenges can be found online at github.com/AbtinDjavadifar/ORDC.	

\end{abstract}

\begin{keyword}
Computer vision, Object recognition, Deep learning
\end{keyword}

\end{frontmatter}


\section{1	Introduction}
\label{sec1}

Object recognition is one of the fundamental computer vision tasks that pertains to 
identifying objects of different classes withing digital visual representations such as images or 3D point clouds. As one of the most important tasks in computer vision, it has found practicality in numerous applications, from autonomous driving and face recognition to human pose estimation and remote sensing.
Throughout the lifetime of computer vision research, datasets have played a critical role in the improvement of the object recognition algorithms. Not only do datasets provide a means to compare and contrast existing algorithms, but they also help push the boundaries for meaningful visual understanding as they evolve in the scale and diversity of covered scenarios. With the emergence of deep learning techniques \citep{Krizhevsky2012b}, the need for large and precisely annotated datasets has further been accentuated. In recent years, object recognition tasks have been transformed in terms of accuracy and complexity, moving from classification of iconic view single-object \citep{Nene1996} images to instance-level segmentation of highly cluttered environments \citep{Cordts2016, Lin2014}. A crucial part of this progress is due to the availability of large-scale and finely annotated datasets.

\subsection{Comparison with Relevant Reviews}

Following the breakthroughs in object recognition in recent years, many notable surveys have been published discussing progressions in algorithm development \citep{Liu2020, Zou2019a, ZhaoObject}, and their practicality in various real-world applications \citep{taghanaki2020deep,FengSemantic, LATEEF2019321, MiaoRemote, TASKIRAN2020102809}. While a number of these surveys provide valuable discussions on relevant datasets \citep{garciagarcia2017review, Zou2019a}, there is still a lack of dedicated surveys providing an in-depth analysis on the recent progressions and trends in object recognition from the dataset development standpoint.  As the striking success of deep learning-based object recognition algorithms in recent years is largely dependent on the availability of large and carefully annotated datasets, an independent survey on relevant datasets can provide a more comprehensive understanding of the current challenges in object recognition research as a whole and serve as a roadmap for future research directions. 

\subsection{Scope}

This paper aims to analyze the progression of prominent object recognition datasets in the past two decades, with a focus on the more evolved entries in recent years.  Along with the datasets, related competitions and common evaluation metrics used for the assessment of recognition algorithms have also been discussed. Due to the wide breadth of object recognition datasets and limited space, we restrict our analysis to datasets containing still 2D images in the visible spectrum. Further, we narrow our coverage to annotated datasets since the challenging tasks in the field are mostly tackled with supervised learning approaches. We understand the importance of other data modalities, and also computer vision tasks that use object recognition as a component, but the discussion of any material other than the discussed limits of this paper would be beyond the scope of any reasonable length publication. Finally, where possible, we tried to gauge importance and popularity by referring to highly cited works presented in the the most accredited publication venues in the field such as CVPR, ICCV, etc. We acknowledge there are certainly valuable research efforts beyond  the publications discussed in this paper that we had to ignore due to limited space. 

\subsection{Contributions of the Paper}

A comprehensive survey evaluating the utility of object recognition datasets in a quantitative manner will be an important contribution to the field. By providing a technical overview of the historical  developments and contrasting the state-of-the-art in critical frames such as data bias, collection sources, annotation quality, readers will be able to form a hierarchical understanding of the role of data in object recognition. Further, by providing concise tabular descriptions of the most prominent datasets in popular applications, this paper can serve as a starting point for practitioners and researchers in the field looking to narrow their search for appropriate datasets. Finally, as object recognition algorithms mature, the existing datasets become saturated, thus making it necessary to develop more challenging datasets. To this end, we analyze the current trends to provide researchers with insight regarding future dataset collection directions.

\section{Background}
 
In this section, a history of the prominent datasets evolving through time, and an introduction to milestones in object recognition techniques before and after the emergence of deep learning models are provided. The most eminent evaluation metrics are also discussed.

\subsection{An Overview of Object Recognition Tasks}

Before introducing various object recognition algorithms, it is noteworthy to discern different but similar computer vision tasks like image classification, object detection, object localization, instance segmentation, and semantic segmentation, to avoid any confusion that may happen later. An image classification method assigns a class label to a whole image, whilst object localization draws a bounding box around each object present in the image. Object detection is a combination of these two tasks and assigns a class label to each object of interest after drawing a bounding box around Object recognition is a general term used for referring to all of these tasks together \citep{Russakovsky2015}. Figure 1 shows how these concepts are related to each other.

\begin{figure}[!h]
\centering
\includegraphics[scale=.5]{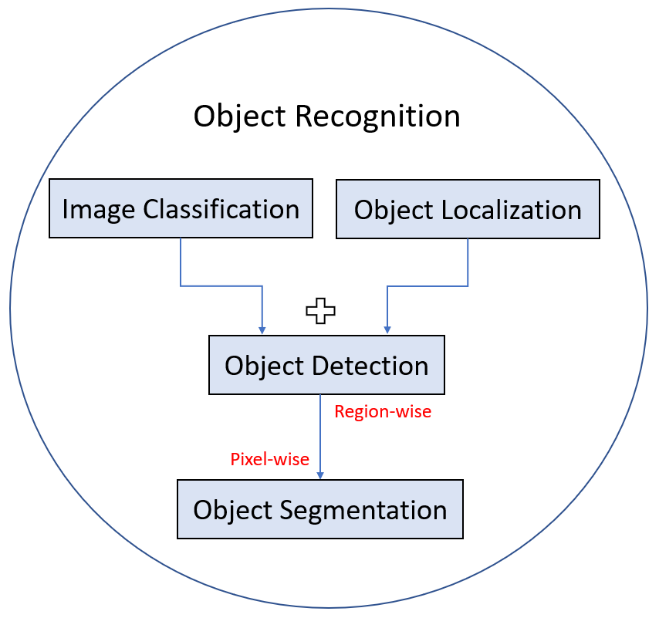}
\caption{Overview of object recognition tasks in computer vision research}
\end{figure}

By applying object detection models, we will only be able to build a bounding box for every object present in the image. However, it will not tell anything about the object’s shape as the bounding boxes are either rectangular or square in shape. Image segmentation models, on the other hand, will create a mask in pixel level for every object that appeared in the image. This method provides us with a more comprehensive understanding of the object(s) present in the image. The image segmentation task itself can be performed in two ways: 1) semantic segmentation which labels each pixel of the image with class of the object surrounded inside the pixel (class-aware labeling), 2) instance segmentation which identifies object boundaries in pixel level and distinguishes between separate objects from the same class (instance-aware labeling).

\subsection{Historical Object Recognition Milestones}

In the early developments of data-driven learning algorithms in the 90's, object recognition algorithms mainly relied on sophisticated hand-crafted features designed for specific applications. As a reult, researchers tried to create ad-hoc datasets for their specific research problem, and collective efforts for data collection and annotation were not as common as today. Early datasets usually focused on one particular application, face detection \citep{Osuna1997, Sung1996,Jesorsky2001}, face recognition \citep{Phillips1998,Beumier2000}, pedestrian detection \citep{Papageorgiou2000}, and handwriting detection \citep{Botta1993,Lecun1997} being among popular ones. Images typically were of low resolution and lacked high level annotations such as segmentation masks, and the objects in question were spatially well-defined. Cropping was also a widely adopted technique to present canonical views of the target object. Here, a canonical or iconic view refers to a clear and distinctive depiction of an object category in an image.
Classification is the most frequent type of annotation in earlier datasets.

MNIST \citep{Lecun1998}, sometimes referred to as the “Hello World” dataset of machine vision, is an example of the early image classification datasets, where each sample contains one hand-written digit in an uncluttered background. COIL-20 \citep{Nene1996} and COIL-100 \citep{Nene1996a} are two other datasets of common objects with different orientations in a black background. FERET \citep{Phillips1998} is the cornerstone of face recognition datasets by offering a large collection of face galleries and a testing framework tuned to the accuracy levels of the then state-of the-art algorithms \citep{Phillips1998,JonathonPhillips2000}. With the introduction of  milestone feature descriptors such as SIFT \citep{Lowe1999}, and Viola Jones \citep{Viola2001, Viola2001a} , and later on HOG \citep{Dalal2005}, SURF \citep{bay2006}, and DPM \citep{Pedro2007}, datasets grew in size and object class coverage. Caltech-101 \citep{Fei-Fei2004} and its successor Caltech-256 \citep{Griffin2007} made a striking turn by including objects in their natural environments, but these datasets were still limited to one class per image. CIFAR-10 and CIFAR-100 \citep{Krizhevsky2012} are 10 and 100-class subsets of the Tiny Images dataset \citep{Torralba2008} in 32×32 pixel resolution. Numerous annotation errors occur in these datasets because the automatically generated annotations were not manually verified.
In the early 2000’s, several attempts were also made to create datasets with higher quality of annotations. The first prominent dataset with segmentation mask annotation was BSDS \citep{Martin2001}, where human annotators manually segmented each image. An error measure was also introduced to evaluate the performance of segmentation algorithms against ground truth human annotations. However, the size of the dataset is by orders of magnitude smaller than the modern-day segmentation datasets, such as Microsoft COCO \citep{Lin2014}, and the evaluation criteria predicts zero error in cases where there is a proper subset relationship between the predicted and ground truth segments. LabelMe \citep{Russell2008} leveraged crowdsourcing to acquire polygon annotation for objects in natural contexts but lacked uniform annotations because of arbitrary captioning and inaccurate polygons of annotators \citep{Russakovsky2015}. However, the online LabelMe annotation tool was extensively used to annotate other datasets, e.g., MIT-CSAIL \citep{Torralba2004}, SUN \citep{Xiao2010}, and ADE20K \citep{Zhou2017}. PASCAL VOC \citep{Everingham2006} was a stepping stone among the datasets of this era as it was adopted to hold annual object recognition competitions to benchmark the state-of-the-art algorithms .
Up until 2012, algorithms utilizing local descriptors and hand tuned feature vectors dominated these  competitions \citep{Everingham2014}.

The current enthusiasm for deep learning-based methods goes back to 2012, wherein the winning entry of the ImageNet Large Scale Visual Recognition Challenge (ILSVRC) was a Deep Convolutional Neural Network (DCNN), i.e., AlexNet \citep{Krizhevsky2012b}.The use of CNNs for image classification dates back to the 1980s. The early works focused on the identification of hand-written digits, specifically as related to automated zip code detection \citep{ Lecun1990, Lecun1989}.  Research on the method continued through the late 1990s, but with little adoption \citep{Lecun1997}. Lack of parallel compute power available at the time was the primary reason why CNNs were not used in large scale \citep{Rawat2017}. Advancement in GPU  computation and increased availability of digitalized datasets led to a renewal of interest in 2006 and brought about technological advances such as the first application of maximum pooling for dimensionality reduction \citep{Chellapilla2006, Hinton2006,Hinton2006a, Bengio2007, Ranzato2007}. ILSVRC 2012 showed the full potential of deep algorithms can only be released when large-scale and accurately annotated datasets are available. Not only did DCNNs achieved staggering improvements in generic object recognition competitions, but they also opened doors for new applications where object recognition was never possible before.  This lead to a flurry of organized efforts to collect generic and application-specific datasets. With the resulting jump in recognition accuracy levels, datasets leaned more towards finer annotations, moving from classification of object classes to pixel-level annotations \citep{Lin2014, Cordts2016}. Also, datasets were now being gathered in less controlled settings. Recently, there has been a focus to  fully annotate images with a dense number object instances in them to extract richer contextual information \citep{Gupta2019}. 

The evolution of milestone DCNNs emerging after AlexNet is discussed in the following. It needs to be mentioned that since a thorough analysis of these algorithms is beyond the scope of this paper, we only outline the development trajectory of these algorithm. For more information readers are encouraged to refer to relevant surveys \citep{Zou2019, Liu20200}.    
\textbf{R-CNN} \citep{Girshick2014} is a two-step Region-based Convolutional Neural Network. The R-CNN structure is based on two steps. First, selecting a feasible number of bounding-box object regions as candidates (also known as RoI) using a selective search approach. Second, extracting CNN features to perform classification while taking the features from each region independently. 
To overcome the expensive and slow training process of the R-CNN model, Girshick et al. enhanced the training process by joining three independent models together and training the new framework, called \textbf{Fast R-CNN} \citep{Girshick2015}. 
Although Fast R-CNN was noticeably faster during training and testing, the achieved improvement was not significant because of the high expenses of separately generating the region proposals by another model. \textbf{Faster R-CNN} \citep{Ren2015} tried to speed up this process by integrating the region proposal algorithm into the CNN model by constructing a single, joined model made of fast R-CNN and RPN (Regional Proposal Network) having the same convolutional feature layers.
\textbf{Single Shot Detector} (SSD) was introduced by C. Szegedy et al. in \citep{Liu2015c} in 2016 and could reach a new record in object detection task performance. Despite the RPN-based approaches like R-CNN series that generate region proposals and detect each proposal's object in two separate stages, SSD detects multiple objects present in the image in only one shot which lets it perform faster.
Single Shot means that object classification and localization tasks are both done in a single forward pass of the CNN. SSD takes the bounding boxes' output space and discretizes them into a group of default boxes over various aspect ratios and then scales them for every feature map location. SSD proved its competitive accuracy against other techniques with an object proposal step by achieving better results on the MS COCO, PASCAL VOC, and ILSVRC datasets.
Contrary to R-CNN family that locate the objects present in an image using regions by only looking at the areas of the images where are more probable to contain an object, \textbf{YOLO} \citep{Redmon} framework considers the whole image as its input and makes predictions about the coordinates of the bounding boxes and their probabilities. Three main advantages of YOLO are 1) its incredibly fast speed by processing images at 45 fps  in real-time, 2) its ability to perform global reasoning and seeing the entire image while making predictions despite the region proposal-based and sliding window techniques 3) its capability to understand generalized object representations. 
After the introduction of YOLO \citep{Redmon} in 2016, various versions of YOLO have been developed trying to enhance its performance in different aspects. Fast YOLO is a version of YOLO using 9 convolutional layers instead of 24 which performs about 3 times faster than YOLO but has lower mAP (mean Average Precision) scores \citep{Shafiee2017}. YOLOv2 is focused on reducing the significant number of localization errors and improving the low recall of original YOLO while maintaining classification accuracy \citep{Redmon2016}. YOLOv3 is the latest member of the YOLO family with some improvements compared to YOLOv2 such as a better feature extractor, a backbone with shortcut connections, and a more accurate object detector with feature map upsampling and concatenation \citep{Redmon2018}.
Pixel-level segmentation needs more accurate alignment compared to bounding boxes. \textbf{Mask R-CNN} \citep{He2017} was an effort to improve the RoI pooling layer to provide more precise mapping of the RoI to the regions of the original image. Mask-RCNN beat all the records in different parts of the COCO suite of challenges \citep{Lin2014}.
In 2019, a new framework called TensorMask \citep{Chen2019} was introduced for extremely accurate instance segmentation tasks \citep{Chen2019}. TensorMask uses a dense, sliding-window technique and novel architectures and operators to capture the 4D geometric structure with rich and effective representations for dense images. 
\textbf{DeepLab} \citep{Chen2017deep} is a revolutionary semantic segmentation model that uses atrous convolutions to simply upsample the output of the last convolutional layer and compute a pixel-wise loss to make dense predictions. 
DeepLab V1 added a number of advancements to the previous models by the employment of Fully Convolutional Networks (FCN) that led to overcoming two main challenges: 1) reduced feature resolution because of the multiple pooling and downsampling layers in DCNNs, 2) reduced localization accuracy because of DCNNs invariance. 
DeepLab V2 attempted to further enhance the performance of the DeepLab V1 by addressing the challenge of  the existence of objects at multiple scales. It used a novel technique called Atrous Spatial Pyramid Pooling (ASPP), wherein multiple atrous convolutions with distinct sampling rates were applied to the input feature map and then outputs were combined \citep{Chen2017}. DeepLab V3 was the last iteration of the DeepLab family that employed a novel encoder-decoder with atrous separable convolution which could obtain boundaries of sharper objects.\citep{Chen2017a}.
\textbf{U-Net} \citep{Ronneberger2015} is an encoder-decoder based architecture first developed for biomedical image segmentation. One of the important challenges in the medical computer vision field is the limited number of datasets as many are not publicly available to protect patient confidentiality. Moreover, accurately annotating medical images requires trained personnel. U-Net employs data augmentation techniques to effectively use the available labeled samples which makes it a useful tool in cases without large annotated datasets. U-Net has also been shown to perform well on grayscale datasets. 

\begin{figure}[!t]
\centering
\includegraphics[scale=.43]{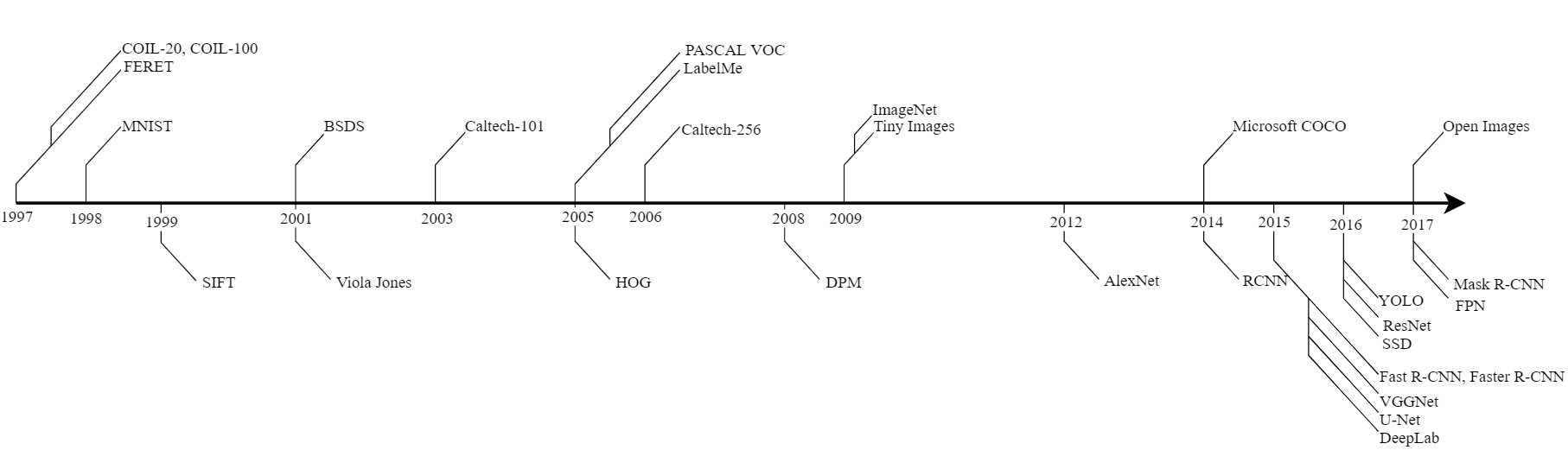}
\caption{Milestones in object recognition algorithm and dataset development.(MNIST \citep{Lecun1998,}, FERET \citep{Phillips1998}, COIL-20 \citep{Nene1996}, \citep{Nene1996a}, BSDS \citep{Martin2001}, Caltech-101 \citep{Fei-Fei2004}, Caltech-256 \citep{Griffin2007}, LabelMe \citep{Russell2008}, Pascal VOC \citep{Everingham2006}, Tiny Images \citep{Torralba2008}, ImageNet \citep{Deng2010}, Microsoft COCO \citep{Lin2014}, Open Images \citep{Kuznetsova2018}, SIFT \citep{Lowe1999}, Viola Jones\citep{Viola2001, Viola2001a}, HOG \citep{Dalal2005}, AlexNet \citep{Krizhevsky2012b}, R-CNN \citep{Girshick2014}, Fast R-CNN \citep{Girshick2015}, Faster-RCNN \citep{Ren2015}, VGGNet \citep{Chatfield2014}, U-Net \citep{Ronneberger2015}, DeepLab \citep{Chen2017deep}, Mask-RCNN \citep{He2017}, FPN \citep{LinFeature}.}
\end{figure}

\subsection{Metrics}
In order to dive into the common metrics used for the evaluation of computer vision algorithms, a few basic definitions need to be first established:

\begin{enumerate}
\item True Positive (TP): A correct prediction of a present condition
\item False Positive (FP): An incorrect prediction of a condition when it is not present.
\item True Negative (TN): A correct prediction of absence of a condition.
\item False Negative (FN): An incorrect prediction of absence of a condition when it is present. 
\end{enumerate}

Having defined the above terms, precision and recall could be defined as follows: 
Precision: Precision is a measure of the accuracy of the positive case predictions. A high precision represents a low level of wrong predictions (FP). It is defined as:

\begin{equation}
Precision=\frac{TP}
 {TP+FP}
\end{equation}

Recall (Sensitivity, True Positive Rate): Recall quantifies the number of positive cases that have been correctly predicted by the algorithm. A high recall value shows a low level of undetected positive cases.

\begin{equation}
Recall=\frac{TP}
 {TP+FN}
\end{equation}

Specificity (True Negative Rate): Specificity measures the share of actual negatives that have been correctly predicted as such.

\begin{equation}
Specificity=\frac{TN}
 {TN+FP}
\end{equation}

Accuracy: Accuracy shows the ratio of correctly predicted instances to all the available instances.

\begin{equation}
Accuracy=\frac{TP+TN}
 {TP+TN+FP+FN}
\end{equation}

It needs to be noted that a similarity threshold between the annotated data and the predictions could be modified in algorithms to determine whether the prediction of a positive case is as accurate enough to be considered as a TP. Tuning the similarity threshold will in turn alter the precision and recall values. For instance, an increase in precision stemmed from a strict threshold comes at the expense of a reduction in the number of detected positive cases and hence recall. 
Receiver Operative Characteristic (ROC) Curve: Mostly used in the medical diagnostic and face verification problems, the ROC curve is obtained by plotting True Positive Rate (Recall) as a function of False Positive Rate (1-Specificity). Each point on the ROC curve corresponds to a set of sensitivity and specificity associated with a decision threshold. For a binary classification problem with equal number of data entries for each class, the curve will be a line of slope 1, given the predictions are entirely random. As the model starts to make better predictions, the curve is bent towards the upper left corner (Fig. 2). The area under the ROC curve is used as a performance metric. A higher area under the ROC curve indicates a higher detection rate of true positives (high sensitivity) while making less false positive predictions (high specificity), therefore specifying a better performance. 

\begin{figure}[!t]
\centering
\includegraphics[scale=.35]{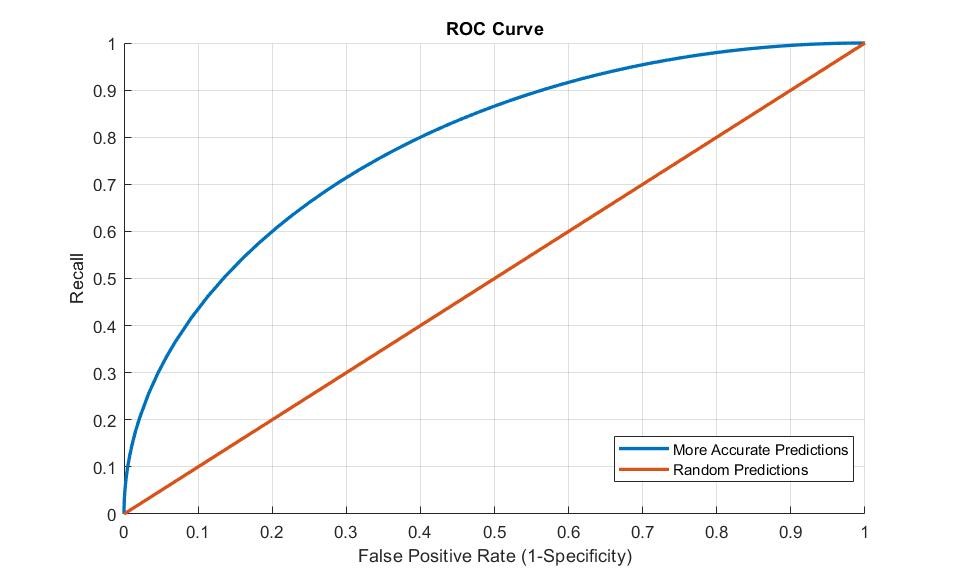}
\caption{ROC Curve}
\end{figure}

In the ROC curve, if the false positive rate in the horizontal axis is substituted with the number of false positive per image, the Free Operative Characteristic (FROC) is produced instead. 
\(F_1\) Score: As previously discussed, these metrics serve different evaluation goals. The \(F_1\) scores has been proposed to combine both precision and recall values into their harmonic mean:

\begin{equation}
F1=2 \frac{precision\times recall}
 {precision+recall}
\end{equation}

Dice coefficient: First defined for discrete sets, the Dice coefficient is just reformulation of the \(F_1\) score when applied to Boolean data with TP, FN, and FP definitions:

\begin{equation}
Dice=\frac{2TP}
 {2TP+FP+FN}
\end{equation}

For a given detection/segmentation task, the numerator represents twice the area of the ground truth and prediction overlaps, and the denominator shows the total pixel area of the ground truth and the prediction. 
Intersection over Union (IoU, Jaccard Index): IoU is used for thresholding the TPs in objection detection and segmentation tasks. It is defined as the intersection of the ground truth segmentation/bounding box and the predicted segmentation/bounding box over the union of those the ground truth and predictions:

\begin{equation}
IoU=\frac{Area of Overlap}
 {Area of Union}=\frac{TP}
 {TP+FN+FP}
\end{equation}

Bad predictions (FP and FN) are more penalized in IoU than in the Dice coefficient, thereby the IoU score of an algorithm is always less than or equal to that of the Dice coefficient. However, in an overall sense these metrics are quite similar and positively correlated, meaning given a set of algorithms for a specific task, both metrics will rank their performance in the same manner. The IoU of different classes in an image entry could also be averaged to calculate mean Intersection over Union (mIoU).
Average Precision (AP):  Average Precision and its variations are probably the most popular metrics among computer vision challenges. AP essentially computes the precision of an algorithm over recalls ranging from 0 to 1.  
In order to calculate the AP of an algorithm for a specific class and over a dataset, the data entries need to be first ranked with respect to their confidence score. The confidence score of a class could be the outputted probability of a classifier, or the IoU score of a segmentation/object detection algorithm. The mean of the precision values for the top-k ranked predictions with k ranging from 1 to n (the number of positive ground truths) forms the average precision:

\begin{equation}
AP = \sum_{k= 1}^{n}P(k)\Delta r(k)	
\end{equation}

Where \(P(k)\) is the precision score of the top-k predictions, and \(r(k)\) is the recall change between steps \(k\) and \(k+1\). It needs to be noted that the recall value at step k is defined as the number of TPs in the top \(k\) confident predictions over the total number of TPs in the dataset for the class in question. The above equation can be reformulated as:

\begin{equation}
AP = \frac{\sum_{k= 1}^{n}P'(k)}{total \: number \: of \: TPs}	
\end{equation}

Where \(P'(k)\) is defined in the same manner as \(P(k)\) but returns zero is the kth prediction is not a true positive. 
To measure the performance over several classes instead of one, the AP score could be averaged to compute the mean Average Precision (mAP):

\begin{equation}
mAP = \frac{\sum_{C= 1}^{C_{total}}AP(c)}{C_{total}}
\end{equation}

Where C\textsubscript{total} is the total number of classes. 
Panoptic Quality: The performance of panoptic segmentation tasks is measured using the proposed Panoptic Quality (PQ) metric. PQ incorporates both the detection and segmentation quality of algorithms and is formulated as:

\begin{equation}
PQ = \frac{\sum_{(p,q)\in TP}IoU(p,g)}{|TP|}\times 
\frac{|TP|}{|TP|+\frac{1}{2}|FP| + \frac{1}{2}|FN|}
\end{equation}

Where TP, FP, and FN correspond to the number of correct detections with 0.5 \(<\) IoU segmentation score, incorrect detections with IoU \(>\) 0.5, and unmatched ground truth segments (not detected or 0.5 \(>\) IoU) respectively. Note that the first fraction is the average of IoUs for the matched segmentations and does not take the bad predictions into account whereas the \(F_1\) score in the fraction to the right evaluates the detection quality with the \(\frac{1}{2}|FP|+\frac{1}{2}|FN|\) term penalizing bad and missed predictions. Cancelling TP in the two fractions we can write the PQ score as follows:

\begin{equation}
\begin{centering}
PQ = \frac{\sum_{(p,q)\in TP}IoU(p,g)}{|TP|+\frac{1}{2}|FP| + \frac{1}{2}|FN|}
\end{centering}
\end{equation}

The PQ metric is computed for every class and then averaged over all classes.

\section{Generic Object Recognition Datasets}
The number of the publicly available labelled datasets has soared in recent years. Such datasets are usually annotated through crowdsourcing. Amazon Mechanical Turk is a popular crowdsourcing platform, used for many of the famous datasets \citep{Deng2010, Krishna2017, Song2015}. For several of the datasets, there also exists a challenge to benchmark the performance of the state-of-the-art algorithms. In this section, a comprehensive overview of the datasets and the corresponding challenges focusing on the generic object recognition task have been provided. In Sec. 3.1, the four major large-scale object recognition datasets are discussed, with other generic datasets to follow in Sec. 3.2 and 3.3.

\subsection{Major Large-scale Datasets}
There are four major object recognition datasets widely adopted by researchers. The statistics of these datasets are provided in Table 1. Each datasets comes with a corresponding challenge held annually to evaluate and benchmark state-of-the-art algorithms. These challenges have played a critical role in the advancements of object recognition techniques in recent years, as the state-of-the-art algorithms emerge to outperform the winners from previous competitions. Figure 3 shows the rise in the performance of the winner algorithms in each of the four major challenges.

\begin{figure}[!h]
\centering
\includegraphics[scale=0.4]{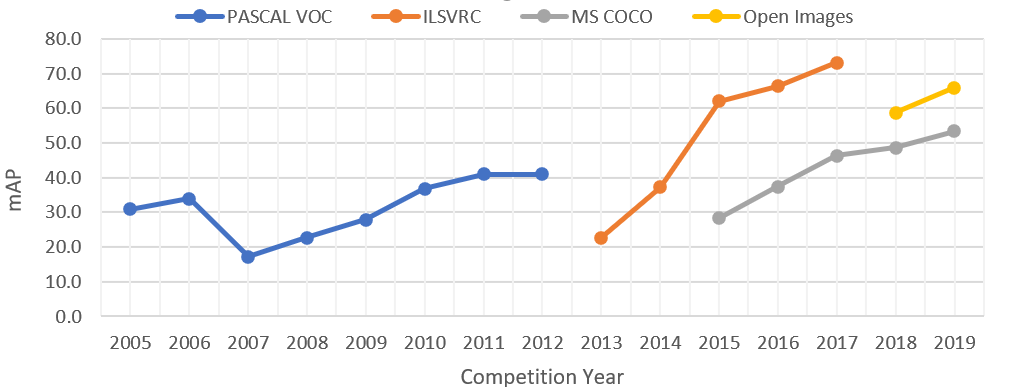}
\caption{Accuracy improvement of winner algorithms in the object detection track of major challenges. The fall in accuracy levels in PASCAL VOC 2007 is due to the increase of object classes from 4 to 20.}
\end{figure}

\begin{table*}[h]
\scriptsize
\caption{Dataset statistics for PASCAL VOC, ImageNet, MS COCO, and Open Images }
\centering
\begin{tabular}{|l|l|l|l|l|}
\hline
    Dataset & Number of Classes & Number of Images & Average Objects Per Image & First Introduced  \\
    \hline
    PASCAL VOC & 20    & 22,591 & 2.3   & 2005 \\
    \hline
    ImageNet & 21,841 & 14,197,122 & 3     & 2009 \\ \hline
    Microsoft COCO & 91    & 328,000 & 7.7   & 2014 \\ \hline
    Open Images & 600   & 9,178,275 & 8.1   & 2017 \\   
    \hline

\end{tabular}
\end{table*}

\subsubsection{Overview of Datasets}
\textbf {Pascal VOC}:
The earliest of the four is the Pattern Analysis, Statistical Modelling and Computational Learning Visual Object Classes (PASCAL VOC) dataset \citep{Everingham2010}. It was first released in 2005 with four object classes, which were then expanded to 20 in the following years. The PASCAL VOC annual challenges (2005-2012) \citep{Everingham2014} laid the foundation of the standard object recognition evaluation metrics, widely adopted by future competitions. The dataset fell out of fashion after the release of a significantly larger dataset named ImageNet, but it is still sometimes used as a convenient benchmark of newly developed algorithms because of its relatively small size. \\ 
\textbf{ImageNet}: Following the footsteps of PASCAL VOC, the ImageNet dataset was introduced in 2010 \citep{Deng2010}. The dataset contains 14 million annotated images in highly fine-grained categories. Samples are organized based on the hierarchical structure of WordNet \citep{Fellbaum1998} that is a lexical database that groups conceptually similar words into sets of cognitive synonyms (synsets). The database consists of 27 main classes such as fish, vehicles, and tools. Each of these classes are then divided into semantically arranged sysnset subcategories, resulting in over 21000 synsets. A subset of 1.2 million images of the dataset forms the baseline for the annual ImageNet Large Scale Visual Recognition Challenge (ILSVRC) \citep{Russakovsky2015}, held between 2010 and 2017. When first introduced, the only task in the challenge was image classification, but single-object localization and object detection tasks were also added in the later competitions. Due to its large size and extensive class hierarchy, ImageNet is still used to pretrain complex models (e.g.VGG \citep{Chatfield2014}, ResNet \citep{He2016}). However, ImageNet is one of the largest object recognition datasets to date although issues such as partial image annotations (only one annotated object per image) and lack of rich natural contextual information have instigated attempts to create more intricate datasets.\\
\textbf {Microsoft COCO}: Microsoft COCO \citep{Lin2014} is an effort to address the problems of ImageNet by offering a relatively small (328,000 images) but a much richer dataset. In order to improve generalizability, images are annotated with segmentation mask for objects in their contextual backgrounds instead of canonical views. The main argument for the inclusion of contextual environments in the dataset is that detection of many object categories depends on whether they are depicted in their natural surroundings or not. Specifically, the algorithms trained on a particular dataset will be more generalizable if the objects are presented in their natural environment. By training a model on both PASCAL VOC and MS COCO, and then testing it on the other dataset, it was shown that although the algorithm trained on PASCAL VOC performed better, the performance drop of the MS COCO trained algorithm on the test dataset was less significant; this indicates its superior generalizability. There has also been a focus on balancing the number of instances per category, ensuring that a threshold of 5000 entries is surpassed for all object classes. Additionally, a supplementary COCO stuff dataset \citep{Caesar2018} has been released that provides segmentations for surrounding-level classes like grass, sky, and floor, which lack distinctive structure or parts. The downside of the COCO dataset is that the object categories are entry-level, and fine-grained categorization is ignored so that each category includes at least a sufficient number of samples.\\ 
\textbf{Open Images}: Open Images \citep{Kuznetsova2018} is the most recent large-scale dataset that includes more than nine million labeled images, with bounding boxes for 600 object classes. Segmentation masks were also added to the 2019 release of the dataset. Moreover, a crowdsourced extension adds 478,000 labeled, but not segmented, images to the dataset. Another feature of the dataset is visual relationship annotations that capture meaningful interactions between pairs of objects, allowing for more holistic interpretations. Image classes include coarse-grained object classes, fine-grained object classes, events, scenes, materials and attributes (e.g., texture and color). Image class labels are divided into positive labels, instances that exist in the image, and negative labels, instances that do not appear in the image. By running classifiers on the dataset with and without the help of negative labels, it was shown that inclusion of negative labels considerably enhances classification accuracy.

\subsubsection{Challenge Tasks} In this section we analyze the development of the popular object recognition tasks covered in the corresponding challenges of the datasets discussed in section 3.3.1.  Table 2 lists the criteria, dataset attributes, and performance metrics of the four annual challenges. Each challenge consists of multiple tasks including classification, object detection and segmentation elaborated in the following. \\

\begin{table*}[!h]
  \scriptsize
  \centering
  \resizebox{\textwidth}{!}{
    \begin{tabular}{|l| c | c | c | c | c | p{6.5 cm} | p { 5 cm}|}
    
\hline
Challenge & Tasks Covered & Classes& Images & Annotated Objects & Years active & Task Description & Evaluation Metric \\ \hline

 \multirow{5}[1]{*}{  PASCAL VOC} & Image Classification & 20    & 11,540 & 27,450 & 2005 - 2012 & Absence/presence prediction of at least one instance of every class in each image & AP \\ 
      & Detection & 20    & 11,540 & 27,450 & 2005 - 2012 & Bounding box prediction for every instance of the challenge classes present in images & AP with \(IoU > 0.5\) \\ 
      & Segmentation & 20    & 2913  & 6929 & 2007 - 2012 & Semantic segmentation for the object classes & IoU \\
      & Action Classification & 10    & 4588  & 6278 & 2010 - 2012 & Bounding box prediction or single points for persons performing an action and annotate with the corresponding action label & AP over action classes  \\
      & Person Layout Taster & 3     & 609   & 850 & 2007 - 2012 & Body part (hands, head, feet) detection with bounding boxes & AP calculated separately for parts with \(IoU > 0.5\) \\ \hline
   \multirow{5}[0]{*}{ILSVRC} &Image Classification & 1000 & 1,331,167 & 1,331,167 & 2010 - 2014 & Classification for one annotated class per image & Binary class error over the top 5 predictions per image \\
       & Object Localization & 1000  & 573,966 & 657,231 & 2011 - 2017 & Bounding box prediction for only one object per image & Binary class and bounding box IoU error over the top 5 predictions \\
      & Object Detection & 200   & 476,688 & 534,309 & 2013 - 2017 & Bounding box prediction for all instances per image & AP with flexible recall threshold varied proportional to bounding box size \\
      & Object Detection from Video & 30    & 5,314 (video snippets) & - & 2015 - 2017 & Continuous bounding box prediction throughout video sequences & AP with flexible recall threshold varied proportional to bounding box size \\ \hline
    \multirow{5}[0]{*}{MS COCO} & Detection & 80    & 123,000+ & 500,000+ & 2015 - present & Instance Segmentation over object classes (things) & AP at IoU range of \([0.5:0.05:0.95]\) \\
      & Keypoints & 17    & 123,000+ & 250,000+ & 2017 - present & Simultaneous object detection and keypoint localization & AP based on Object Keypoint Similarity (OKS) \\
      & Stuff & 91    & 123,000+ & -     & 2017 - present & Pixelwise segmentation of background categories & mIoU \\
      & Panoptic & 171   & 123,000+ & 500,000+ & 2018 - present & Full segmentation of images (stuff and things) & Panoptic Quality \\
      & DensePose & - & 39,000 & 56,000 & 2019 - present & Human body segmentation and mapping all the pixels of the body to a template 3D model & AP based on Geodesic Point Similarity (GPS) \\ \hline
\multirow{3}[1]{*}{Open Images} & Object Detection & 500   & 1,743,042 & 12,421,955 & 2018 - present & Hierarchical-based bounding box detection & mAP \\
      & Instance Segmentation & 300   & ~ 848,000 & 2,148,896 & 2018 - present & Instance segmentation over object classes; negative labels included to refine training & mAP at \(IoU > 0.5\) \\
      & Visual Relationship Detection & 57    & 1,743,042 & 380,000 relationship triplets & 2018 - present & Labeling images with relationship triplets containing the interacting objects and the action class & A weighted sum of mAP and recall of number of relationships at \(IoU > 0.5\) 
\\ \hline
    \end{tabular}}
      \caption{Challenge description for PASCAL VOC, ILSVRC, MS COCO, and Open Images}
\end{table*}
\textbf{Classification}: This task is only included in PASCAL VOC and ILSVRC challenges. More recent challenges do not include classification due to its relative simplicity. In PASCAL VOC, two competitions were organized. Participants could either train their models on the annotated VOC training/validation dataset, or use their pre-trained models in a separate competition. In ILSVRC, algorithms were tested on a list of images, each having one ground-truth annotation. Since there might be several objects in a single image, it was possible that the algorithm could not distinguish the desired object for labeling. As a result, for each image, the algorithm was allowed to label up to five objects, and the evaluation would then be made based on whether the annotated object was listed between the predictions for that particular image. The error of an algorithm was measured with the top-5 criterion, defined as the fraction of test examples for which the algorithm does not classify the desired object in its top five predicted labels. The error is given by:

\begin{equation}
PQ = \frac{1}{n}\sum_{k}\min_{j}d(l_j,g_k)
\end{equation}

where \(d(x,y)=0\) if   \(x=y\)  and 1 otherwise. The top-5 criterion has been also been adopted by other classification datasets for benchmarking proposes \citep{Barbu2019,Horn2018}. \\
\textbf{Object detection}: All the four competitions offer this track. However, bounding box annotations will probably be sidelined for segmentation masks. As of 2018, this task is not featured in the MS COCO challenges. All competitions used a variation of mAP with IoU thresholding used to assess the algorithms. In essence, the difference between the evaluation criteria of these competitions is the way the IoU threshold is defined. PASCAL VOC and Open Images use a threshold of 0.5 whereas in ILSVRC the threshold value increases proportional to the size of each object. Since all the datasets have images containing unannotated object classes, predictions for such classes are not assessed.\\
\textbf{Segmentation}: This task is aimed for pixel-wise object classification of test images. All the competitions except for ILSVRC offer this track. In PASCAL VOC, models were required to classify every pixel as belonging to one of the 20 specified classes or background. A 5-pixel wide margin around objects were labeled as void so as to identify the border between the objects as and the background. Difficult objects (far, small, or highly occluded), were also segmented with a void mask and hence excluded from the training and the test datasets. Open Images follows the same evaluation metrics as its detection task. segmentation masks are evaluated using mAP with mask-to-mask or box-to-box IoU greater than 0.5 (depending on whether the prediction is tested against the ground-truth segmentation or bounding box) over 300 challenge classes. MS COCO tackles the segmentation task more comprehensively. Along with the standard instance segmentation, two other segmentation tasks are presented: stuff segmentation and panoptic segmentation. The former is aimed towards pixel-wise segmentation of the background-level categories such as grass and wall, instances of which are hard to separate from one another. Stuff classes are divided into two general outdoor and indoor categories, each of which are narrowed down in two more levels. The panoptic segmentation task is aimed for achieving fully segmented images by combining semantic segmentation (pixel-wise class labels) and instance segmentation (masking object instances). The performance of panoptic segmentation tasks is measured using the Panoptic Quality metric.\\
\textbf{Competition-specific tasks}:In addition to the three mainstream tasks mentioned above, there are competition-specific tasks that focus on a particular recognition application, e.g. human body part recognition or action classification. 
PASCAL VOC’s Person Layout Taster is a human detection task reported with a confidence score. Body parts were also required to be identified with bounding boxes; the requirements for prediction to be considered as a true positive were expanded to correct detection of body parts (legs, hands, and head) with their corresponding bounding boxes, having an IOU score of over 50\%. \\
Open Images offers the Visual Relationship Detection, where participants are required to identify visual relationship triplets in two scenarios. In the first scenario, for each relationship class, algorithms generate two bounding boxes for the interacting objects, their corresponding labels, and a label for the visual relationship. In the second scenario, the output is a bounding box encapsulating the two interacting objects, and three labels, indicating the relationship triplet. Finally, a weighted sum of the recall of the first, and the mAPs of both scenarios is computed to rank the performance of the algorithms. \\
MS COCO’s Keypoint Detection is a human pose estimation task. It involves localization of important body keypoints that well describe the person’s pose. Different types of keypoints, such as eyes and hips, are labeled to describe person’s pose in unspecified and uncontrolled conditions. The accuracy of the keypoint detection algorithms is measured by calculating the mAP over a range of similarity criteria called Object Keypoint Similarity (OKS) defined as:

\begin{equation}
OKS = \frac{\sum_{i}e^\frac{-d_i^2}{2s^2k_i^2}\delta(v_i>0)}{\sum_{i}\delta(v_i>0)}
\end{equation}

where d is the Euclidean distance between the ground truth and the predicted keypoint, s is the scale parameter which determines the sensitivity of this similarity criterion to the surface of the keypoint, and k is the keypoint constant, which is used to take into account the importance of the keypoint type. For instance, since eye location is more important than hip location, \(k_{eye}<k_{hip}\). Finally, \(v_i\) is the visibility factor which is zero for obscured ground-truth keypoints and 1 otherwise. This means that the occluded keypoints will not be used for benchmarking the performance of predictions. The mAP in the range of \(OKS=[0.5:0.05:0.95]\) is then calculated to determine the best method in this task.

\subsection{Other Object Recognition Datasets}
Besides the datasets discussed in Sec. 3.1 many other object recognition datasets also exist that are used to a lesser extent but may still be useful for research and development purposes.

\subsubsection{Object Detection Datasets}
As previously discussed, the most important object detection datasets in terms of impact and size (at the time of their release) are PASCAL VOC and Open Images. However, there are numerous other recognition datasets with bounding box annotations. Many of these datasets, such as Caltech Pedestrian, and KITTI are application-specific and are therefore discussed in Sec. 4. Since the rapid improvement of state-of-the-art recognition algorithms demand for holistic scene annotations, there has been a shift in focus towards segmentation mask annotations. However, bounding boxes are still considerably less time-consuming to annotate and also well-defined, being more robust to inconsistencies arising from subjective annotations of different human annotators working on the same dataset \citep{Djavadifar2020}. Inconsistencies exist even between two segmentation annotations of the same image by one annotator \citep{Zhou2017}. Consequently, bounding box annotations are still used in many datasets. In Table 3, the information of some generic object detection datasets is provided. We ignored non-commercial, e.g. Google’s internal JFT-300m \citep{Chollet2017}, or sparsely annotated (e.g., YouTube Objects \citep{Prest2012}, YFCC100M \citep{Thomee2016}) datasets.  

\begin{table*}[!h]
  \centering
  \scriptsize
    \begin{tabular}{|l |c| c| c| c|}
    \hline
    Dataset & Number of Images & Classes & Number of Bounding Boxes &Year \\ \hline
    Caltech 101 \citep{Fei-Fei2004} & 9,144 & 102   & 9144 & 2003 \\ \hline
    MIT CSAIL \citep{Torralba2004} & 2,500 & 21    & 2500 & 2004 \\ \hline
    Caltech 256 \citep{Griffin2007} & 30,307 & 257   & 30,307 & 2006 \\ \hline
    Visual Genome \citep{Krishna2017} & 108,000 & 76,340 & 4,102,818 & 2016 \\ \hline
    YouTube BB \citep{Real2017} & 5.6 m & 23    & 5.6 m & 2017 \\ \hline
    Objects 365 \citep{Shao2019} & 638,000 & 365   & 10.1 m & 2019 \\ \hline
    \end{tabular}
  \caption{Generic object detection datasets (except for those covered in Sec. 3.1).}
\end{table*}

\subsubsection{Object Segmentation Datasets}
Object segmentation datasets offer either instance-level or semantic segmentation masks. Stuff classes are either not annotated or labeled as background. A list of impactful generic object segmentation datasets is provided in Table 4. Since stuff could also be segmented, there has been an inclination towards creating datasets with fully segmented images (these datasets are discussed in Sec.3.3). However, object segmentation datasets are still popular for Video Object Segmentation (VOS). Pixel-wise annotation of foreground objects with temporal correspondence between frames is vital to applications such as action recognition, motion tracking, and interactive video editing. Most VOC datasets are small in size, usually containing only a dozen of video sequences \citep{inproceedings,10.1145/2816795.2818105,6751383, 10.1007/978-3-642-15555-0_21, 6682905}. DAVIS \citep{Caelles2019} and YouTube-VOC \citep{Xu2018} are two recent entries established as the main benchmarks for VOC. Corresponding challenges are held annually for both of the datasets, where algorithms are evaluated based on IoU and \(F_1\) scores. As of 2020, an instance segmentation challenge is also held for the LVIS \citep{Gupta2019} dataset with the same AP evaluation metric as MS COCO.

\begin{table*}[!h]
  \centering
  \scriptsize
  \resizebox{\textwidth}{!}{

    \begin{tabular}{|l| c | c | c | c | c | p{9 cm} |}
    \hline
    Dataset & Number of Images  & Classes & Number of Objects & Year & Challenge & Description \\ \hline
    SUN \citep{Xiao2010} & 130,519 & 3819  & 313,884 & 2010  & No    & The main purpose of the dataset is scene recognition, however instance-level segmentation masks have also been provided \\ \hline
    SBD \citep{Hariharan2011} & 10,000 & 20    & 20,000 & 2011  & No    & Object contours on the train/validation images of PASCAL VOC \\ \hline
    Pascal Part \citep{chen2014} & 11,540 & 191   & 27,450 & 2014  & No    & Object part segmentations for all the 20 class in the PASCAL VOC dataset \\ \hline
    DAVIS \citep{Caelles2019} & 150 (videos) & 4     & 449   & 2016  & Yes   & A video object segmentation dataset and challenge focused on semi-supervised and unsupervised segmentation tasks \\ \hline
    YouTube-VOS \citep{Xu2018a} & 4,453 &94 & 7,755 & 2018 & Yes & videos object segmentation dataset collected of short (3s-6s) video snippets \\ \hline
    LVIS \citep{Gupta2019} & 164,000 & 1000  & 2 m & 2019  & Yes   & Instance segmentation annotations for a long-tail of classes with few samples \\ \hline
    LabelMe\citep{Russell2008} & 62,197 & 182   & 250,250 & 2005  & No    & Instance-level segmentation; some of the background classes have also been annotated \\ \hline
    \end{tabular}}
  \caption{Object segmentation datasets.}
\end{table*}

\subsection{Object Recognition in Scene Understanding Datasets}
A comprehensive understanding of an image requires the determination of scene characteristics in addition to mere object recognition. Although the identification of object categories may provide some clues about the context of an image (e.g., beds and drawers in an image possibly represent a bedroom), but such information alone can be misleading since many common objects are shared across different contexts. Additionally, the amorphous background properties of an image (sometimes referred to as stuff categories versus objects/things) are ignored in object recognition datasets, despite being crucial in providing deeper visual information such as geometric relationship of the objects or contextual reasoning (e.g. trees are more likely to appear on grass rather than street). As a result of these shortcomings, many scene-centric datasets have been proposed in the computer vision community. \\
Scene recognition datasets are directly tailored for scene type identification by including classification labels for each of the images, describing the corresponding scene of that particular image. Scene is defined as a place where a human can navigate \citep{Xiao2010,Zhou2017a}. Classification labels do not allow for intra-class variations, which can be significant in the case scene types (e.g., the variations between different forests). Additionally, multiple categories might exist in an image, and limiting the number of annotations to only one classification label will result in an inevitable loss of information for some scenes. In a few scene recognition datasets, additional annotations have been added to address these issues. In a subset of SUN database \citep{Xiao2010}, different sub-scenes are annotated to allow for the detection of multiple scenes in one image. Creators of OpensSurfaces \citep{Bell2013} took a different approach by annotating scene labels for segmented interior surfaces in each scene. An attribute-based categorization is used in SUN Attributes \citep{Patterson2012} to take into account the inter-class scene variations. The most popular scene recognition datasets are presented in Table 5. A 365-class subset of Places2 \citep{zhou} was used for the Places scene classification challenge with a top-5 evaluation criterion similar to ILSVRC.

\begin{table*}[!h]
  \scriptsize
  \centering
  \resizebox{\textwidth}{!}{
    \begin{tabular}{|l| c | c | c | c | p{9 cm} |}
    \hline
    Dataset & Number of Images  & Classes & Additional Annotations &Year & Description \\ \hline
    15-Scene \citep{Lazebnik2006} & 4,485 & 15    & -     & 2006  & One of the earliest major scene classification datasets \\ \hline
    MIT Indoor67 \citep{Quattoni2010} & 15,620 & 67    & - & 2009  & Indoor scene classification in 5 main groups: Store, Home, Public Space, Leisure, and Working Place \\ \hline
    SUN \citep{Xiao2010} & 130,519 & 899   & 313,844 SM (Objects) & 2010  & Classification dataset of navigable scenes with additional object recognition annotations  \\ \hline
    SUN Attribute \citep{Patterson2012} & 14,000 & 700   & 102 binary attributes per image & 2012  & Attribute-based representation of scenes for a subset of the original SUN database \\ \hline
    Open Surfaces \citep{Bell2013} & 25,357 & 160 & 71,460 SM & 2013 & Segmented surfaces in interior scenes with texture and material information \\ \hline
    Places2 \citep{Zhou2017a} & 10 m  & 476   & -     & 2017  & Classification of scenes bounded by spaces a human body would fit; comes with binary attributes \\
\hline
    \end{tabular}}
      \caption{Popular scene recognition datasets.}
\end{table*}

Scene parsing datasets are another type of scene understanding datasets that go beyond the recognition of object classes by including pixel-wise segmentations for all the stuff and object categories in an image. Compared to object recognition datasets, these datasets require more extensive class categorizations and much richer annotations, thus often sacrificing the scale. The majority of scene parsing datasets focus on outdoor scenes \citep{Cordts2016,Gould2009,10.1007/978-3-642-15555-0_26,Wrenninge2018}. Indoor scene parsing is still a more challenging task compared to outdoor scene parsing due to the higher diversity of scene categories and smaller amount of training data \citep{Tighe2013}. The prominent scene parsing datasets are summarized in Table 6.  Among the listed datasets, scene parsing challenges have been held for ADE20K, MS COCO Stuff, and SUN RGB-D, all using IoU as the evaluation metric.

\begin{table*}[!h]
  \scriptsize
  \centering
  \resizebox{\textwidth}{!}{
    \begin{tabular}{|l| c | c | c | c | c | p{9 cm} |}
    \hline
    Dataset & Number of Images  & Stuff Classes & Object Classes & Year & Challenge & Highlights \\ \hline
    MSRC 21 \citep{10.1007/11744023_1} & 591   & 6     & 15    & 2006  & No    & One of the earliest semantic scene parsing datasets; images were later used in PASCAL VOC and Stanford Background \\ \hline
    Stanford Background \citep{Gould2009} & 715   & 7     & 1     & 2009  & No    & Outdoor scene parsing dataset collected from LabelMe, MSRC, and PASCAL VOC geometric features also included \\ \hline
    SiftFlow \citep{Liu2015b} & 2688  & 18    & 15    & 2009  & No    & An early dataset on outdoor environment scene parsing labeled using LabelMe \\ \hline
    Barcelona \citep{10.1007/978-3-642-15555-0_26} & 15,150 & 31    & 139   & 2010  & No    & A subset of the LabelMe dataset \\ \hline
    NYU Depth V2 \citep{Silberman2012} & 1,449 & 26    & 893   & 2012  & No    & Parsing of 464 cluttered indoor scenes; depth maps also included; semantic segmentation masks for objects \\ \hline
    SUN+LM \citep{Tighe2013} & 45,676 & 52    & 180   & 2013  & No    & A fully annotated subset of LabelMe and SUN datasets with both indoor and outdoor images \\ \hline
    PASCAL Context \citep{Mottaghi2014} & 10,103 & 152   & 388   & 2014  & No    & Pixel-wise semantic segmentation on the PASCAL VOC dataset; 520 new object and stuff categories were added to the original dataset \\ \hline
    SUN RGB-D \citep{Song2015} & 10,335 & 47    & 800   & 2015  & Yes   & Indoor scene parsing dataset and benchmark; 3D bounding boxes also provided \\ \hline
    Cityscapes \citep{Cordts2016} & 25,000 & 14 & 13 & 2016 & No & Images captured from a vehicle driving in urban environments across 50 cities in different weather conditions in Europe; instance-level segmentation masks  \\ \hline
    ADE20K \citep{Zhou2017} & 25,210 & 1,242 & 1,451 & 2017  & Yes   & Includes object part labels, and attributes; instance-level segmentation masks \\ \hline
    Synscapes \citep{Wrenninge2018} & 25,000 & 14    & 13    & 2018  & No    & Photo-realistic synthetic scene parsing of urban environments; annotation categories are the same as Cityscapes instance-level segmentation masks \\ \hline
    MS COCO Stuff \citep{Caesar2018} & 163,957 & 91    & 80    & 2018  & Yes   & Pixel-wise semantic segmentation for the entire MS COCO dataset  \\
   \hline
    \end{tabular}}
  \caption{Scene parsing datasets.}
\end{table*}

Apart from generic object recognition datasets, a number of scene understanding datasets include annotations only for a portion of objects that provide meaningful information about the context of the scene. These include salient  and camouflaged object detection datasets. Salient object detection refers to detecting the most salient objects within images. Salient objects form the most visually distinctive regions of an image and are most likely to be the first objects to catch the attention of human observers. As a result,   accurate detection of such objects can provide a quick and efficient understanding of the context of the scene. Salient object detection has been widely studied throughout the years \citep{LiSalient,LiSalient2}, and a number of popular datasets have emerged throughout the years to fuel research in the field, see Table 7. Camouflage object detection on the other hand, is aimed to go beyond human perception capabilities by identifying object categories that are visually indistinguishable or hard to distinguish from background elements. Camouflaged object detection is still an understudied area primarily due to the inherent annotation difficulty and subsequent lack of large-scale datasets geared towards the task \citep{COD10,ZhaiCam}. The very few predominant datasets in the field include CHAMELEON \citep{CHAM}, CAMO \citep{CAMO}, and COD10K \citep{COD10}.

\begin{table*}[!h]
  \scriptsize
  \centering
  \resizebox{\textwidth}{!}{
    \begin{tabular}{|l| c | c | c | c | p{9 cm} |}
    \hline
    Dataset & Number of Images  & Objects per Image & Annotation Type & Year & Highlights \\ \hline
MSRA-A  \citep{LiuSalient} & 20,840 & 1-2                                                             & BB                                                          & 2007 & One of the earliest large-scale salient object detection datasets. 5,000 and 10,000 image subsets   of the dataset were further created with pixel-wise annotations
\\ \hline
ASD  \citep{ASD}       & 1,000  & 1                                                               & SM                                                          & 2009 & A subset of   MSRA-A dataset with pixel-wise annotations          \\ \hline                                 
THUR15K \citep{THUR15K} & 15,000 & 1                                                               & SM                                                          & 2013 & Images  are crawled from the web using butterfly, coffee mug, dog jump, giraffe, and plane  keywords
\\ \hline
DUT-OMRON \citep{DUT-OMRON}   & 5,172  & 1-4 & BB                                                          & 2013 & Saliency detection dataset in complex environments, more challenging compared to MSRA \\ \hline
PASCAL-S \citep{PASCAL-S}    & 850    & 1-5                                                             & SM                                                          & 2014 & A subset of the PASCAL VOC dataset. Also includes human gaze annotations
\\ \hline
ECSSD  \citep{ECSSD}      & 1,000  & 1-4                                                             & SM                                                          & 2015 & Focused on providing images with challenging natural backgrounds
\\ \hline
HKU-IS  \citep{HKU}     & 4,447  & 1-4                                                             & SM                                                          & 2015 & Low contrast images   often containing multiple disconnected salient objects that touch the image   boundary                                     
\\ \hline
SOS  \citep{SOS}    & 14,000 & 0-4+                                                            & BB                                                          & 2015 & Each image is   labelled as having 0,1,2,3, or 4+ salient objects.
\\ \hline
DUTS \citep{DUTS}        & 15,572 & 1-4+                                                            & SM                                                          & 2017 & Popular large-scale dataset with images from the ImageNet and SUN datasets
\\ \hline
XPIE \citep{XPIE}        & 10,000 & 1-4                                                             & SM                                                          & 2017 & Objects are   selected as salient if they have high objectness, topological simplicity, and less than four similar candidates within the image
\\ \hline
SOC \citep{SOC}        & 6,000  & 0-4                                                             & SM                                                          & 2018 & Half of the dataset does not contain salient objects. Object category and challenging factor annotations are also provided.
\\ \hline
COCO-CapSal \citep{CapSal} & 6,724  & 1-4                                                             & SM                                                          & 2019 & Contains 80 salient object categories collected from the Microsoft COCO dataset. Also includes human gaze annotations                       

\\ \hline
    
    \end{tabular}}
  \caption{Popular salient object detection datasets.}
\end{table*}

\section{Fine-grained Object Recognition Datasets}
One of the issues associated with generic object recognition datasets is class imbalance. Since many of these datasets are created using online sources, namely Flicker, there is a tendency to have a greater number of instances for categories that are more common on the web, such as airplanes, and cars. Additionally, large-scale datasets may lack a fine-grained categorization, and even if they are fine-grained (e.g., ImageNet), the number of instances naturally drops as we move down a hierarchical class categorization. As a result, the performance of an algorithm will suffer if it is trained on a generic dataset that is not as dense in the context in which the algorithm is used as it is in other categories. This problem can be addressed by fine-tuning the algorithm on datasets with richer annotations in the specific application where it is deployed. More precisely, first a backbone network (e.g., AlexNet, VGG16, ResNet) is chosen. Trained on large-scale generic object recognition datasets, such backbone architectures have already learned the invariant features applicable to all object recognition tasks and applications. Then, the baseline network is trained on a fine-grained dataset to tune the parameters according to the desired application. \\
The highly investigated object recognition applications, the corresponding datasets, and competitions are discussed in the following.
\subsection{Autonomous Driving}
Perception is one of the main components of Autonomous Vehicle (AV) navigation \citep{VanBrummelen2018}. Success in path planning, control, localization and mapping of the vehicle depends on a robust visual understanding of the surrounding environment. In the past 20 years, there have been vast research efforts on AV-related perception problems such as object recognition (e.g., traffic lights, pedestrians, and other vehicles), object tracking, visual odometry/SLAM, and optical flow. Since the early research developments on autonomous vehicles, there has been a focus on annotated datasets for relevant object recognition tasks. Such datasets are collected either using drones (bird’s eye view images), traffic cameras, or by sampling frames from hours of videos recorded by onboard cameras of one or a fleet of vehicles.  The range of tasks covered in AV perception datasets is truly extensive. In accordance with the scope of the paper, we focus on the object recognition statistics of datasets.\\
The majority of AV perception datasets are taken in street-view, being similar to the viewpoint of perception sensors mounted on AVs. While some of the datasets only offer annotated RGB images (e.g., CamVid \citep{Brostow2008}, Cityscapes \citep{Cordts2016}, Mapillary Vistas \citep{Neuhold2017}, \(D^2\)-City \citep{Caesar2019}, BDD100k \citep{Yu2018}), a number of AV street-view datasets are multimodal, e.g., KITTI dataset \citep{Geiger2013}, Apolloscape dataset \citep{Wang2019}, Waymo \citep{Sun2019}, nuScenes \citep{Caesar2019}, Oxford-1000km \citep{Maddern}, Lyft V5 \citep{Kesten2019}. These datasets are collected using a research vehicle equipped with a variety of sensors such as LiDAR, RADAR, RGB cameras, inertial measurement units, and driven in various locations, weather conditions, and time of the day. While incorporating valuable layered perception data, multimodal datasets are expensive to collect and bring about other difficulties such as correct integration, synchronization and calibration of sensors.\\
The main annotation choice for street-view datasets is 2D and 3D bounding boxes \citep{Caesar2019,Geiger2012,Pham2019, Chang2019} as it is more convenient for object tracking purposes \citep{DiDi2019}. Multimodal datasets usually offer a temporal correspondence between the annotated objects in different perception modules. Several datasets are annotated with rich pixel-wise segmentation masks instead of bounding boxes. CamVid \citep{Brostow2008a} is the first dataset offering semantic segmentation. Apolloscape \citep{Wang2019} provides semantic segmentation masks for 140k camera images captured in diverse traffic conditions. The CityScapes dataset \citep{Cordts2016} is the first dataset with significantly large numbers of pixel-wise labelled frames. It has been collected through 50 different cities depicting varied driving conditions. The Mapillary Vistas dataset \citep{Neuhold2017} surpasses the amount and diversity of labelled data compared to Cityscapes \citep{Girshick2014}. The virtual Semantic KITTI dataset \citep{Behley2019} provides sequential images with depth information and pixel-wise labelled dense point cloud video frames. BDD-100k dataset \citep{Yu2018} is focused on frame annotation and annotation tool provision. Diversification of driving scenarios is another important factor with regards to street-view AV datasets. A number of datasets are collected with an emphasis on diverse weather and illumination conditions, e.g., D2 city \citep{Caesar2019}, Mapillary Vistas \citep{Neuhold2017}, and BDD-100k dataset \citep{Yu2018}. There are also datasets that are collected by ego-vehicles driven in various cities or even countries. The details of the most popular street-view datasets are provided in Table 8.

\begin{table*}[!h]
  \scriptsize
  \centering
  \resizebox{\textwidth}{!}{
    \begin{tabular}{|l| c | c | c | c | c | p{9 cm} |}
    \hline
    Dataset & Year & Location& Annotated frames &Number of Classes & Object Annotations &Highlights \\ \hline KITTI \citep{Geiger2013}  & 2012  &Karlsruhe, Germany &15k & 8 & 200k 3D BB & Pioneer benchmark dataset for 3D object detection; multimodal  \\ \hline
    Cityscapes \citep{Cordts2016} & 2016 & 50 cities in EU& 25k & 27 & 65k SM & Rich annotations provided; scene variability and complexity; depth information provided \\ \hline
              BDD 100k \citep{Yu2018} & 2017 &NY, SF & 100k & 40 Objects, 8 Lanes & 1.8M BB & Diversified in location and weather conditions; instance segmentation masks provided for 10k images of the dataset \\ \hline
    KAIST \citep{Choi2018} & 2018 & Seoul & 8.9k &3 & 308k BB &All-day time conditions (e.g., sunrise, morning, noon); multimodal  \\ \hline
    ApolloScape \citep{Wang2019} & 2018 & 4x China & 144k & 25 Objects 28 Lanes & 70k 3D BB &Contains lane markings based on the lane colors and styles; instance level annotations are available \\ \hline A*3D \citep{Pham2019} & 2019 & Singapore & 39k & 7 & 230k 3D BB & Focused on pedestrian detection; high driving speed and low annotation speed \\ \hline Argoverse \citep{Chang2019} & 2019 & Miami, Pittsburgh & 22k & 15 & 993k 3D BB & Focused on 3D object tracking and motion forecasting; annotated HD semantic maps included \\ \hline
    Automative RADAR \citep{Meyer2019} & 2019 & Germany & 500 & 7 & 3000 3D BB & RADAR data and object detection based on RADAR data \\ \hline H3D \citep{Patil2019} & 2019 & San Francisco & 27k & 8 & 1.1M 3D BB & Full-surround 3D multi object detection and tracking \\ \hline nuScenes \citep{Caesar2019} &2019 & Boston & 40k & 23 & 1.4M 3D BB & First dataset providing 3D dataset with attribute annotations; first to provide RADAR data; rich multimodal information \\ \hline
Waymo \citep{Sun2019} & 2019 & 3x USA &200k  &4 & 9.9M BB, 12 3D BB & Diverse weather and geographic conditions in the US; multimodal \\ \hline
    Mapillary Vistas \citep{Neuhold2017} & 2017 & Global & 25k & 152 & 8M SM & Scene-parsing with instance-level object segmentation with a diverse geographic, weather, season and daytime extent \\ \hline
    Lyft L5 \citep{Kesten2019} & 2019 & Paolo Alto & 46k & 9 & 1.3M 3D BB & Multimodal captured by a fleet of vehicles; an annotated LiDAR semantic map is provided \\ \hline
     \(D^2\)-City \citep{Che2019} & 2019 & China & 700k & 12 & ~ 50k BB & Sampled from dashcam video sequences; Bounding box annotations \\
             \hline
    \end{tabular}}
    \caption{Popular street-view autonomous driving datasets. BB = Bounding Boxes, SM = Segmentation Masks.}
\end{table*}

A number of datasets cover specific recognition problems in the AV domain e.g., pedestrian detection, car detection and traffic sign detection. Pedestrian detection has been heavily investigated in autonomous driving and other applications such as video surveillance. Such datasets could be divided into two categories. ‘Person’ datasets contain people instances in all kinds of poses while ‘Pedestrian’ datasets only consider upright people (walking and standing) typically viewed from more restricted viewpoints but often containing motion information and detailed labelling \citep{Gan2017}. The famous pedestrian detection datasets are described in Table 9.

\begin{table*}[!h]
  \scriptsize
  \centering
  \resizebox{\textwidth}{!}{
    \begin{tabular}{|l| c | c | c| c | p{9 cm} |}
    \hline
    Dataset Dataset & Year & Number of Cities & Number of Images & Number of Pedestrians & Highlights \\ \hline
    CityPersons \citep{Zhang2017} & 2017  & 27 cities in EU & 5000  & 35016 & Built on top of the Cityscapes dataset \\ \hline
    INRIA \citep{Dalal2005} & 2005  & -     & 614   & 902   & Occlusion labels included \\ \hline
    Caltech \citep{Dollar2010} & 2009  & 1     & 250,000 & 2300  & Temporal correspondence and occlusion labels included; sampled from 10 hours of video \\ \hline
    MIT \citep{Papageorgiou2000} & 2000  & -     & 1800  & 1800  & Labelled using the LabelMe annotation tool \\ \hline
    EuroCity \citep{Braun2018} & 2018  & 31 cities in EU & 47,000 & 238,000 & Largest pedestrian detection dataset to date \\ \hline
    NightOwls \citep{Neumann2019} & 2018  & 7     & 32    & 55,000 & Pedestrian detection at night time; attribute annotations include pose, occlusion, and height \\ \hline
    Daimler \citep{Enzweiler2009} & 2009  & 1     & 21,790 & 56,492 & Occlusion attributes provided; monocular images \\
             \hline
    \end{tabular}}
      \caption{Pedestrian detection datasets. Number of images does not include unannotated images. Unique pedestrians are considered for the number of pedestrians.}
\end{table*}

Bird’s eye view datasets are captured using cameras mounted on drones, or from traffic signal lights or buildings in suitable vantagepoints over vehicles and pedestrians. Examples of such datasets are INTERACTION \citep{Zhan2019}, NIGSM \citep{Coifman2017}, HighD \citep{Krajewski2018} and KITTI bird’s eye view. Bird’s eye datasets are usually collected with the purpose of vehicle and pedestrian behavior prediction, however since many of them are annotated, they could also be studied under the object recognition dataset category. A list of the popular bird's eye view datasets with object annotations can be found in Table 10. 

\begin{table*}[!h]
  \scriptsize
  \centering
  \resizebox{\textwidth}{!}{
    \begin{tabular}{|l| c | p{2 cm} | p{3 cm}| p{3 cm} | p{9 cm} |}
    \hline
        Dataset & Year & Location & Road span/Area & Size of data & Highlights \\ \hline
    NGSIM \citep{Coifman2017} & 2005  & USA   & 500-640m Span of road & 90 min & Video cameras attached to the adjacent buildings; speed levels more than 75km/h are not included in the dataset \\ \hline
    High D \citep{Krajewski2018} & 2017  & Germany & 420m Span of road & 16.5 hours & Drone based dataset with five scenario description layers, the first 3 layers include static scenario description, 4th layer includes dynamic description,5th layer includes environment conditions \\ \hline
    The inD \citep{Bock2019} & 2017  & Germany & Altitude 100m 80x40 meters to 140x70 meters & 10 hours of video recording & The dataset includes more than 11,500 road users including pedestrians, bicyclists and vehicles at intersections \\ \hline
    INTERACTION \citep{Zhan2019} & 2019  & USA, China, Bulgaria, Germany  & n/a   & 365min+ 433min+133min + 60min & Data collected from drones and traffic cameras; Multimodal \\ \hline
    AU-AIR \citep{Bozcan2020} & 2019  & Aarhus, Denmark & Flight altitude (5m to 30m) and camera angle 45 to 90 degree & 2 hours  & Multimodal; compare and contrast of natural and aerial images in the context of object detection  \\
                 \hline
    \end{tabular}}
      \caption{Bird's eye view datasets}
\end{table*}

A number of challenges have also been held to boost research in recognition tasks. Table 13 provides the details of the the predominant AV-related object recognition datasets.

\subsection{Medical Imaging}
The promotion of computer vision applications in the medical field has made a significant improvement to healthcare industry all over the world. Artificial intelligence (AI) in medicine can be generally classified into two categories, virtual and physical \citep{Amisha2019}.  A virtual dataset is made up of images taken by medical equipment such as X-ray, and computed tomography scanner. These datasets aim to increase the accuracy and speed of certain diseases diagnosis. Thus, the doctors can plan treatment in advance and treat the patients effectively. A physical dataset consists of videos filmed during surgery procures. The goal of a physical dataset is to promote the implementation of computer and robots to surgery. \\
Numerous datasets and benchmarks have been produced using various medical imaging equipment for different organs. Figure 4 shows different types of medical images included in mentioned datasets, and a summary of major datasets and challenges could be found in Tables 11 and 12 respectively. 

\begin{figure}[!t]
\centering
\includegraphics[scale=0.4]{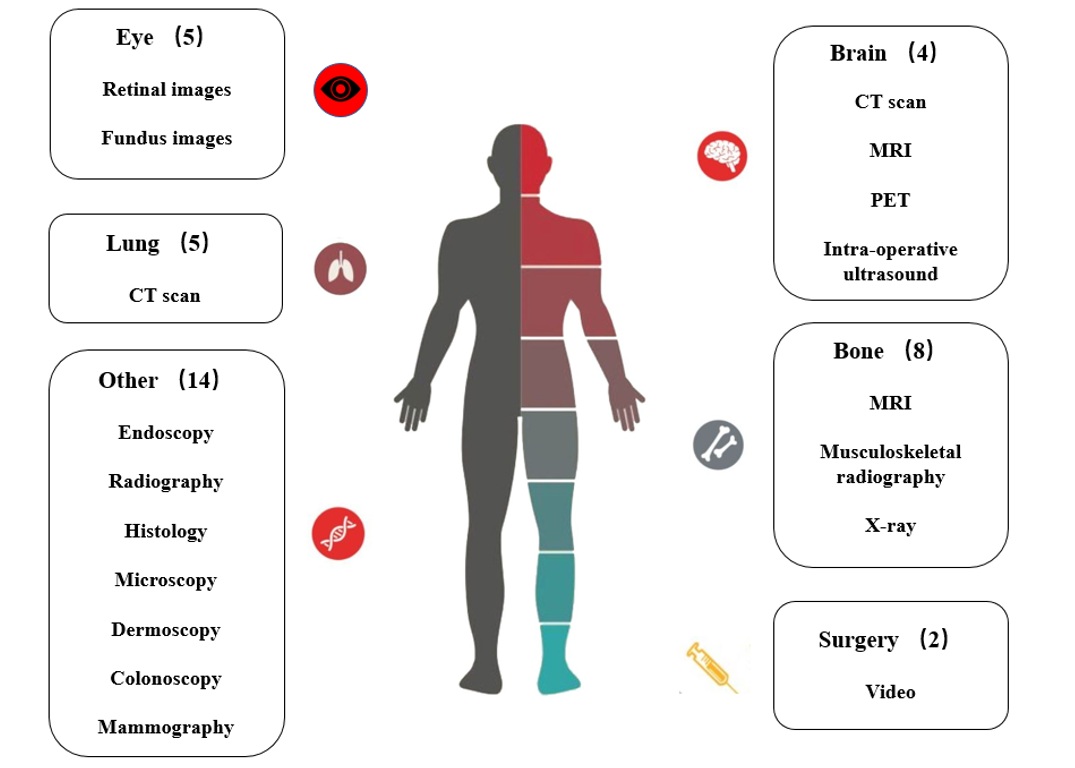}
\caption{Different types of medical images included in datasets. The number of available datasets related to each organ is shown in parentheses.}
\end{figure}

\begin{table*}[!h]
  \centering

  \resizebox{\textwidth}{!}{
    \begin{tabular}{|l| c | p{3.5 cm} | p{3 cm}| p{2 cm} | p{5 cm} |}
    \hline
    Challenge/Benchmark & Year &  Task &  Dataset & Metric & Task(s) \\ \hline
    CVPR 2018 - Video Segmentation Challenge \citep{Kaggle2018} & 2018  & Video Segmentation & - & mAP \& IoU & Segmentation of movable object from video frames \\ \hline
     CVPR 2018 - Berkeley DeepDrive challenges \citep{BerkeleyDeepDrive2018} &  2018 & Road Object Detection \& Drivable Area Segmentation \& Domain adaptation &  BDD 100K dataset  & AP \& IoU & Object detection, segmentation and tracking \\ \hline
    nuScenecs 3D detection challenge \citep{Caesar2019} & 2019  & 3D model generation & nuScenes dataset & mAP \& TP & Object detection with attributes  using 3D bounding boxes \\ \hline
    Lyft 3D Detection for Autonomous Vehicle \citep{Lyft2019} & 2019  & Object detection & Lyft Level 5 dataset & IoU   &  Object detection with 3D bounding boxes over semantic maps. \\ \hline
    NightOwls Pedestrian Detection Challenge \citep{Neumann2019} & 2019  & Pedestrian detection & NightOwls dataset & Standard average missing rate & Pedestrian detection at night tim \\ \hline
    D²-City Detection Domain Adaptation Challenge \citep{DiDi2019} & 2019  & Object detection \& Domain adaptation &  Image-Net \& BDD 100K datasets & AP \& IoU &  Object detection in diverse conditions with a focus on transfer learning to new unseen environments  \\ \hline
    WIDER Face \& Person Challenge & 2019  & Pedestrian detection & WIDER dataset & mAP \& IoU &  Detection of pedestrians and cyclist in unconstrained environment. \\ \hline
    CVPR 2019 - Beyond Single-frame Perception \citep{Apolloscape} & 2019  & 3D object detection & - & mAP \& IoU & Object detection on Lidar data \\ \hline
    The KITTI 3D Object Evaluation Benchmark \citep{Geiger} & 2017  & Object detection & KITTI dataset & precision-recall curve \& AP & Objection detection using 3D bounding boxes \\ \hline
    Caltech Pedestrian Detection Benchmark \citep{Dollar2010} & 2012  & Pedestrian detection & Caltech Pedestrian dataset & IoU   & Pedestrian detection \\ \hline The KITTI 2D Object Evaluation Benchmark \citep{Geigera} & 2012 &  Object detection \& Object Orientation & KITTI dataset & precision-recall curve \& AP \&  average orientation similarity &  Objection detection and orientation estimation \\ \hline
    \end{tabular}}
      \caption{AV-related object recognition and scene understanding challenges.}
\end{table*}

\begin{table*}[!t]
  \centering
  \tiny
  \resizebox{\textwidth}{!}{
    \begin{tabular}{|p{3.5 cm}| c | c | l | p{2 cm} | l | p{5 cm}|}
    \hline
Dataset & Size & Year &  Targeted disease/organ & Content & Challenge & Description \\ \hline
NLM's MedPix Dataset \citep{UniformedServicesUniversity2006} & 59000 images & - & - & Integrated images & no & A free online dataset contains   more than 12000 patient cases \\ \hline
STARE Database \citep{Goldbaum1975} & 400 cases & - & Eye & retinal images & no & Blood vessel segmentation images \\ \hline
SMIR \citep{Sicas} & 350425 images & - & - & CT scans & yes & 51 subjects of   whole-body postmortem CT scans \\ \hline
EchoNet-Dynamic \citep{David2019} & 10030 images & 2020 & Heart & Echocardiographic   video frames & yes & An expert labeled   dataset for the study of cardiac motion and chamber size. \\ \hline
Atlas of Digital Pathology \citep{Hosseini2019} & 17668   images & 2020 & Radiological diagnosis & Histological   patch images & yes & Images of   different organs with 57 types of hierarchical tissue annotated \\ \hline
COVID-CT Dataset \citep{Zhan2019} & 349 images & 2020 & COVID19 & CT scans & no & - \\ \hline
SARAS-ESAD Dataset \citep{SARAS2020} & 22601 frames & 2020 & Prostatectomy   procedure & Video frames & yes & A dataset of videos   showing the full prostatectomy procedure by surgeons \\ \hline 
The StructSeg 2019 Dataset{\citep{Li2020}} & 120 cases & 2019 & Radiotherapy   planning & CT scans & yes & A dataset for the   treatment of cancers \\ \hline
ODIR-5K \citep{PekingUniversity2019a} & 5000 images & 2019 & Eye & fundus photographs & yes & Fundus images taken by various cameras with different resolutions \\ \hline
DRIVE \citep{Staal2013} & 400 cases & 2019 & Eye & Rential images & yes & Images of 400   different patients between 25-90 years of age. \\ \hline
The RSNA Brain Hemorrhage CT Dataset \citep{Flanders} & 874035 images & 2019 & Brain Hemorrhage & CT scans & yes & Images gathered from 2 medical societies and 60 neuroradiologists \\ \hline
The KiTs19 Challenge Dataset \citep{Heller2019} & 300 cases & 2019 & Kidney tumor & CT scans & yes & A dataset of   multi-phase CT imaging with segmentation masks \\ \hline
SegTHOR \citep{Lambert2019} & 60 scans & 2019 & Lung & CT scans & No & A dataset focused   on the segmentation of organs at risk in the thorax \\ \hline
The EAD Challenge Dataset \citep{Ali2019a} & 2700 images & 2019 & Hollow organs & Endoscopic video   frames & yes & Images collected from 6 different data centers \\ \hline
Oasis Brains Dataset \citep{LaMontagne2019} & $\sim$1000 cases & 2019 & Brain & MRI \& PET   images & no & A dataset collected over 30 years \\ \hline
CheXpert \citep{Irvin2019} & 224316 images & 2019 & Chest & Chest radiographs & yes & A dataset labeled by an automatic labeler \\ \hline
LERA \citep{StanfordUniversity} & 182 patients & 2019 & Musculoskeletal   disorder & Radiographs & yes & Images of hip, foot, ankle and knee of patients for the study of   musculoskeletal disorders \\ \hline
CAMEL colorectal adenoma Dataset \citep{Xu2019} & 177 cases & 2019 & Cancer & Histology images & no & A dataset for   segmentation of cancerous parts in organ \\ \hline
BACH Dataset \citep{Aresta2019} & 430 images & 2019 & Breast cancer & Microscopy \&   whole-slide images & yes & Microscopy images labelled by 2 experts \\ \hline
MRNet \citep{Bien2018} & 1370 patients & 2018 & Knee & MRI & yes & A dataset for   autonomous MRI diagnosis \\ \hline
The REFUGE Challenge Dataset \citep{Fu} & 1200   images & 2018 & Glaucoma & Fundus   photographs & yes & The dataset   was collected using two types of devices. \\ \hline
MURA \citep{Rajpurkar2017} & 40561 images & 2018 & Bone & musculoskeletal   radiographs & yes & A manually labeled dataset by board-certificated Stanford radiologists,  containing 7 body types: finger, hand, elbow, forearm, humerus, wrist and   shoulder \\ \hline
Calgary-Campinas Public Brain MR Dataset \citep{Souza2018} & 167 scans & 2018 & Brain & MRI & no & A dataset for   analysis of brain MRI \\ \hline
HAM 10000 Dataset \citep{Tschandl2018, Codella2019} & 10015 images & 2018 & Skin lesions & Dermatoscopic   images & yes & A multi-modal and multi-population dataset \\ \hline
NIH Chest X-ray Dataset \citep{Wang2017a} & 100000 images & 2017 & Chest & X-ray images & no & A dataset of x-ray   images \\ \hline
609 Spinal Anterior-posterior X-ray Dataset\citep{Wu2017} & 609 images & 2017 & Spine & X-ray images & No & Each vertebra was   located by a landmark and the landmark is used to calculate Cobb angles. \\ \hline
RESECT \citep{Xiao2017} & 23 patients & 2017 & Cerebral Tumor & MRI \&   intra-operative ultrasound & yes & A dataset of   homologous landmarks \\ \hline
Cancer Digital Slide Archive \citep{WinshipCancerInstitute} & - & 2017 & Cancers & Glass slides of   histologic images & no & High resolution   detailed images of tissue microenvironments and cytologic details \\ \hline
Cholec80 \citep{Twinanda2017} & 80 videos & 2016 & Surgery & Video frames & no & A dataset containing 80 videos of surgeries performed by 13 different   surgeons \\ \hline
CSI 2014 Vertebra Segmentation Challenge Dataset \citep{Yao2016} & 10 scans & 2016 & Spine & CT scan & yes & Entire thoracic and lumbar spine were covered by the images. The   in-plane resolution is from 0.31 to 0.45mm. The slice-thickness is 1mm or   2mm. \\ \hline
CRCHistoPhenotypes - Labeled Cell Nuclei   Data\citep{Sirinukunwattana2016} & 100 images & 2016 & Cell & Histology images & no & 100 H\&E stained histology images of colorectal adenocarcinomas \\ \hline
Multi-Modality Vertebra Dataset \citep{Cai2015} & 20 cases & 2015 & Vertebra & MRI \& CT scan & no & The 3D vertebra centre location and orientation are annotated. \\ \hline
CVC colon DB \citep{Bernal2012} & 1200   frames & 2012 & colon   \& rectum & Colonoscopy   video frames & no & The dataset's region of interest has been annotated. The   video frames were specifically chosen for maximum visual distinction among   them. \\ \hline
LIDC-IDRI Database \citep{Armato2011} & 1018 cases & 2011 & Lung nodule & CT scans & yes & A database created   by 7 academic centers and 8 medical imaging companies \\ \hline
Computed Tomography Emphysema Dataset \citep{Sorensen2010} & 115 slices & 2010 & COPD & CT scans & no & High-resolution CT   scans \\ \hline
DIARETDB1 \citep{Kauppi2007} & 89 images & 2007 & Diabetic   retinopathy & fundus photographs & no & A database for   benchmarking the detection of diabetic retinography \\ \hline
ELCAP Public Lung Image Database \citep{SIMBA} & 50 sets & 2003 & Lung & CT scans & no & 50 low-dose documented CT scans for lungs containing nodules \\ \hline
The Digital Database for   Screening Mammography \citep{Heath,Heath2001} & 2620 cases & 1998 & Breast & Mammography images & no & The database has the function for user to search classes among   normal, benign and cancer. \\ \hline
    \end{tabular}}
      \caption{Medical imaging datasets.}
\end{table*}

\begin{table*}[!h]
  \centering
  \tiny
  \resizebox{\textwidth}{!}{
    \begin{tabular}{|p{2.5 cm}| c | l | l | p{2 cm} | p{5 cm}|}
    \hline
Challenge & Year & Task & Dataset & Metric & Targeted area \\ \hline
SARAS \citep{Cuzzolin2020} & 2020 & Detection & SARAS-ESAD Dataset & mAP & AI integrated minimally invasive surgery; Detection of surgeon's actions \\ \hline
REFUGE \citep{Fu} & 2020 & Detection Segmentation & REFUGE Challenge   Dataset & - & Automated   segmentation and detection of glaucoma \\ \hline
StructSeg \citep{Li2019a} &2019 & Segmentation& The   StructSeg 2019 Dataset & DSC \&   95\% Hausdoff Distance & Lung cancer and nasopharynx cancer;Gross   target volume and organs at risk \\ \hline
DRIVE \citep{Staal2013} & 2019 & Segmentation & DRIVE & Overall   prediction accuracy and s score & Screening programs for diabetic retinopathy;  Diagnosis   of hypertension and computer-assisted laser surgery \\ \hline
ODIR \citep{PekingUniversity2019a} & 2019 & Classification & ODIR-5K & Precision, accuracy   and dice similarity & AI in retinal image analysis \\ \hline
RSNA Intracranial Hemorrhage Detection \citep{RadiologicalSocietyofNorthAmerica2019} & 2019 & Detection & The RSNA Brain   Hemorrhage CT Dataset & Weighted   multi-label log loss & Detection of acute   intracranial hemorrhage and respective subtypes \\ \hline
KiTS \citep{UniversityofMinnesota2019} & 2019 & Segmentation & The KiTs19   Challenge Dataset & FROC & Kidney tumor   semantic segmentation \\ \hline
EAD \citep{Ali2019} & 2019 & Classification Detection Segmentation & The EAD Challenge   Dataset & average Dice   coefficient & Diagnosis and   treatment for diseases in hollow organs \\ \hline
AASCE \citep{Li2019} & 2019 & Regression & AASCE Challenge   Dataset & Symmetric mean   absolute percentage error & Automated spinal   curvature estimation and correction of error from x-ray images \\ \hline
CuRIOUS \citep{Reinertsen2019} & 2019 & Registration & RESECT & Threshold Jaccard   Index and normalized multi-class accuracy & Employment of AI in   surgery \\ \hline
ISIC \citep{Canfield} & 2018 & Classification & The HAM 10000 & mAP, IoU, Dice   coefficient, Jaccard Index, F2 score and deviation score. & Automated diagnosis   of melanoma \\ \hline
Data Science Bowl \citep{Caicedo2019} & 2018 & Classification & - & IoU & Detection of   nucleus \\ \hline
ICIAR \citep{Araujo2018} & 2018 & Classification Segmentation & BACH & Mean target   registration errors & Early diagnosis of   breast cancer \\ \hline
LUNA \citep{Jacobs2016} & 2016 & Classification Detection & LIDC-IDRI & Kappa score, \(F_1\)   score and AUC & Automatic detection   of lung nodule algorithms \\ \hline
SARAS \citep{Cuzzolin2020} & 2015 & Segmentation & - & Average Dice coefficient & Detection of dental caries \\
\hline
    \end{tabular}}
      \caption{Medical imaging challenges and benchmarks.}
\end{table*}

\subsection{Face Recognition}

Face Recognition (FR) is one of the highly investigated object recognition tasks with datasets and algorithms dating back to the early 1990’s. The main two face recognition tasks are face verification and face identification. The former determines if a pair of images correspond to the same human subject (e.g., used in access control systems). This task has been widely studied with algorithms outperforming human recognition rates \citep{Kemelmacher-Shlizerman2016} on the benchmark datasets such as LFW \citep{Huang2008}. For face identification, given two gallery and probe face image sets, the task is to rank the images in the gallery set based on their similarity to each image in the probe set. Forensic identification, watchlist identification, and de-duplication (e.g., mug shut repositories, ID card databases) are among the applications of face identification paradigm \citep{Klare2015}.\\
In the past three decades, there has been a steady expansion of the FR datasets as evolving algorithms reached near-perfect accuracy levels on previous datasets. The early face recognition datasets offered laboratory-controlled samples with limited scale, lighting condition, and background clutter. FERET \citep{Phillips1998}, CMU-PIE \citep{Sim2002}, and the AR Face database \citep{Martinez1998} are among the prominent datasets of this era. A complete list of early milestone datasets is provided in \citep{Abate2007}. After the introduction of shallow learning methods, which used hand-engineered descriptors such as SIFT and HOG, these datasets became saturated, and a new era of datasets containing unconstrained images emerged to challenge the state-of-the-art. LFW was among the first datasets to offer unrestricted images in the wild, varying in illumination levels, subject pose, and facial expression. Due to its increased difficulty compared to the laboratory-controlled datasets, LFW remained the main dataset for benchmarking FR algorithms for several years. \citep{Wolf2011,Kumar2009, Berg2004, Everingham2006} are among other datasets in this era. The third generation of FR datasets were increased in scale to fuel the data-hungry deep FR algorithms. Similar to other object recognition paradigms, it was shown that the accuracy of deep FR networks benefits from the availability of large-scale training data \citep{Zhou2015}. The early large-scale datasets were private and hence could neither be used as fair performance benchmarks nor as training sets for deep models developed in the academia. Facebook’s DeepFace \citep{Taigman}, Google’s FaceNet \citep{Schroff}, and the DeepID series \citep{Sun2014,Sun, Sun2015}, are the major models trained on private datasets of 4M, 500M, and 0.2M images respectively. Such models reached near perfect accuracies on benchmarks such as LFW and YTF, hence calling for more challenging benchmarks. To this end, public datasets were gradually introduced to both enable large-scale training of academic deep models (e.g., CASIA-WebFace \citep{Yi2014}, MS-Celeb-1M \citep{Guo2016}, VGGFace2 \citep{Cao2018})  and also to create a fair performance evaluation medium (e.g. \citep{Klare2015, Whitelam2017,Maze2018} and MegaFace \citep{Kemelmacher-Shlizerman2016,Nech} benchmarks). \\
FR datasets expand either in breadth or depth. Broad datasets are focused on increasing the number of subjects and are suitable for low-shot learning purposes \citep{Nech, Rothe, Guo2016}, while deep datasets increase their size by containing more images per subject \citep{Cao2018, Chenio,Parkhi2015}, therefore allowing for intra-class variations resulted from age, pose, and expression variations. Some datasets also offer additional annotations such as facial keypoints, age, pose, craniofacial features, and binary attributes. The prominent datasets from the second and third eras are listed in Table 14. \\
Most FR datasets are collected using internet sources and consist of celebrity faces in formal occasions. This limits the studied demographic. Data bias exists with regards to gender, race and age. It has been shown that commercial and state-of-the-art algorithms perform best on Caucasians while having the most error rates on African faces for the verification task \citep{Wang}.  To this end, there have been recent efforts to correct the recognition bias by decreasing class imbalance \citep{Wang,Karkkainen} and including diversity-based features \citep{Merler2019}.

\begin{table*}[!h]
  \centering
  \tiny
   \resizebox{\textwidth}{!}{
    \begin{tabular}{|l| l | p{1 cm} | l | p{1.5 cm} | p{7 cm}|}
    \hline
    Dataset & Year & Number of Subjects & Number of Images & Additional Information & Highlights \\ \hline
VGGFace \citep{Parkhi2015} & 2015 & 2,622 & 2.6 M & A & Large-scale celebrity recognition with high intra-class variations \\ \hline
VGGFace2 \citep{Cao2018} & 2018 & 9,131 & 3.31M & A, P & Diversified pose, age, and ethnicity of celebrity faces \\ \hline
LFW \citep{Huang2008} & 2007 & 5,749 & 13,233 & - & The first unconstrained FR dataset \\ \hline
MegaFace \citep{Nech} & 2016 & 672,052 & 4.7 M & - & Raised difficulty by including 1 M  distractors; non-celebrity subjects \\ \hline
YTF \citep{Wolf2011} & 2011 & 1,595 & 3,425 V & - & Designed for face verification in videos; same format as LFW \\ \hline
CASIA-WebFace \citep{Yi2014} & 2014 & 10,577 & 494,414 & - & First publicly available large-scale FR dataset \\ \hline
IJB-A \citep{Klare2015} & 2015 & 500 & 5,712 & BB, KP & Manually verified bounding boxes for face detection, nose and eye   keypoints included \\ \hline
MS-Celeb-1M \citep{Guo2016} & 2016 & 100,000 & 10 M & - & Celebrity identification dataset and benchmark with a linked celebrity   knowledge base \\ \hline
Pubfig \citep{Kumar2009} & 2009 & 200 & 60,000 & A, E, G, P, BB & 73 automatically generated attributes provided; same format as LFW \\ \hline
CelebA \citep{Liu2015a} & 2015 & 10,177 & 202,599 & KP & Designed for face attribute prediction in the wild; 40 binary attributes   included \\ \hline
DiF \citep{Merler2019} & 2019 & - & 0.97 M & A, P, S, BB, KP & Quantitative facial features included to reduce recognition bias across   different demographics \\ \hline
IMDB-Face \citep{Rothe2018} & 2015 & 100,000 &460,723 & A, G & Age and gender prediction on a set of celebrities collected   from IMDB \\ \hline
UMDFaces \citep{Bansal2016} & 2016 & 8,501 & 367,920 & A, P, G, BB, KP & Detailed human-verified attributes and annotations \\ \hline
IJB-B \citep{Whitelam2017} & 2015 & 1,845 & 21,798 & A, G, P, S & A superset of IJB-A with additional occlusion and illumination   annotations \\ \hline
IJB-C \citep{Maze2018} & 2018 & 3,531 & 31,334 & A, G, P, S & An improvement upon IJB-B with a focus on diversifying the geographic   coverage of subjects \\ \hline
FaceScrub \citep{Hong-Wei2014} & 2014 & 695 & 141,130 & G & A broad dataset of movie celebrities gathered from IMDB \\ \hline
CACD\citep{Chenio} & 2014 & 2,000 & 163,446 & A & Images include age variations for each subject for cross-age face   recognition and retrieval; only 200 subjects are manually annotated \\
\hline
    \end{tabular}}
      \caption{Well-known face recognition datasets. Abbreviations in the table: Oclusion (O), Pose (P), Age (A), Expression (E), Skin color (S), Gender (G), Bounding Boxes (BB), Keypoints (KP), V (video).}
\end{table*}

Since face verification and face identification are two different matching problems (one-to-one versus one-to-many), different evaluation metrics are used to assess the algorithms in these paradigms. Face verification algorithms are usually evaluated using the ROC curve. For a given test set of image pairs, a ROC curve can be drawn by changing the threshold above which the output confidence score is considered a matched pair. Each point on the ROC curve corresponds to a True Positive Rate (TPR) on the vertical and a False Positive Rate (FPR) on the horizontal axis for a threshold \textit{t} (the independent variable). The TPR is the fraction of detected face matches that correctly exceed \textit{t}, and FPR is the fraction of detected face matches that falsely exceed the threshold. The area under the ROC curve (AUC) or the TPR value at a fixed FPR can then be calculated as a measure of the verification accuracy. Another face verification metric is LFW’s \textit{estimated mean accuracy} \citep{Huang2008}, which is the percentage of correct binary classifications. The Cumulative Match Characteristic (CMC) curve is the most common evaluation metric used for face identification. Given a set of probe images and a gallery, the CMC curve can be drawn by ranking the images in the gallery according to predicted similarity scores for each of the probes. For each point on the CMC curve, the vertical axis value corresponds to the percentage of the probe images for which the algorithm has found the corresponding gallery mate within the top-\textit{k} ranked predictions, and the horizontal axis value represents the rank \textit{k}, varied between one and the total number of images in the gallery. The MS-Celeb-1M is the only major face identification benchmark that uses a precision-coverage curve instead of the CMC curve \citep{Guo2016}. A list of metrics used in common FR benchmarks is provided in \citep{Masi2019}.
\subsection{Remote Sensing}
The increased access to aerial images captured by satellites or UAVs has paved the way for a wide series of applications such as traffic flow management, precision agriculture, natural disaster response, and map extraction. In order to achieve a scaled utility, such applications call for automated detection of map features such as urban infrastructure, vegetation, and other points of interest. \\
Remote sensing datasets usually are limited in size and the number of covered classes compared to object recognition datasets in ground-level view. Vehicles, roads, and buildings are among the highly investigated object classes due to their significance in a wide range of applications such as traffic management, surveillance, urban planning and modeling population dynamics \citep{Demir, Mundhenk2016}. In recent years, a number of general-purpose object recognition datasets have also emerged, covering more classes \citep{Cheng2017,Zou2018,Zhu2015} with a few exceptions such as xView \citep{Lam2018} and fMoW \citep{Christie2017}, these datasets still lack the scale required for training deep models from scratch. An overview of some of the popular remote sensing datasets is provided in Table 15. \\
While many of the datasets are scraped from Google Earth (e.g., NWPU \citep{Cheng2017}, DOTA \citep{Xia2017} , TAS \citep{Heitz}) and thus lack multispectral data and temporal views, some datasets utilize 4-band or 8-band images to include a wider range of electromagnetic spectrum, especially the near infrared channel \citep{Razakarivony2016}. \\ 
The dominant choice of annotation for remote sensing datasets is bounding boxes. Since objects appear in different orientations when looked from above, remote sensing images contain high levels of intra-class aspect ratio variations. Thus, many of the remote sensing object detection datasets offer rotated bounding box annotations to alleviate this problem \citep{Xia2017, Razakarivony2016}. Several datasets have also been proposed for semantic segmentation purposes, e.g., the TorontoCity \citep{Wang2016}, ISPRS 2D semantic labeling \citep{Sensing2017}, DeepGlobe+Vivid \citep{Demir} datasets. A number of competitions have also been held to promote research in aerial object recognition. A list of important remote sensing challenges is provided in Table 16.

\begin{table*}[!h]
\centering
\scriptsize
\resizebox{\textwidth}{!}{%
\begin{tabular}{|l|l|l|l|l| p{6 cm}|}
\hline
Dataset & Year & Annotation & Size & Spatial Resolution (cm per pixel) & Description \\ \hline
SpaceNet C.1\&C.2 & 2019 &685,000 buildings & 5,555 \(km^2\) & 30-50 & Building footprints annotated using   polygons; five cities \\ \hline
SpaceNet C.3 & 2019 &8,676 km road & 5,555 \(km^2\) &30-50 & Road centerlines labeled based on the OpenStreetMap scheme \\ \hline
 
COWC \citep{Mundhenk2016} & 2016 & 32,716 vehicles & - & 15 & Car detection dataset gathered from six cities in North America and   Europe; cars annotated with points on centroids \\ \hline
xView \citep{Lam2018} & 2018 & 1M   objects & 1,400 \(km^2\) & 30 & Large-scale object overhead object detection dataset with   bounding box annotations \\ \hline
FMoW \citep{Christie2017} & 2017 & 132,700 objects & 1M & - & Temporal image sequences from over 200 countries with the purpose visual   reasoning about location, time, and sun angles; bounding box annotations \\ \hline
NWPU-RESISC45 \citep{Cheng2017} & 2017 & 31,500 scenes & 31,500 & 20-3000 & Aerial scene classification dataset with variations in spatial   resolution, illumination, object pose, occlusion \\ \hline
TorontoCity \citep{Wang2016} &2016 & 400,000   buildings &712 \(km^2\) &10 &RGB and LiDAR Aerial imagery of the greater Toronto area   augmented with and street-view stereo and LiDAR \\ \hline
DOTA \citep{Xia2017} & 2018 & 188,282 objects & 2,806 & - & Rotated bounding box annotations verified by expert annotators; 15   common classes \\ \hline
TAS \citep{Heitz} & 2008 & 1,319 vehicles & 30 & - & An early annotated remote sensing dataset from collected from google   earth; bounding box annotations \\ \hline
DLR3K \citep{Liu2015} & 2013 & 3,472 vehicles & 20 & 13 & Rotated bounding boxes with additional orientation annotations \\ \hline
NWPU VHR-10 \citep{Cheng2016} & 2016 & 3,775 objects & 715 & 50-200 & generic object detection dataset with 10 classes; bounding box   annotations \\ \hline
LEVIR \citep{Zou2018} & 2018 & 11,000 objects & 22,000 & 20-100 & Bounding box annotations provided for airplanes, ships, and oilpots \\ \hline
VEDAI \citep{Razakarivony2016} & 2016 & 3,600 vehicles & 1,210 & 12.5 & Small vehicle detection consisting of 9 vehicle classes; rotated   bounding box annotations \\ \hline
UCAS-AOD \citep{Zhu2015} & 2015 & 6,000 objects & 910 & - & Rotated bounding box annotations; vehicle and airplane detection;   scraped from Google Earth \\ \hline
AID\citep{Xia2016} & 2016 & 10,000 scenes & 10,000 & 50-800 & Aerial scene classification with 30 classes \\ \hline
\end{tabular}
}
\caption{ Remote sensing object detection datasets. Dataset size is the number of images unless states otherwise.}
\end{table*}

\begin{table*}[!h]
\centering
\scriptsize
\resizebox{\textwidth}{!}{%
\begin{tabular}{|l|l|l |l | l |p{5 cm}|}
\hline
Challenge & Year & Dataset size & Number of Classes & Evaluation Metric & Task \\ \hline
SpaceNet C.1\&C.2\citep{Etten} & 2019 & 5,555 \(km^2\) &2 & \(F_1\) score & Building footprint detection in 5 cities \\ \hline
SpaceNet C.3\citep{Etten} &2019 &5,555 \(km^2\) & 1 & Average Path Length Similarity &Road network extraction \\ \hline
FMoW \citep{Christie2017} & 2017 & 1 M & 63 & \(F_1\) score & Object classification \\ \hline
DSTL \citep{dstl} & 2017 & 57 & 10 & IoU & Semantic segmentation \\ \hline
NWPU- RESISC45 \citep{Cheng2017} & 2017 & 31,500 & 45 & Accuracy & Scene classification \\ \hline
DIUx xView \citep{Lam2018} & 2018 & 1,400 \(km^2\) & 60 & IoU &  Object detection \\ \hline
DeepGlobe \citep{Demir} & 2018 & 10,000 &7 * & IoU, \(F_1\) score & Building segmentation, road extraction; land cover   classification \\ \hline
\end{tabular}
}
\caption{Remote sensing challenges. * the number of classes for the land cover classification task.}
\end{table*}

\subsection{Species Recognition}
Numerous fine-grained datasets have been created for species recognition tasks. Although there are many images of various pet animal categories available online, datasets including other categories e.g., wild animals are very sparse. Therefore, many efforts have been made to collect images of diverse species in their natural habitats \citep{Horn2018,Swanson2015,Goeau2019}. Another challenge with regards to fine-grained species recognition is the required expertise knowledge for fine-grained classification annotations, which makes crowdsourcing less practical. However, crowdsourcing workers have also been employed to help with bounding box or keypoint annotations \citep{Khan2019,Everingham2014}. The popular species recognition datasets are described in Table 17. Species recognition competitions are also held for some of these datasets. iNaturalist is an annual Kaggle-hosted competition offering a classification task of a subset of the main dataset and is evaluated using a combination of ILSVRC’s top-k classification evaluation criteria. LifeCLEF is another annual competition with a plant identification task that is based on the PLANTCLEF dataset and is evaluated by the top-1 accuracy criterion.

\begin{table*}[!h]
\centering
\scriptsize
\resizebox{\textwidth}{!}{%
\begin{tabular}{|l|l|l|l|l|l| p{5 cm}|}
\hline
Dataset & Number of Images & Number of Classes & Number of Annotations & Year & Challenge & Description \\ \hline
Flower 102 \citep{Nilsback2008} & 8,189 & 103 & 8,189 SM & 2008 & No & Flower recognition dataset of 103 flower categories common in the United   Kingdom \\ \hline
Caltech-Birds 2011\citep{wah2011caltech} & 11,788 & 200 & 11,788 BB & 2011 & No & 15 part locations and 28 attributes for each bird \\ \hline
Stanford Dogs \citep{Khosla2011} & 22,000 & 120 & 22,000 BB & 2011 & No & Single-object per image dataset for dog breed recognition \\ \hline
F4K \citep{Boom2012} & 27,370 & 23 & 27,370 CL & 2012 & No & Fish recognition dataset annotated by following marine biologists \\ \hline
Snapshot Serengeti \citep{Swanson2015} & 1.2 m & 61 & 406,433 CL, 150,000 BB & 2014 & No & Wild animal classification dataset gathered using 225 camera-traps in   Serengeti national park in Africa \\ \hline
NABirds \citep{Horn} & 48,562 & 555 & 48,562 BB & 2015 & No & Expert-curated dataset of North American birds; 11 bird parts annotated   in every image \\ \hline
PlantCLEF \citep{Goeau2019} & 434,251 & 10,000 & 10,000 CL & 2015 & Yes & Plant classification dataset gathered in the Amazon   rainforest \\ \hline
iNat \citep{Horn2018} & 675,175 & 5,089 & 561,767 BB & 2017 & Yes & Manually collected dataset of 13 super-class and 5k sub-class species;   organized in a hierarchical taxonomy; highly imbalanced \\ \hline
Dogs-in-the-Wild \citep{Sun2018a} & 300,000 & 362 & 300,000 CL & 2018 & No & A large dataset for dog breed classification in natural environments \\ \hline
AnimalWeb \citep{Khan2019} & 21,900 & 334 & 198k KP & 2019 & No & Hierarchically categorized dataset for animal face recognition with 9   keypoint annotations per face \\ \hline
IP102\citep{Wu2019} & 75,000 & 102 & 75,000 CL, 19,000 BB & 2019 & No & Hierarchically categorized dataset for insect pest   recognition \\ \hline
\end{tabular}
}
\caption{Species recognition datasets}
\end{table*}

\subsection{Clothing Detection}
Fashion is another milieu that has attracted a lot of attention in the computer vision community. There are two vision tasks that are most popular: 1) retrieval of clothing images similar to an item in a given image, in the same manner as the recommendation systems used in cloths shopping websites, and 2) parsing individual fashion items where the goal is to identify the exact model of a clothing item in an image taken by the customer and retrieve the same product or similar products from the shopping floor \citep{Zheng2018}. To this end, the created datasets usually offer multiple clothing attributes for each item in the form of binary tags (e.g., has v-shaped collar, is red), and some of them also offer clothing pairs, one from an item in the shopping floor and the other taken by a customer in an unconstrained condition to facilitate the retrieval procedure. Although mere recognition is usually not the main purpose of clothing datasets, some of the clothing datasets offer fine-grained detection/segmentation annotations, which can be used for accurate recognition purposes. Clothing recognition remains a challenging task mainly due to the non-rigid structure of cloths that cause unpredictable deformations in various body poses. Other challenges include occlusions, sensitivity to lighting conditions, and inconsistencies between commercial and customer images \citep{Ge2019}. The prominent clothing datasets with annotated instances are discussed in Table 18. 
\begin{table*}[!h]
\centering
\scriptsize
\resizebox{\textwidth}{!}{%
\begin{tabular}{|l|l|l|l|l|l|l|l|}
\hline
Dataset & Number of Images & Classes & Number of Annotated Clothing   instances & Annotation Type & Year & Challenge/Benchmark & Number of Attributes \\ \hline
DARN\citep{Huang2015a} & 182,000 & 20 & 182,000 & BB & 2015 & No & 9 \\ \hline
Street2Shop\citep{Kiapour2015} & 404,000 & 11 & 20,357 & BB & 2015 & No & - \\ \hline
DeepFashion\citep{Liu2016} & 800,000 & 50 & 180,000 & KP & 2016 & Yes & 5 \\ \hline
ModaNet\citep{Zheng2018} & 55,000 & 13 & 240,000 & BB, SM & 2018 & No & - \\ \hline
FashionAI\citep{Zou2019} & 324,000 & 41 & 324,000 & KP & 2018 & No & 68 \\ \hline
Deepfashion2 \citep{Ge2019} & 491,000 & 13 & 801,000 & BB, SM, KP & 2019 & Yes & 4 \\ \hline
\end{tabular}%
}
\caption{Clothing detection datasets}
\end{table*}
The “annotated clothing instances number” column in the table shows the number of cloths having annotations that could be used for detection purposes. As mentioned before, some datasets have unannotated images from shopping retailers that can only be used for image retrieval and not for detection. Deepfashion 2 is the richest annotated clothing dataset, which also comes with an evaluation benchmark. The benchmark tasks are cloth detection, landmark estimation, and segmentation, which are evaluated based on analogous tasks in MS COCO challenges. The cloth detection and segmentations tasks are evaluated using AP for bounding boxes and segmentation masks respectively. The landmark estimation task, which is the localization of clothing keypoints, is evaluated by AP for the annotated keypoints (same as MS COCO keypoint detection).

\section{Conclusion and Future Work}
In this paper, a detailed analysis on the milestone datasets in generic object recognition was provided. This review also focused on six popular applications namely autonomous driving, medical imaging, face recognition, remote sensing, species detection, and clothing detection. With a focus on the most recent datasets generated in the deep learning era, details such as dataset size, annotation type, and individual descriptions are mentioned, and the challenges faced by the community in data collection, annotation, and algorithm design are also discussed. Our concentration on the study of the publicly available datasets and cataloguing the researchers’ efforts involved in generation and optimizing them were in view of the fact that the deep learning community generally believes that the most critical upcoming challenge in way of improving the utility of deep learning and other data-driven methods involves introducing and augmenting comprehensive and flawless training datasets. \\
In addition to the datasets, the major generic and application-specific object recognition competitions were investigated, with information on the adopted evaluation metrics, covered tasks, and dataset statistics. A comprehensive overview of the evaluation criteria frequently used in the competitions and benchmarks is also provided to more clarify the evaluation processes used for assessing the performance of different tasks. This will provide the curious reader with clues about the reasons for success of the approaches taken and an insight to where to begin in their own projects. \\
In conclusion, this review is an effort to help researchers shorten the process of finding the appropriate training and testing mediums for their desired applications and to shed light upon promising data collection directions as the state-of-the-are models inevitably saturate the existing datasets.

\section*{Acknowledgements}
We would like to acknowledge the financial support received by Aria Salari for this survey from Mathematics of Information Technology and Complex Systems (MITACS) under the Accelerate Program Internship no. IT16412 and Vancouver Computer Vision.

\bibliographystyle{Object_Recognition_Datasets_and_Challenges_-_A_Review}
\bibliography{Object_Recognition_Datasets_and_Challenges_-_A_Review}

\begin{thebibliography}{280}
\expandafter\ifx\csname natexlab\endcsname\relax\def\natexlab#1{#1}\fi
\providecommand{\url}[1]{\texttt{#1}}
\providecommand{\href}[2]{#2}
\providecommand{\path}[1]{#1}
\providecommand{\DOIprefix}{doi:}
\providecommand{\ArXivprefix}{arXiv:}
\providecommand{\URLprefix}{URL: }
\providecommand{\Pubmedprefix}{pmid:}
\providecommand{\doi}[1]{\href{http://dx.doi.org/#1}{\path{#1}}}
\providecommand{\Pubmed}[1]{\href{pmid:#1}{\path{#1}}}
\providecommand{\bibinfo}[2]{#2}
\ifx\xfnm\relax \def\xfnm[#1]{\unskip,\space#1}\fi
\bibitem[{()}]{}
, .
\newblock \bibinfo{title}{{IEEE Xplore Full-Text PDF:}}.
\newblock \URLprefix
  \url{https://ieeexplore.ieee.org/stamp/stamp.jsp?tp={\&}arnumber=8241865}.
\bibitem[{Abate et~al.(2007)Abate, Nappi, Riccio and Sabatino}]{Abate2007}
\bibinfo{author}{Abate, A.F.}, \bibinfo{author}{Nappi, M.},
  \bibinfo{author}{Riccio, D.}, \bibinfo{author}{Sabatino, G.},
  \bibinfo{year}{2007}.
\newblock \bibinfo{title}{{2D and 3D face recognition: A survey}}.
\newblock \bibinfo{journal}{Pattern Recognition Letters} \bibinfo{volume}{28},
  \bibinfo{pages}{1885--1906}.
\newblock \DOIprefix\doi{10.1016/j.patrec.2006.12.018}.
\bibitem[{Achantay et~al.(2009)Achantay, Hemamiz, Estraday and
  Süsstrunky}]{ASD}
\bibinfo{author}{Achantay, R.}, \bibinfo{author}{Hemamiz, S.},
  \bibinfo{author}{Estraday, F.}, \bibinfo{author}{Süsstrunky, S.},
  \bibinfo{year}{2009}.
\newblock \bibinfo{title}{Frequency-tuned salient region detection}.
\newblock \bibinfo{journal}{2009 IEEE Computer Society Conference on Computer
  Vision and Pattern Recognition Workshops, CVPR Workshops 2009} ,
  \bibinfo{pages}{1597--1604}\DOIprefix\doi{10.1109/CVPRW.2009.5206596}.
\bibitem[{Ali et~al.(2019a)Ali, Zhou, Daul, Braden, Bailey, Realdon, East,
  Wagni{\`{e}}res, Loschenov, Grisan, Blondel and Rittscher}]{Ali2019a}
\bibinfo{author}{Ali, S.}, \bibinfo{author}{Zhou, F.}, \bibinfo{author}{Daul,
  C.}, \bibinfo{author}{Braden, B.}, \bibinfo{author}{Bailey, A.},
  \bibinfo{author}{Realdon, S.}, \bibinfo{author}{East, J.},
  \bibinfo{author}{Wagni{\`{e}}res, G.}, \bibinfo{author}{Loschenov, V.},
  \bibinfo{author}{Grisan, E.}, \bibinfo{author}{Blondel, W.},
  \bibinfo{author}{Rittscher, J.}, \bibinfo{year}{2019}a.
\newblock \bibinfo{title}{{Endoscopy artifact detection (EAD 2019) challenge
  dataset}} , \bibinfo{pages}{1--13}\DOIprefix\doi{10.17632/C7FJBXCGJ9.1}.
\bibitem[{Ali et~al.(2019b)Ali, Zhou, Daul and Loschenov}]{Ali2019}
\bibinfo{author}{Ali, S.}, \bibinfo{author}{Zhou, F.}, \bibinfo{author}{Daul,
  C.}, \bibinfo{author}{Loschenov, M.}, \bibinfo{year}{2019}b.
\newblock \bibinfo{title}{{EAD 2019}}.
\newblock \URLprefix \url{https://ead2019.grand-challenge.org/}.
\bibitem[{Amisha et~al.(2019)Amisha, Malik, Pathania and Rathaur}]{Amisha2019}
\bibinfo{author}{Amisha}, \bibinfo{author}{Malik, P.},
  \bibinfo{author}{Pathania, M.}, \bibinfo{author}{Rathaur, V.K.},
  \bibinfo{year}{2019}.
\newblock \bibinfo{title}{{Overview of artificial intelligence in Medicine}}.
\newblock \bibinfo{journal}{Journal of Family Medicine and Primary Care}
  \bibinfo{volume}{8}, \bibinfo{pages}{2328--2331}.
\newblock \DOIprefix\doi{10.4103/jfmpc.jfmpc_440_19}.
\bibitem[{Apolloscape(2019)}]{Apolloscape}
\bibinfo{author}{Apolloscape}, \bibinfo{year}{2019}.
\newblock \bibinfo{title}{{CVPR 2019 WAD Beyond Single-frame Perception
  Challenge}}.
\newblock \URLprefix \url{http://wad.ai/2019/index.html}.
\bibitem[{Ara{\'{u}}jo et~al.(2018)Ara{\'{u}}jo, Aresta, Eloy, Ant{\'{o}}nio
  and Aguiar}]{Araujo2018}
\bibinfo{author}{Ara{\'{u}}jo, T.}, \bibinfo{author}{Aresta, G.},
  \bibinfo{author}{Eloy, C.}, \bibinfo{author}{Ant{\'{o}}nio, P.},
  \bibinfo{author}{Aguiar, P.}, \bibinfo{year}{2018}.
\newblock \bibinfo{title}{{ICIAR 2018}}.
\newblock \URLprefix \url{https://iciar2018-challenge.grand-challenge.org/}.
\bibitem[{Aresta et~al.(2019)Aresta, Ara{\'{u}}jo, Kwok, Chennamsetty, Safwan,
  Alex, Marami, Prastawa, Chan, Donovan, Fernandez, Zeineh, Kohl, Walz, Ludwig,
  Braunewell, Baust, Vu, To, Kim, Kwak, Galal, Sanchez-Freire, Brancati,
  Frucci, Riccio, Wang, Sun, Ma, Fang, Kone, Boulmane, Campilho, Eloy,
  Pol{\'{o}}nia and Aguiar}]{Aresta2019}
\bibinfo{author}{Aresta, G.}, \bibinfo{author}{Ara{\'{u}}jo, T.},
  \bibinfo{author}{Kwok, S.}, \bibinfo{author}{Chennamsetty, S.S.},
  \bibinfo{author}{Safwan, M.}, \bibinfo{author}{Alex, V.},
  \bibinfo{author}{Marami, B.}, \bibinfo{author}{Prastawa, M.},
  \bibinfo{author}{Chan, M.}, \bibinfo{author}{Donovan, M.},
  \bibinfo{author}{Fernandez, G.}, \bibinfo{author}{Zeineh, J.},
  \bibinfo{author}{Kohl, M.}, \bibinfo{author}{Walz, C.},
  \bibinfo{author}{Ludwig, F.}, \bibinfo{author}{Braunewell, S.},
  \bibinfo{author}{Baust, M.}, \bibinfo{author}{Vu, Q.D.}, \bibinfo{author}{To,
  M.N.N.}, \bibinfo{author}{Kim, E.}, \bibinfo{author}{Kwak, J.T.},
  \bibinfo{author}{Galal, S.}, \bibinfo{author}{Sanchez-Freire, V.},
  \bibinfo{author}{Brancati, N.}, \bibinfo{author}{Frucci, M.},
  \bibinfo{author}{Riccio, D.}, \bibinfo{author}{Wang, Y.},
  \bibinfo{author}{Sun, L.}, \bibinfo{author}{Ma, K.}, \bibinfo{author}{Fang,
  J.}, \bibinfo{author}{Kone, I.}, \bibinfo{author}{Boulmane, L.},
  \bibinfo{author}{Campilho, A.}, \bibinfo{author}{Eloy, C.},
  \bibinfo{author}{Pol{\'{o}}nia, A.}, \bibinfo{author}{Aguiar, P.},
  \bibinfo{year}{2019}.
\newblock \bibinfo{title}{{BACH: Grand challenge on breast cancer histology
  images}}.
\newblock \bibinfo{journal}{Medical Image Analysis} \bibinfo{volume}{56},
  \bibinfo{pages}{122--139}.
\newblock \DOIprefix\doi{10.1016/j.media.2019.05.010}.
\bibitem[{Armato et~al.(2011)Armato, McLennan, Bidaut, McNitt-Gray, Meyer,
  Reeves, Zhao, Aberle, Henschke, Hoffman, Kazerooni, MacMahon, {Van Beek},
  Yankelevitz, Biancardi, Bland, Brown, Engelmann, Laderach, Max, Pais, Qing,
  Roberts, Smith, Starkey, Batra, Caligiuri, Farooqi, Gladish, Jude, Munden,
  Petkovska, Quint, Schwartz, Sundaram, Dodd, Fenimore, Gur, Petrick, Freymann,
  Kirby, Hughes, {Vande Casteele}, Gupte, Sallam, Heath, Kuhn, Dharaiya, Burns,
  Fryd, Salganicoff, Anand, Shreter, Vastagh, Croft and Clarke}]{Armato2011}
\bibinfo{author}{Armato, S.G.}, \bibinfo{author}{McLennan, G.},
  \bibinfo{author}{Bidaut, L.}, \bibinfo{author}{McNitt-Gray, M.F.},
  \bibinfo{author}{Meyer, C.R.}, \bibinfo{author}{Reeves, A.P.},
  \bibinfo{author}{Zhao, B.}, \bibinfo{author}{Aberle, D.R.},
  \bibinfo{author}{Henschke, C.I.}, \bibinfo{author}{Hoffman, E.A.},
  \bibinfo{author}{Kazerooni, E.A.}, \bibinfo{author}{MacMahon, H.},
  \bibinfo{author}{{Van Beek}, E.J.}, \bibinfo{author}{Yankelevitz, D.},
  \bibinfo{author}{Biancardi, A.M.}, \bibinfo{author}{Bland, P.H.},
  \bibinfo{author}{Brown, M.S.}, \bibinfo{author}{Engelmann, R.M.},
  \bibinfo{author}{Laderach, G.E.}, \bibinfo{author}{Max, D.},
  \bibinfo{author}{Pais, R.C.}, \bibinfo{author}{Qing, D.P.},
  \bibinfo{author}{Roberts, R.Y.}, \bibinfo{author}{Smith, A.R.},
  \bibinfo{author}{Starkey, A.}, \bibinfo{author}{Batra, P.},
  \bibinfo{author}{Caligiuri, P.}, \bibinfo{author}{Farooqi, A.},
  \bibinfo{author}{Gladish, G.W.}, \bibinfo{author}{Jude, C.M.},
  \bibinfo{author}{Munden, R.F.}, \bibinfo{author}{Petkovska, I.},
  \bibinfo{author}{Quint, L.E.}, \bibinfo{author}{Schwartz, L.H.},
  \bibinfo{author}{Sundaram, B.}, \bibinfo{author}{Dodd, L.E.},
  \bibinfo{author}{Fenimore, C.}, \bibinfo{author}{Gur, D.},
  \bibinfo{author}{Petrick, N.}, \bibinfo{author}{Freymann, J.},
  \bibinfo{author}{Kirby, J.}, \bibinfo{author}{Hughes, B.},
  \bibinfo{author}{{Vande Casteele}, A.}, \bibinfo{author}{Gupte, S.},
  \bibinfo{author}{Sallam, M.}, \bibinfo{author}{Heath, M.D.},
  \bibinfo{author}{Kuhn, M.H.}, \bibinfo{author}{Dharaiya, E.},
  \bibinfo{author}{Burns, R.}, \bibinfo{author}{Fryd, D.S.},
  \bibinfo{author}{Salganicoff, M.}, \bibinfo{author}{Anand, V.},
  \bibinfo{author}{Shreter, U.}, \bibinfo{author}{Vastagh, S.},
  \bibinfo{author}{Croft, B.Y.}, \bibinfo{author}{Clarke, L.P.},
  \bibinfo{year}{2011}.
\newblock \bibinfo{title}{{The Lung Image Database Consortium (LIDC) and Image
  Database Resource Initiative (IDRI): A completed reference database of lung
  nodules on CT scans}}.
\newblock \bibinfo{journal}{Medical Physics} \bibinfo{volume}{38},
  \bibinfo{pages}{915--931}.
\newblock \DOIprefix\doi{10.1118/1.3528204}.
\bibitem[{Bansal et~al.(2016)Bansal, Nanduri, Castillo, Ranjan and
  Chellappa}]{Bansal2016}
\bibinfo{author}{Bansal, A.}, \bibinfo{author}{Nanduri, A.},
  \bibinfo{author}{Castillo, C.}, \bibinfo{author}{Ranjan, R.},
  \bibinfo{author}{Chellappa, R.}, \bibinfo{year}{2016}.
\newblock \bibinfo{title}{{UMDFaces: An Annotated Face Dataset for Training
  Deep Networks}}.
\newblock \bibinfo{journal}{IEEE International Joint Conference on Biometrics,
  IJCB 2017} \bibinfo{volume}{2018-Janua}, \bibinfo{pages}{464--473}.
\newblock \URLprefix \url{http://arxiv.org/abs/1611.01484}.
\bibitem[{Barbu et~al.(2019)Barbu, Mayo, Alverio, Luo, Wang, Gutfreund,
  Tenenbaum and Katz}]{Barbu2019}
\bibinfo{author}{Barbu, A.}, \bibinfo{author}{Mayo, D.},
  \bibinfo{author}{Alverio, J.}, \bibinfo{author}{Luo, W.},
  \bibinfo{author}{Wang, C.}, \bibinfo{author}{Gutfreund, D.},
  \bibinfo{author}{Tenenbaum, J.}, \bibinfo{author}{Katz, B.},
  \bibinfo{year}{2019}.
\newblock \bibinfo{title}{{ObjectNet: A large-scale bias-controlled dataset for
  pushing the limits of object recognition models}}.
\newblock \bibinfo{journal}{Advances in neural information processing systems}
  , \bibinfo{pages}{1--11}\URLprefix \url{https://objectnet.dev.}
\bibitem[{Bay et~al.(2006)Bay, Tuytelaars and Van~Gool}]{bay2006}
\bibinfo{author}{Bay, H.}, \bibinfo{author}{Tuytelaars, T.},
  \bibinfo{author}{Van~Gool, L.}, \bibinfo{year}{2006}.
\newblock \bibinfo{title}{Surf: Speeded up robust features} ,
  \bibinfo{pages}{404--417}.
\bibitem[{Behley et~al.(2019)Behley, Garbade, Milioto, Quenzel, Behnke,
  Stachniss and Gall}]{Behley2019}
\bibinfo{author}{Behley, J.}, \bibinfo{author}{Garbade, M.},
  \bibinfo{author}{Milioto, A.}, \bibinfo{author}{Quenzel, J.},
  \bibinfo{author}{Behnke, S.}, \bibinfo{author}{Stachniss, C.},
  \bibinfo{author}{Gall, J.}, \bibinfo{year}{2019}.
\newblock \bibinfo{title}{{SemanticKITTI: A Dataset for Semantic Scene
  Understanding of LiDAR Sequences}} \URLprefix
  \url{http://arxiv.org/abs/1904.01416}.
\bibitem[{Bell et~al.(2013)Bell, Upchurch, Snavely and Bala}]{Bell2013}
\bibinfo{author}{Bell, S.}, \bibinfo{author}{Upchurch, P.},
  \bibinfo{author}{Snavely, N.}, \bibinfo{author}{Bala, K.},
  \bibinfo{year}{2013}.
\newblock \bibinfo{title}{{OPENSURFACES: A richly annotated catalog of surface
  appearance}}.
\newblock \bibinfo{journal}{ACM Transactions on Graphics} \bibinfo{volume}{32}.
\newblock \DOIprefix\doi{10.1145/2461912.2462002}.
\bibitem[{Bengio et~al.(2007)Bengio, Lamblin, Popovici and
  Larochelle}]{Bengio2007}
\bibinfo{author}{Bengio, Y.}, \bibinfo{author}{Lamblin, P.},
  \bibinfo{author}{Popovici, D.}, \bibinfo{author}{Larochelle, H.},
  \bibinfo{year}{2007}.
\newblock \bibinfo{title}{{Greedy layer-wise training of deep networks}}, in:
  \bibinfo{booktitle}{Advances in neural information processing systems}, pp.
  \bibinfo{pages}{153--160}.
\bibitem[{Berg et~al.(2004)Berg, Berg, Edwards, Maire, White, Teh,
  Learned-Miller and Forsyth}]{Berg2004}
\bibinfo{author}{Berg, T.L.}, \bibinfo{author}{Berg, A.C.},
  \bibinfo{author}{Edwards, J.}, \bibinfo{author}{Maire, M.},
  \bibinfo{author}{White, R.}, \bibinfo{author}{Teh, Y.W.},
  \bibinfo{author}{Learned-Miller, E.}, \bibinfo{author}{Forsyth, D.A.},
  \bibinfo{year}{2004}.
\newblock \bibinfo{title}{{Names and faces in the news}}.
\newblock \bibinfo{journal}{Proceedings of the IEEE Computer Society Conference
  on Computer Vision and Pattern Recognition} \bibinfo{volume}{2}.
\newblock \DOIprefix\doi{10.1109/cvpr.2004.1315253}.
\bibitem[{{Berkeley Deep Drive}(2018)}]{BerkeleyDeepDrive2018}
\bibinfo{author}{{Berkeley Deep Drive}}, \bibinfo{year}{2018}.
\newblock \bibinfo{title}{{CVPR 2018 - Berkeley DeepDrive challenges}}.
\bibitem[{Bernal et~al.(2012)Bernal, S{\'{a}}nchez and Vilarino}]{Bernal2012}
\bibinfo{author}{Bernal, J.}, \bibinfo{author}{S{\'{a}}nchez, J.},
  \bibinfo{author}{Vilarino, F.}, \bibinfo{year}{2012}.
\newblock \bibinfo{title}{{Towards automatic polyp detection with a polyp
  appearance model}}.
\newblock \bibinfo{journal}{Pattern Recognition} \bibinfo{volume}{45},
  \bibinfo{pages}{3166--3182}.
\newblock \DOIprefix\doi{https://doi.org/10.1016/j.patcog.2012.03.002}.
\bibitem[{Beumier and Acheroy(2000)}]{Beumier2000}
\bibinfo{author}{Beumier, C.}, \bibinfo{author}{Acheroy, M.},
  \bibinfo{year}{2000}.
\newblock \bibinfo{title}{{Automatic 3D face authentication}}.
\newblock \bibinfo{journal}{Image and Vision Computing} \bibinfo{volume}{18},
  \bibinfo{pages}{315--321}.
\newblock \DOIprefix\doi{10.1016/S0262-8856(99)00052-9}.
\bibitem[{Bien et~al.(2018)Bien, Rajpurkar, Ball, Irvin, Park, Jones, Bereket,
  Patel, Yeom, Shpanskaya, Halabi, Zucker, Fanton, Amanatullah, Beaulieu,
  Riley, Stewart, Blankenberg, Larson, Jones, Langlotz, Ng and
  Lungren}]{Bien2018}
\bibinfo{author}{Bien, N.}, \bibinfo{author}{Rajpurkar, P.},
  \bibinfo{author}{Ball, R.L.}, \bibinfo{author}{Irvin, J.},
  \bibinfo{author}{Park, A.}, \bibinfo{author}{Jones, E.},
  \bibinfo{author}{Bereket, M.}, \bibinfo{author}{Patel, B.N.},
  \bibinfo{author}{Yeom, K.W.}, \bibinfo{author}{Shpanskaya, K.},
  \bibinfo{author}{Halabi, S.}, \bibinfo{author}{Zucker, E.},
  \bibinfo{author}{Fanton, G.}, \bibinfo{author}{Amanatullah, D.F.},
  \bibinfo{author}{Beaulieu, C.F.}, \bibinfo{author}{Riley, G.M.},
  \bibinfo{author}{Stewart, R.J.}, \bibinfo{author}{Blankenberg, F.G.},
  \bibinfo{author}{Larson, D.B.}, \bibinfo{author}{Jones, R.H.},
  \bibinfo{author}{Langlotz, C.P.}, \bibinfo{author}{Ng, A.Y.},
  \bibinfo{author}{Lungren, M.P.}, \bibinfo{year}{2018}.
\newblock \bibinfo{title}{{Deep-learning-assisted diagnosis for knee magnetic
  resonance imaging: Development and retrospective validation of MRNet}}.
\newblock \bibinfo{journal}{PLoS Medicine} \bibinfo{volume}{15},
  \bibinfo{pages}{1--19}.
\newblock \DOIprefix\doi{10.1371/journal.pmed.1002699}.
\bibitem[{Bock et~al.(2019)Bock, Krajewski, Moers, Runde, Vater and
  Eckstein}]{Bock2019}
\bibinfo{author}{Bock, J.}, \bibinfo{author}{Krajewski, R.},
  \bibinfo{author}{Moers, T.}, \bibinfo{author}{Runde, S.},
  \bibinfo{author}{Vater, L.}, \bibinfo{author}{Eckstein, L.},
  \bibinfo{year}{2019}.
\newblock \bibinfo{title}{{The inD Dataset: A Drone Dataset of Naturalistic
  Road User Trajectories at German Intersections}} .
\bibitem[{Boom et~al.(2012)Boom, Huang, Beyan, Spampinato, Palazzo, He,
  Beauxis-Aussalet, Lin, Chou, Nadarajan, Chen-Burger, van Ossenbruggen,
  Giordano, Hardman, Lin and Fisher}]{Boom2012}
\bibinfo{author}{Boom, B.J.}, \bibinfo{author}{Huang, P.X.},
  \bibinfo{author}{Beyan, C.}, \bibinfo{author}{Spampinato, C.},
  \bibinfo{author}{Palazzo, S.}, \bibinfo{author}{He, J.},
  \bibinfo{author}{Beauxis-Aussalet, E.}, \bibinfo{author}{Lin, S.I.},
  \bibinfo{author}{Chou, H.M.}, \bibinfo{author}{Nadarajan, G.},
  \bibinfo{author}{Chen-Burger, Y.H.}, \bibinfo{author}{van Ossenbruggen, J.},
  \bibinfo{author}{Giordano, D.}, \bibinfo{author}{Hardman, L.},
  \bibinfo{author}{Lin, F.P.}, \bibinfo{author}{Fisher, R.B.},
  \bibinfo{year}{2012}.
\newblock \bibinfo{title}{{Long-term underwater camera surveillance for
  monitoring and analysis of fish populations}}.
\newblock \bibinfo{journal}{Workshop on Visual observation and Analysis of
  Animal and Insect Behavior (VAIB), in conjunction with ICPR 2012} ,
  \bibinfo{pages}{2--5}\URLprefix
  \url{http://homepages.inf.ed.ac.uk/rbf/VAIB12PAPERS/boom.pdf}.
\bibitem[{Botta et~al.(1993)Botta, Giordana and Saitta}]{Botta1993}
\bibinfo{author}{Botta, M.}, \bibinfo{author}{Giordana, A.},
  \bibinfo{author}{Saitta, L.}, \bibinfo{year}{1993}.
\newblock \bibinfo{title}{{Learning fuzzy concept definitions}}.
\newblock \bibinfo{journal}{1993 IEEE International Conference on Fuzzy
  Systems} , \bibinfo{pages}{18--22}\DOIprefix\doi{10.1109/fuzzy.1993.327470}.
\bibitem[{Bozcan and Kayacan(2020)}]{Bozcan2020}
\bibinfo{author}{Bozcan, I.}, \bibinfo{author}{Kayacan, E.},
  \bibinfo{year}{2020}.
\newblock \bibinfo{title}{{AU-AIR: A Multi-modal Unmanned Aerial Vehicle
  Dataset for Low Altitude Traffic Surveillance}} \URLprefix
  \url{http://arxiv.org/abs/2001.11737}.
\bibitem[{Braun et~al.(2018)Braun, Krebs, Flohr and Gavrila}]{Braun2018}
\bibinfo{author}{Braun, M.}, \bibinfo{author}{Krebs, S.},
  \bibinfo{author}{Flohr, F.}, \bibinfo{author}{Gavrila, D.M.},
  \bibinfo{year}{2018}.
\newblock \bibinfo{title}{{The EuroCity Persons Dataset: A Novel Benchmark for
  Object Detection}}.
\newblock \bibinfo{journal}{IEEE Transactions on Pattern Analysis and Machine
  Intelligence} \bibinfo{volume}{41}, \bibinfo{pages}{1844--1861}.
\newblock \URLprefix \url{http://arxiv.org/abs/1805.07193
  http://dx.doi.org/10.1109/TPAMI.2019.2897684},
  \DOIprefix\doi{10.1109/TPAMI.2019.2897684}.
\bibitem[{Brostow et~al.(2008a)Brostow, Shotton, Fauqueur and
  Cipolla}]{Brostow2008a}
\bibinfo{author}{Brostow, G.}, \bibinfo{author}{Shotton, J.},
  \bibinfo{author}{Fauqueur, J.}, \bibinfo{author}{Cipolla, R.},
  \bibinfo{year}{2008}a.
\newblock \bibinfo{title}{{Segmentation and Recognition using SfM Point
  Clouds}}.
\newblock \bibinfo{journal}{Eccv} , \bibinfo{pages}{1--15}.
\bibitem[{Brostow et~al.(2008b)Brostow, Shotton, Fauqueur and
  Cipolla}]{Brostow2008}
\bibinfo{author}{Brostow, G.J.}, \bibinfo{author}{Shotton, J.},
  \bibinfo{author}{Fauqueur, J.}, \bibinfo{author}{Cipolla, R.},
  \bibinfo{year}{2008}b.
\newblock \bibinfo{title}{{Segmentation and Recognition Using Structure from
  Motion Point Clouds}} , \bibinfo{pages}{44--57}.
\bibitem[{Brox and Malik(2010)}]{10.1007/978-3-642-15555-0_21}
\bibinfo{author}{Brox, T.}, \bibinfo{author}{Malik, J.}, \bibinfo{year}{2010}.
\newblock \bibinfo{title}{{Object Segmentation by Long Term Analysis of Point
  Trajectories}}, in: \bibinfo{editor}{Daniilidis, K.},
  \bibinfo{editor}{Maragos, P.}, \bibinfo{editor}{Paragios, N.} (Eds.),
  \bibinfo{booktitle}{Computer Vision -- ECCV 2010},
  \bibinfo{publisher}{Springer Berlin Heidelberg}, \bibinfo{address}{Berlin,
  Heidelberg}. pp. \bibinfo{pages}{282--295}.
\bibitem[{Caelles et~al.(2019)Caelles, Pont-Tuset, Perazzi, Montes, Maninis and
  {Van Gool}}]{Caelles2019}
\bibinfo{author}{Caelles, S.}, \bibinfo{author}{Pont-Tuset, J.},
  \bibinfo{author}{Perazzi, F.}, \bibinfo{author}{Montes, A.},
  \bibinfo{author}{Maninis, K.K.}, \bibinfo{author}{{Van Gool}, L.},
  \bibinfo{year}{2019}.
\newblock \bibinfo{title}{{The 2019 DAVIS Challenge on VOS: Unsupervised
  Multi-Object Segmentation}} , \bibinfo{pages}{1--4}\URLprefix
  \url{http://arxiv.org/abs/1905.00737}.
\bibitem[{Caesar et~al.(2019)Caesar, Bankiti, Lang, Vora, Liong, Xu, Krishnan,
  Pan, Baldan and Beijbom}]{Caesar2019}
\bibinfo{author}{Caesar, H.}, \bibinfo{author}{Bankiti, V.},
  \bibinfo{author}{Lang, A.H.}, \bibinfo{author}{Vora, S.},
  \bibinfo{author}{Liong, V.E.}, \bibinfo{author}{Xu, Q.},
  \bibinfo{author}{Krishnan, A.}, \bibinfo{author}{Pan, Y.},
  \bibinfo{author}{Baldan, G.}, \bibinfo{author}{Beijbom, O.},
  \bibinfo{year}{2019}.
\newblock \bibinfo{title}{{nuScenes: A multimodal dataset for autonomous
  driving}} .
\bibitem[{Caesar et~al.(2018)Caesar, Uijlings and Ferrari}]{Caesar2018}
\bibinfo{author}{Caesar, H.}, \bibinfo{author}{Uijlings, J.},
  \bibinfo{author}{Ferrari, V.}, \bibinfo{year}{2018}.
\newblock \bibinfo{title}{{COCO-Stuff Thing and Stuff Classes in Context -
  Caesar, Uijlings, Ferrari - 2016.pdf}} ,
  \bibinfo{pages}{1209--1218}\URLprefix
  \url{http://openaccess.thecvf.com/content{\_}cvpr{\_}2018/html/Caesar{\_}COCO-Stuff{\_}Thing{\_}and{\_}CVPR{\_}2018{\_}paper.html}.
\bibitem[{Cai et~al.(2015)Cai, Osman, Sharma, Landis and Li}]{Cai2015}
\bibinfo{author}{Cai, Y.}, \bibinfo{author}{Osman, S.},
  \bibinfo{author}{Sharma, M.}, \bibinfo{author}{Landis, M.},
  \bibinfo{author}{Li, S.}, \bibinfo{year}{2015}.
\newblock \bibinfo{title}{{Multi-Modality Vertebra Recognition in Arbitrary
  Views Using 3D Deformable Hierarchical Model}}.
\newblock \bibinfo{journal}{IEEE Transactions on Medical Imaging}
  \bibinfo{volume}{34}, \bibinfo{pages}{1676--1693}.
\newblock \DOIprefix\doi{10.1109/TMI.2015.2392054}.
\bibitem[{Caicedo et~al.(2019)Caicedo, Goodman, Karhohs, Cimini, Ackerman,
  Haghighi, Heng, Becker, Doan, McQuin, Rohban, Singh and
  Carpenter}]{Caicedo2019}
\bibinfo{author}{Caicedo, J.C.}, \bibinfo{author}{Goodman, A.},
  \bibinfo{author}{Karhohs, K.W.}, \bibinfo{author}{Cimini, B.A.},
  \bibinfo{author}{Ackerman, J.}, \bibinfo{author}{Haghighi, M.},
  \bibinfo{author}{Heng, C.}, \bibinfo{author}{Becker, T.},
  \bibinfo{author}{Doan, M.}, \bibinfo{author}{McQuin, C.},
  \bibinfo{author}{Rohban, M.}, \bibinfo{author}{Singh, S.},
  \bibinfo{author}{Carpenter, A.E.}, \bibinfo{year}{2019}.
\newblock \bibinfo{title}{{Nucleus segmentation across imaging experiments: the
  2018 Data Science Bowl}}.
\newblock \bibinfo{journal}{Nature Methods} \bibinfo{volume}{16},
  \bibinfo{pages}{1247--1253}.
\newblock \URLprefix \url{https://doi.org/10.1038/s41592-019-0612-7},
  \DOIprefix\doi{10.1038/s41592-019-0612-7}.
\bibitem[{Canfield et~al.()Canfield, Kittler, Codella, Celebi, Dana, Halpern,
  Helba and Tschandl}]{Canfield}
\bibinfo{author}{Canfield}, \bibinfo{author}{Kittler, H.},
  \bibinfo{author}{Codella, N.}, \bibinfo{author}{Celebi, M.E.},
  \bibinfo{author}{Dana, K.}, \bibinfo{author}{Halpern, A.},
  \bibinfo{author}{Helba, B.}, \bibinfo{author}{Tschandl, P.}, .
\newblock \bibinfo{title}{{ISIC 2018}}.
\newblock \URLprefix \url{https://challenge2018.isic-archive.com/}.
\bibitem[{Cao et~al.(2018)Cao, Shen, Xie, Parkhi and Zisserman}]{Cao2018}
\bibinfo{author}{Cao, Q.}, \bibinfo{author}{Shen, L.}, \bibinfo{author}{Xie,
  W.}, \bibinfo{author}{Parkhi, O.M.}, \bibinfo{author}{Zisserman, A.},
  \bibinfo{year}{2018}.
\newblock \bibinfo{title}{{VGGFace2: A dataset for recognising faces across
  pose and age}}, in: \bibinfo{booktitle}{Proceedings - 13th IEEE International
  Conference on Automatic Face and Gesture Recognition, FG 2018},
  \bibinfo{publisher}{Institute of Electrical and Electronics Engineers Inc.}.
  pp. \bibinfo{pages}{67--74}.
\newblock \DOIprefix\doi{10.1109/FG.2018.00020}.
\bibitem[{Chang et~al.(2019)Chang, Lambert, Sangkloy, Singh, Bak, Hartnett,
  Wang, Carr, Lucey, Ramanan and Hays}]{Chang2019}
\bibinfo{author}{Chang, M.F.}, \bibinfo{author}{Lambert, J.},
  \bibinfo{author}{Sangkloy, P.}, \bibinfo{author}{Singh, J.},
  \bibinfo{author}{Bak, S.}, \bibinfo{author}{Hartnett, A.},
  \bibinfo{author}{Wang, D.}, \bibinfo{author}{Carr, P.},
  \bibinfo{author}{Lucey, S.}, \bibinfo{author}{Ramanan, D.},
  \bibinfo{author}{Hays, J.}, \bibinfo{year}{2019}.
\newblock \bibinfo{title}{{Argoverse: 3D tracking and forecasting with rich
  maps}}.
\newblock \bibinfo{journal}{Proceedings of the IEEE Computer Society Conference
  on Computer Vision and Pattern Recognition} \bibinfo{volume}{2019-June},
  \bibinfo{pages}{8740--8749}.
\newblock \DOIprefix\doi{10.1109/CVPR.2019.00895}.
\bibitem[{Chatfield et~al.(2014)Chatfield, Simonyan, Vedaldi and
  Zisserman}]{Chatfield2014}
\bibinfo{author}{Chatfield, K.}, \bibinfo{author}{Simonyan, K.},
  \bibinfo{author}{Vedaldi, A.}, \bibinfo{author}{Zisserman, A.},
  \bibinfo{year}{2014}.
\newblock \bibinfo{title}{{Return of the devil in the details: Delving deep
  into convolutional nets}}.
\newblock \bibinfo{journal}{BMVC 2014 - Proceedings of the British Machine
  Vision Conference 2014} ,
  \bibinfo{pages}{1--11}\DOIprefix\doi{10.5244/c.28.6}.
\bibitem[{Che et~al.(2019)Che, Li, Li, Jiang, Shi, Zhang, Lu, Wu, Liu and
  Ye}]{Che2019}
\bibinfo{author}{Che, Z.}, \bibinfo{author}{Li, G.}, \bibinfo{author}{Li, T.},
  \bibinfo{author}{Jiang, B.}, \bibinfo{author}{Shi, X.},
  \bibinfo{author}{Zhang, X.}, \bibinfo{author}{Lu, Y.}, \bibinfo{author}{Wu,
  G.}, \bibinfo{author}{Liu, Y.}, \bibinfo{author}{Ye, J.},
  \bibinfo{year}{2019}.
\newblock \bibinfo{title}{{D{\^{}}2-City: A Large-Scale Dashcam Video Dataset
  of Diverse Traffic Scenarios}} \URLprefix
  \url{http://arxiv.org/abs/1904.01975}.
\bibitem[{Chellapilla et~al.(2006)Chellapilla, Puri and
  Simard}]{Chellapilla2006}
\bibinfo{author}{Chellapilla, K.}, \bibinfo{author}{Puri, S.},
  \bibinfo{author}{Simard, P.}, \bibinfo{year}{2006}.
\newblock \bibinfo{title}{{High Performance Convolutional Neural Networks for
  Document Processing}}, in: \bibinfo{editor}{Lorette, G.} (Ed.),
  \bibinfo{booktitle}{Tenth International Workshop on Frontiers in Handwriting
  Recognition}, \bibinfo{publisher}{Suvisoft}, \bibinfo{address}{La Baule
  (France)}.
\bibitem[{Chen et~al.(2014a)Chen, Chen and Hsu}]{Chenio}
\bibinfo{author}{Chen, B.C.}, \bibinfo{author}{Chen, C.S.},
  \bibinfo{author}{Hsu, W.}, \bibinfo{year}{2014}a.
\newblock \bibinfo{title}{{Cross-Age Reference Coding for Age-Invariant Face
  Recognition and Retrieval}} , \bibinfo{pages}{768--783}.
\bibitem[{Chen et~al.(2017a)Chen, Papandreou, Kokkinos, Murphy and
  Yuille}]{Chen2017deep}
\bibinfo{author}{Chen, L.C.}, \bibinfo{author}{Papandreou, G.},
  \bibinfo{author}{Kokkinos, I.}, \bibinfo{author}{Murphy, K.},
  \bibinfo{author}{Yuille, A.L.}, \bibinfo{year}{2017}a.
\newblock \bibinfo{title}{{Deeplab: Semantic image segmentation with deep
  convolutional nets, atrous convolution, and fully connected crfs}}.
\newblock \bibinfo{journal}{IEEE transactions on pattern analysis and machine
  intelligence} \bibinfo{volume}{40}, \bibinfo{pages}{834--848}.
\bibitem[{Chen et~al.(2017b)Chen, Papandreou, Kokkinos, Murphy and
  Yuille}]{Chen2017}
\bibinfo{author}{Chen, L.C.}, \bibinfo{author}{Papandreou, G.},
  \bibinfo{author}{Kokkinos, I.}, \bibinfo{author}{Murphy, K.},
  \bibinfo{author}{Yuille, A.L.}, \bibinfo{year}{2017}b.
\newblock \bibinfo{title}{{Deeplab: Semantic image segmentation with deep
  convolutional nets, atrous convolution, and fully connected crfs}}.
\newblock \bibinfo{journal}{IEEE transactions on pattern analysis and machine
  intelligence} \bibinfo{volume}{40}, \bibinfo{pages}{834--848}.
\bibitem[{Chen et~al.(2017c)Chen, Papandreou, Schroff and Adam}]{Chen2017a}
\bibinfo{author}{Chen, L.C.}, \bibinfo{author}{Papandreou, G.},
  \bibinfo{author}{Schroff, F.}, \bibinfo{author}{Adam, H.},
  \bibinfo{year}{2017}c.
\newblock \bibinfo{title}{{Rethinking atrous convolution for semantic image
  segmentation}}.
\newblock \bibinfo{journal}{arXiv preprint arXiv:1706.05587} .
\bibitem[{Chen et~al.(2019)Chen, Girshick, He and Doll{\'{a}}r}]{Chen2019}
\bibinfo{author}{Chen, X.}, \bibinfo{author}{Girshick, R.},
  \bibinfo{author}{He, K.}, \bibinfo{author}{Doll{\'{a}}r, P.},
  \bibinfo{year}{2019}.
\newblock \bibinfo{title}{{TensorMask: A Foundation for Dense Object
  Segmentation}} .
\bibitem[{Chen et~al.(2014b)Chen, Mottaghi, Liu, Fidler, Urtasun and
  Yuille}]{chen2014}
\bibinfo{author}{Chen, X.}, \bibinfo{author}{Mottaghi, R.},
  \bibinfo{author}{Liu, X.}, \bibinfo{author}{Fidler, S.},
  \bibinfo{author}{Urtasun, R.}, \bibinfo{author}{Yuille, A.},
  \bibinfo{year}{2014}b.
\newblock \bibinfo{title}{Detect what you can: Detecting and representing
  objects using holistic models and body parts}, in:
  \bibinfo{booktitle}{Proceedings of the 2014 IEEE Conference on Computer
  Vision and Pattern Recognition}, \bibinfo{publisher}{IEEE Computer Society},
  \bibinfo{address}{USA}. p. \bibinfo{pages}{1979–1986}.
\newblock \URLprefix \url{https://doi.org/10.1109/CVPR.2014.254},
  \DOIprefix\doi{10.1109/CVPR.2014.254}.
\bibitem[{Cheng and Han(2016)}]{Cheng2016}
\bibinfo{author}{Cheng, G.}, \bibinfo{author}{Han, J.}, \bibinfo{year}{2016}.
\newblock \bibinfo{title}{{A survey on object detection in optical remote
  sensing images}}.
\newblock \DOIprefix\doi{10.1016/j.isprsjprs.2016.03.014}.
\bibitem[{Cheng et~al.(2017)Cheng, Han and Lu}]{Cheng2017}
\bibinfo{author}{Cheng, G.}, \bibinfo{author}{Han, J.}, \bibinfo{author}{Lu,
  X.}, \bibinfo{year}{2017}.
\newblock \bibinfo{title}{{Remote Sensing Image Scene Classification: Benchmark
  and State of the Art}}.
\newblock \bibinfo{journal}{Proceedings of the IEEE} \bibinfo{volume}{105},
  \bibinfo{pages}{1865--1883}.
\newblock \URLprefix \url{http://arxiv.org/abs/1703.00121
  http://dx.doi.org/10.1109/JPROC.2017.2675998},
  \DOIprefix\doi{10.1109/JPROC.2017.2675998}.
\bibitem[{Cheng et~al.(2013)Cheng, Mitra, Huang and Hu}]{THUR15K}
\bibinfo{author}{Cheng, M.M.}, \bibinfo{author}{Mitra, N.},
  \bibinfo{author}{Huang, X.}, \bibinfo{author}{Hu, S.M.},
  \bibinfo{year}{2013}.
\newblock \bibinfo{title}{Salientshape: Group saliency in image collections}.
\newblock \bibinfo{journal}{The Visual Computer} \bibinfo{volume}{30},
  \bibinfo{pages}{1--10}.
\newblock \DOIprefix\doi{10.1007/s00371-013-0867-4}.
\bibitem[{Choi et~al.(2018)Choi, Kim, Hwang, Park, Yoon, An and
  Kweon}]{Choi2018}
\bibinfo{author}{Choi, Y.}, \bibinfo{author}{Kim, N.}, \bibinfo{author}{Hwang,
  S.}, \bibinfo{author}{Park, K.}, \bibinfo{author}{Yoon, J.S.},
  \bibinfo{author}{An, K.}, \bibinfo{author}{Kweon, I.S.},
  \bibinfo{year}{2018}.
\newblock \bibinfo{title}{{KAIST Multi-Spectral Day/Night Data Set for
  Autonomous and Assisted Driving}}.
\newblock \bibinfo{journal}{IEEE Transactions on Intelligent Transportation
  Systems} \bibinfo{volume}{19}, \bibinfo{pages}{934--948}.
\newblock \DOIprefix\doi{10.1109/TITS.2018.2791533}.
\bibitem[{Chollet(2017)}]{Chollet2017}
\bibinfo{author}{Chollet, F.}, \bibinfo{year}{2017}.
\newblock \bibinfo{title}{{Xception: Deep learning with depthwise separable
  convolutions}}.
\newblock \bibinfo{journal}{Proceedings - 30th IEEE Conference on Computer
  Vision and Pattern Recognition, CVPR 2017} \bibinfo{volume}{2017-Janua},
  \bibinfo{pages}{1800--1807}.
\newblock \DOIprefix\doi{10.1109/CVPR.2017.195}.
\bibitem[{Christie et~al.(2017)Christie, Fendley, Wilson and
  Mukherjee}]{Christie2017}
\bibinfo{author}{Christie, G.}, \bibinfo{author}{Fendley, N.},
  \bibinfo{author}{Wilson, J.}, \bibinfo{author}{Mukherjee, R.},
  \bibinfo{year}{2017}.
\newblock \bibinfo{title}{{Functional Map of the World}}.
\newblock \bibinfo{journal}{Proceedings of the IEEE Computer Society Conference
  on Computer Vision and Pattern Recognition} ,
  \bibinfo{pages}{6172--6180}\URLprefix \url{http://arxiv.org/abs/1711.07846}.
\bibitem[{Codella et~al.(2019)Codella, Rotemberg, Tschandl, Celebi, Dusza,
  Gutman, Helba, Kalloo, Liopyris, Marchetti, Kittler and
  Halpern}]{Codella2019}
\bibinfo{author}{Codella, N.}, \bibinfo{author}{Rotemberg, V.},
  \bibinfo{author}{Tschandl, P.}, \bibinfo{author}{Celebi, M.E.},
  \bibinfo{author}{Dusza, S.}, \bibinfo{author}{Gutman, D.},
  \bibinfo{author}{Helba, B.}, \bibinfo{author}{Kalloo, A.},
  \bibinfo{author}{Liopyris, K.}, \bibinfo{author}{Marchetti, M.},
  \bibinfo{author}{Kittler, H.}, \bibinfo{author}{Halpern, A.},
  \bibinfo{year}{2019}.
\newblock \bibinfo{title}{{Skin Lesion Analysis Toward Melanoma Detection 2018:
  A Challenge Hosted by the International Skin Imaging Collaboration (ISIC)}} ,
  \bibinfo{pages}{1--12}.
\bibitem[{Coifman and Li(2017)}]{Coifman2017}
\bibinfo{author}{Coifman, B.}, \bibinfo{author}{Li, L.}, \bibinfo{year}{2017}.
\newblock \bibinfo{title}{{A critical evaluation of the Next Generation
  Simulation (NGSIM) vehicle trajectory dataset}}.
\newblock \bibinfo{journal}{Transportation Research Part B: Methodological}
  \bibinfo{volume}{105}, \bibinfo{pages}{362--377}.
\newblock \DOIprefix\doi{10.1016/j.trb.2017.09.018}.
\bibitem[{Cordts et~al.(2016)Cordts, Omran, Ramos, Rehfeld, Enzweiler,
  Benenson, Franke, Roth and Schiele}]{Cordts2016}
\bibinfo{author}{Cordts, M.}, \bibinfo{author}{Omran, M.},
  \bibinfo{author}{Ramos, S.}, \bibinfo{author}{Rehfeld, T.},
  \bibinfo{author}{Enzweiler, M.}, \bibinfo{author}{Benenson, R.},
  \bibinfo{author}{Franke, U.}, \bibinfo{author}{Roth, S.},
  \bibinfo{author}{Schiele, B.}, \bibinfo{year}{2016}.
\newblock \bibinfo{title}{{The Cityscapes Dataset for Semantic Urban Scene
  Understanding}}.
\newblock \bibinfo{journal}{Proceedings of the IEEE Computer Society Conference
  on Computer Vision and Pattern Recognition} \bibinfo{volume}{2016-Decem},
  \bibinfo{pages}{3213--3223}.
\newblock \DOIprefix\doi{10.1109/CVPR.2016.350}.
\bibitem[{Cuzzolin et~al.(2020a)Cuzzolin, Bawa, Skarga-Bandurova and
  Singh}]{Cuzzolin2020}
\bibinfo{author}{Cuzzolin, F.}, \bibinfo{author}{Bawa, V.S.},
  \bibinfo{author}{Skarga-Bandurova, I.}, \bibinfo{author}{Singh, G.},
  \bibinfo{year}{2020}a.
\newblock \bibinfo{title}{{SARAS-ESAD 2020}}.
\bibitem[{Cuzzolin et~al.(2020b)Cuzzolin, Bawa, Skarga-Bandurova and
  Singh}]{SARAS2020}
\bibinfo{author}{Cuzzolin, F.}, \bibinfo{author}{Bawa, V.S.},
  \bibinfo{author}{Skarga-Bandurova, I.}, \bibinfo{author}{Singh, G.},
  \bibinfo{year}{2020}b.
\newblock \bibinfo{title}{{SARAS-ESAD Dataset}}.
\newblock \URLprefix \url{https://saras-esad.grand-challenge.org/Dataset/}.
\bibitem[{Dalal and Triggs(2005)}]{Dalal2005}
\bibinfo{author}{Dalal, N.}, \bibinfo{author}{Triggs, B.},
  \bibinfo{year}{2005}.
\newblock \bibinfo{title}{{Histograms of oriented gradients for human
  detection}}.
\newblock \bibinfo{journal}{Proceedings - 2005 IEEE Computer Society Conference
  on Computer Vision and Pattern Recognition, CVPR 2005} \bibinfo{volume}{I},
  \bibinfo{pages}{886--893}.
\newblock \DOIprefix\doi{10.1109/CVPR.2005.177}.
\bibitem[{David et~al.(2019)David, Bryan, Amirata, {Matt P.}, {Euan A.}, {David
  H.} and {James Y.}}]{David2019}
\bibinfo{author}{David, O.}, \bibinfo{author}{Bryan, H.},
  \bibinfo{author}{Amirata, G.}, \bibinfo{author}{{Matt P.}, L.},
  \bibinfo{author}{{Euan A.}, A.}, \bibinfo{author}{{David H.}, L.},
  \bibinfo{author}{{James Y.}, Z.}, \bibinfo{year}{2019}.
\newblock \bibinfo{title}{{EchoNet-Dynamic Dataset}}.
\bibitem[{Demir et~al.()Demir, Koperski, Lindenbaum, Pang, Huang, Basu, Hughes,
  Tuia, Raskar and Works}]{Demir}
\bibinfo{author}{Demir, I.}, \bibinfo{author}{Koperski, K.},
  \bibinfo{author}{Lindenbaum, D.}, \bibinfo{author}{Pang, G.},
  \bibinfo{author}{Huang, J.}, \bibinfo{author}{Basu, S.},
  \bibinfo{author}{Hughes, F.}, \bibinfo{author}{Tuia, D.},
  \bibinfo{author}{Raskar, R.}, \bibinfo{author}{Works, C.}, .
\newblock \bibinfo{title}{{DeepGlobe 2018: A Challenge to Parse the Earth
  through Satellite Images}}.
\newblock \bibinfo{type}{Technical Report}.
\bibitem[{Deng et~al.(2010)Deng, Dong, Socher, Li, {Kai Li} and {Li
  Fei-Fei}}]{Deng2010}
\bibinfo{author}{Deng, J.}, \bibinfo{author}{Dong, W.},
  \bibinfo{author}{Socher, R.}, \bibinfo{author}{Li, L.J.},
  \bibinfo{author}{{Kai Li}}, \bibinfo{author}{{Li Fei-Fei}},
  \bibinfo{year}{2010}.
\newblock \bibinfo{title}{{ImageNet: A large-scale hierarchical image
  database}} ,
  \bibinfo{pages}{248--255}\DOIprefix\doi{10.1109/cvpr.2009.5206848}.
\bibitem[{DiDi(2019)}]{DiDi2019}
\bibinfo{author}{DiDi}, \bibinfo{year}{2019}.
\newblock \bibinfo{title}{{D2-City Detection Domain Adaptation Challenge}}.
\bibitem[{Djavadifar(2020)}]{Djavadifar2020}
\bibinfo{author}{Djavadifar, A.}, \bibinfo{year}{2020}.
\newblock \bibinfo{title}{{Automatic detection of geometrical anomalies in
  composites manufacturing : a deep learning-based computer vision approach}}.
\newblock Ph.D. thesis.
\bibitem[{Dollar et~al.(2010)Dollar, Wojek, Schiele and Perona}]{Dollar2010}
\bibinfo{author}{Dollar, P.}, \bibinfo{author}{Wojek, C.},
  \bibinfo{author}{Schiele, B.}, \bibinfo{author}{Perona, P.},
  \bibinfo{year}{2010}.
\newblock \bibinfo{title}{{Pedestrian detection: A benchmark}},
  \bibinfo{publisher}{Institute of Electrical and Electronics Engineers
  (IEEE)}. pp. \bibinfo{pages}{304--311}.
\newblock \DOIprefix\doi{10.1109/cvpr.2009.5206631}.
\bibitem[{ELCAP(2003)}]{SIMBA}
\bibinfo{author}{ELCAP}, \bibinfo{year}{2003}.
\newblock \bibinfo{title}{{ELCAP Public Lung Image Database}}.
\newblock \URLprefix \url{http://www.via.cornell.edu/lungdb.html}.
\bibitem[{Enzweiler and Gavrila(2009)}]{Enzweiler2009}
\bibinfo{author}{Enzweiler, M.}, \bibinfo{author}{Gavrila, D.M.},
  \bibinfo{year}{2009}.
\newblock \bibinfo{title}{{Monocular pedestrian detection: Survey and
  experiments}}, in: \bibinfo{booktitle}{IEEE Transactions on Pattern Analysis
  and Machine Intelligence}, pp. \bibinfo{pages}{2179--2195}.
\newblock \DOIprefix\doi{10.1109/TPAMI.2008.260}.
\bibitem[{Etten et~al.()Etten, Lindenbaum and Bacastow}]{Etten}
\bibinfo{author}{Etten, A.V.}, \bibinfo{author}{Lindenbaum, D.},
  \bibinfo{author}{Bacastow, T.}, .
\newblock \bibinfo{title}{{SpaceNet: A Remote Sensing Dataset and Challenge
  Series}}.
\newblock \bibinfo{type}{Technical Report}.
\bibitem[{Everingham et~al.(2014)Everingham, Eslami, {Van Gool}, Williams, Winn
  and Zisserman}]{Everingham2014}
\bibinfo{author}{Everingham, M.}, \bibinfo{author}{Eslami, S.M.},
  \bibinfo{author}{{Van Gool}, L.}, \bibinfo{author}{Williams, C.K.},
  \bibinfo{author}{Winn, J.}, \bibinfo{author}{Zisserman, A.},
  \bibinfo{year}{2014}.
\newblock \bibinfo{title}{{The Pascal Visual Object Classes Challenge: A
  Retrospective}}.
\newblock \bibinfo{journal}{International Journal of Computer Vision}
  \bibinfo{volume}{111}, \bibinfo{pages}{98--136}.
\newblock \DOIprefix\doi{10.1007/s11263-014-0733-5}.
\bibitem[{Everingham et~al.(2006)Everingham, Sivic and
  Zisserman}]{Everingham2006}
\bibinfo{author}{Everingham, M.}, \bibinfo{author}{Sivic, J.},
  \bibinfo{author}{Zisserman, A.}, \bibinfo{year}{2006}.
\newblock \bibinfo{title}{{"Hello! My name is... Buffy" - Automatic naming of
  characters in TV video}}.
\newblock \bibinfo{journal}{BMVC 2006 - Proceedings of the British Machine
  Vision Conference 2006} , \bibinfo{pages}{899--908}.
\bibitem[{Everingham et~al.(2010)Everingham, {Van Gool}, Williams, Winn and
  Zisserman}]{Everingham2010}
\bibinfo{author}{Everingham, M.}, \bibinfo{author}{{Van Gool}, L.},
  \bibinfo{author}{Williams, C.K.}, \bibinfo{author}{Winn, J.},
  \bibinfo{author}{Zisserman, A.}, \bibinfo{year}{2010}.
\newblock \bibinfo{title}{{The pascal visual object classes (VOC) challenge}}.
\newblock \bibinfo{journal}{International Journal of Computer Vision}
  \bibinfo{volume}{88}, \bibinfo{pages}{303--338}.
\newblock \DOIprefix\doi{10.1007/s11263-009-0275-4}.
\bibitem[{Fan et~al.(2020a)Fan, Guolei, Cheng, Shen and Shao}]{CHAM}
\bibinfo{author}{Fan, D.p.}, \bibinfo{author}{Guolei, G.p.J.},
  \bibinfo{author}{Cheng, S.M.m.}, \bibinfo{author}{Shen, J.},
  \bibinfo{author}{Shao, L.}, \bibinfo{year}{2020}a.
\newblock \bibinfo{title}{{Camouflaged Object Detection}} .
\bibitem[{Fan et~al.(2020b)Fan, Ji, Sun, Cheng, Shen and Shao}]{COD10}
\bibinfo{author}{Fan, D.P.}, \bibinfo{author}{Ji, G.P.}, \bibinfo{author}{Sun,
  G.}, \bibinfo{author}{Cheng, M.M.}, \bibinfo{author}{Shen, J.},
  \bibinfo{author}{Shao, L.}, \bibinfo{year}{2020}b.
\newblock \bibinfo{title}{Camouflaged object detection} ,
  \bibinfo{pages}{2774--2784}\DOIprefix\doi{10.1109/CVPR42600.2020.00285}.
\bibitem[{Fan et~al.(2018)Fan, Liu, Gao, Hou, Borji and Cheng}]{SOC}
\bibinfo{author}{Fan, D.P.}, \bibinfo{author}{Liu, J.J.}, \bibinfo{author}{Gao,
  S.}, \bibinfo{author}{Hou, Q.}, \bibinfo{author}{Borji, A.},
  \bibinfo{author}{Cheng, M.M.}, \bibinfo{year}{2018}.
\newblock \bibinfo{title}{Salient objects in clutter: Bringing salient object
  detection to the foreground}.
\newblock \bibinfo{journal}{European Conference on Computer Vision (ECCV)} .
\bibitem[{Fan et~al.(2015)Fan, Zhong, Lischinski, Cohen-Or and
  Chen}]{10.1145/2816795.2818105}
\bibinfo{author}{Fan, Q.}, \bibinfo{author}{Zhong, F.},
  \bibinfo{author}{Lischinski, D.}, \bibinfo{author}{Cohen-Or, D.},
  \bibinfo{author}{Chen, B.}, \bibinfo{year}{2015}.
\newblock \bibinfo{title}{{JumpCut: Non-Successive Mask Transfer and
  Interpolation for Video Cutout}}.
\newblock \bibinfo{journal}{ACM Trans. Graph.} \bibinfo{volume}{34}.
\newblock \URLprefix \url{https://doi.org/10.1145/2816795.2818105},
  \DOIprefix\doi{10.1145/2816795.2818105}.
\bibitem[{{Fei- Fei} et~al.(2004){Fei- Fei}, Fergus and Perona}]{Fei-Fei2004}
\bibinfo{author}{{Fei- Fei}, L.}, \bibinfo{author}{Fergus, R.},
  \bibinfo{author}{Perona, P.}, \bibinfo{year}{2004}.
\newblock \bibinfo{title}{{Learning Generative Visual Models from Few Training
  Examples :}}.
\newblock \bibinfo{journal}{Conference on Computer Vision and Pattern
  Recognition Workshop (CVPR 2004)} \bibinfo{volume}{00}, \bibinfo{pages}{178}.
\newblock \URLprefix \url{http://dx.doi.org/10.1109/CVPR.2004.109},
  \DOIprefix\doi{10.1109/CVPR.2004.109}.
\bibitem[{Fellbaum(1998)}]{Fellbaum1998}
\bibinfo{author}{Fellbaum, C.}, \bibinfo{year}{1998}.
\newblock \bibinfo{title}{{WordNet: an Electronic Lexical Database}}.
\newblock \bibinfo{publisher}{Bradford Books}.
\bibitem[{{Felzenszwalb} et~al.(2010){Felzenszwalb}, {Girshick}, {McAllester}
  and {Ramanan}}]{Pedro2007}
\bibinfo{author}{{Felzenszwalb}, P.F.}, \bibinfo{author}{{Girshick}, R.B.},
  \bibinfo{author}{{McAllester}, D.}, \bibinfo{author}{{Ramanan}, D.},
  \bibinfo{year}{2010}.
\newblock \bibinfo{title}{Object detection with discriminatively trained
  part-based models}.
\newblock \bibinfo{journal}{IEEE Transactions on Pattern Analysis and Machine
  Intelligence} \bibinfo{volume}{32}, \bibinfo{pages}{1627--1645}.
\newblock \DOIprefix\doi{10.1109/TPAMI.2009.167}.
\bibitem[{Feng et~al.(2021)Feng, Haase-Schütz, Rosenbaum, Hertlein, Gläser,
  Timm, Wiesbeck and Dietmayer}]{FengSemantic}
\bibinfo{author}{Feng, D.}, \bibinfo{author}{Haase-Schütz, C.},
  \bibinfo{author}{Rosenbaum, L.}, \bibinfo{author}{Hertlein, H.},
  \bibinfo{author}{Gläser, C.}, \bibinfo{author}{Timm, F.},
  \bibinfo{author}{Wiesbeck, W.}, \bibinfo{author}{Dietmayer, K.},
  \bibinfo{year}{2021}.
\newblock \bibinfo{title}{Deep multi-modal object detection and semantic
  segmentation for autonomous driving: Datasets, methods, and challenges}.
\newblock \bibinfo{journal}{IEEE Transactions on Intelligent Transportation
  Systems} \bibinfo{volume}{22}, \bibinfo{pages}{1341--1360}.
\newblock \DOIprefix\doi{10.1109/TITS.2020.2972974}.
\bibitem[{Flanders et~al.()Flanders, Prevedello, Shih, Halabi, Kalpathy-Cramer,
  Ball, Mongan, Stein, Kitamura, {Lungren, Mattew}, Choudhary, Cala.lesley,
  Coelho, Mogensen, Moron, Miller, Ikuta, Zohrabian, Mcdonnell, Lincoln, Shah,
  Joyner, Agarwal, Lee and Nath}]{Flanders}
\bibinfo{author}{Flanders, A.E.}, \bibinfo{author}{Prevedello, L.M.},
  \bibinfo{author}{Shih, G.}, \bibinfo{author}{Halabi, S.S.},
  \bibinfo{author}{Kalpathy-Cramer, J.}, \bibinfo{author}{Ball, R.},
  \bibinfo{author}{Mongan, J.T.}, \bibinfo{author}{Stein, A.},
  \bibinfo{author}{Kitamura, f.C.}, \bibinfo{author}{{Lungren, Mattew}, P.},
  \bibinfo{author}{Choudhary, G.}, \bibinfo{author}{Cala.lesley},
  \bibinfo{author}{Coelho, L.}, \bibinfo{author}{Mogensen, M.},
  \bibinfo{author}{Moron, F.}, \bibinfo{author}{Miller, E.},
  \bibinfo{author}{Ikuta, I.}, \bibinfo{author}{Zohrabian, V.},
  \bibinfo{author}{Mcdonnell, O.}, \bibinfo{author}{Lincoln, C.},
  \bibinfo{author}{Shah, L.}, \bibinfo{author}{Joyner, D.},
  \bibinfo{author}{Agarwal, A.}, \bibinfo{author}{Lee, R.K.},
  \bibinfo{author}{Nath, J.}, .
\newblock \bibinfo{title}{{Construction of a Machine Learning Dataset through
  Collaboration: The RSNA 2019 Brain CT Hemorrhage Challenge}}.
\newblock \URLprefix
  \url{https://www.kaggle.com/c/rsna-intracranial-hemorrhage-detection},
  \DOIprefix\doi{https://doi.org/10.1148/ryai.2020190211}.
\bibitem[{Gan et~al.(2017)Gan, Lin, Wu, Peng, Zhang, Liao, Huang, Zheng and
  Zhang}]{Gan2017}
\bibinfo{author}{Gan, D.}, \bibinfo{author}{Lin, G.}, \bibinfo{author}{Wu, H.},
  \bibinfo{author}{Peng, J.}, \bibinfo{author}{Zhang, Y.},
  \bibinfo{author}{Liao, B.}, \bibinfo{author}{Huang, Y.},
  \bibinfo{author}{Zheng, Q.}, \bibinfo{author}{Zhang, N.},
  \bibinfo{year}{2017}.
\newblock \bibinfo{title}{{Research and development of power grid dispatching
  operation control system based on transmission section control}}.
\newblock \bibinfo{journal}{Dianli Xitong Baohu yu Kongzhi/Power System
  Protection and Control} \bibinfo{volume}{45}, \bibinfo{pages}{117--124}.
\newblock \DOIprefix\doi{10.7667/PSPC161855}.
\bibitem[{Garcia-Garcia et~al.(2017)Garcia-Garcia, Orts-Escolano, Oprea,
  Villena-Martinez and Garcia-Rodriguez}]{garciagarcia2017review}
\bibinfo{author}{Garcia-Garcia, A.}, \bibinfo{author}{Orts-Escolano, S.},
  \bibinfo{author}{Oprea, S.}, \bibinfo{author}{Villena-Martinez, V.},
  \bibinfo{author}{Garcia-Rodriguez, J.}, \bibinfo{year}{2017}.
\newblock \bibinfo{title}{A review on deep learning techniques applied to
  semantic segmentation} .
\bibitem[{Ge et~al.(2019)Ge, Zhang, Wang, Tang and Luo}]{Ge2019}
\bibinfo{author}{Ge, Y.}, \bibinfo{author}{Zhang, R.}, \bibinfo{author}{Wang,
  X.}, \bibinfo{author}{Tang, X.}, \bibinfo{author}{Luo, P.},
  \bibinfo{year}{2019}.
\newblock \bibinfo{title}{{Deepfashion2: A versatile benchmark for detection,
  pose estimation, segmentation and re-identification of clothing images}}, in:
  \bibinfo{booktitle}{Proceedings of the IEEE Computer Society Conference on
  Computer Vision and Pattern Recognition}, pp. \bibinfo{pages}{5332--5340}.
\newblock \DOIprefix\doi{10.1109/CVPR.2019.00548}.
\bibitem[{Geiger et~al.(a)Geiger, Lenz, Stiller and Urtasun}]{Geigera}
\bibinfo{author}{Geiger, A.}, \bibinfo{author}{Lenz, P.},
  \bibinfo{author}{Stiller, C.}, \bibinfo{author}{Urtasun, R.}, a.
\newblock \bibinfo{title}{{The KITTI 2D Object Evaluation Benchmark}}.
\bibitem[{Geiger et~al.(b)Geiger, Lenz, Stiller and Urtasun}]{Geiger}
\bibinfo{author}{Geiger, A.}, \bibinfo{author}{Lenz, P.},
  \bibinfo{author}{Stiller, C.}, \bibinfo{author}{Urtasun, R.}, b.
\newblock \bibinfo{title}{{The KITTI 3D Object Evaluation Benchmark}}.
\bibitem[{Geiger et~al.(2013)Geiger, Lenz, Stiller and Urtasun}]{Geiger2013}
\bibinfo{author}{Geiger, A.}, \bibinfo{author}{Lenz, P.},
  \bibinfo{author}{Stiller, C.}, \bibinfo{author}{Urtasun, R.},
  \bibinfo{year}{2013}.
\newblock \bibinfo{title}{{Vision meets robotics: The KITTI dataset}}.
\newblock \bibinfo{journal}{International Journal of Robotics Research}
  \bibinfo{volume}{32}, \bibinfo{pages}{1231--1237}.
\newblock \DOIprefix\doi{10.1177/0278364913491297}.
\bibitem[{Geiger et~al.(2012)Geiger, Lenz and Urtasun}]{Geiger2012}
\bibinfo{author}{Geiger, A.}, \bibinfo{author}{Lenz, P.},
  \bibinfo{author}{Urtasun, R.}, \bibinfo{year}{2012}.
\newblock \bibinfo{title}{{Are we ready for autonomous driving? the KITTI
  vision benchmark suite}}.
\newblock \bibinfo{journal}{Proceedings of the IEEE Computer Society Conference
  on Computer Vision and Pattern Recognition} ,
  \bibinfo{pages}{3354--3361}\DOIprefix\doi{10.1109/CVPR.2012.6248074}.
\bibitem[{Girshick(2015)}]{Girshick2015}
\bibinfo{author}{Girshick, R.}, \bibinfo{year}{2015}.
\newblock \bibinfo{title}{{Fast r-cnn}}, in: \bibinfo{booktitle}{Proceedings of
  the IEEE international conference on computer vision}, pp.
  \bibinfo{pages}{1440--1448}.
\bibitem[{Girshick et~al.(2014)Girshick, Donahue, Darrell and
  Malik}]{Girshick2014}
\bibinfo{author}{Girshick, R.}, \bibinfo{author}{Donahue, J.},
  \bibinfo{author}{Darrell, T.}, \bibinfo{author}{Malik, J.},
  \bibinfo{year}{2014}.
\newblock \bibinfo{title}{{Rich feature hierarchies for accurate object
  detection and semantic segmentation}}, in: \bibinfo{booktitle}{Proceedings of
  the IEEE conference on computer vision and pattern recognition}, pp.
  \bibinfo{pages}{580--587}.
\bibitem[{Go{\"{e}}au et~al.(2019)Go{\"{e}}au, Bonnet and Joly}]{Goeau2019}
\bibinfo{author}{Go{\"{e}}au, H.}, \bibinfo{author}{Bonnet, P.},
  \bibinfo{author}{Joly, A.}, \bibinfo{year}{2019}.
\newblock \bibinfo{title}{{Overview of LifeCLEF Plant identification task 2019:
  Diving into data deficient tropical countries}}, in: \bibinfo{booktitle}{CEUR
  Workshop Proceedings}, pp. \bibinfo{pages}{9--12}.
\bibitem[{Goldbaum(1975)}]{Goldbaum1975}
\bibinfo{author}{Goldbaum, M.}, \bibinfo{year}{1975}.
\newblock \bibinfo{title}{{STARE Database}}.
\bibitem[{Gould et~al.(2009)Gould, Fulton and Koller}]{Gould2009}
\bibinfo{author}{Gould, S.}, \bibinfo{author}{Fulton, R.},
  \bibinfo{author}{Koller, D.}, \bibinfo{year}{2009}.
\newblock \bibinfo{title}{{Decomposing a scene into geometric and semantically
  consistent regions}}.
\newblock \bibinfo{journal}{Proceedings of the IEEE International Conference on
  Computer Vision} ,
  \bibinfo{pages}{1--8}\DOIprefix\doi{10.1109/ICCV.2009.5459211}.
\bibitem[{Griffin and Greg(2007)}]{Griffin2007}
\bibinfo{author}{Griffin}, \bibinfo{author}{Greg}, \bibinfo{year}{2007}.
\newblock \bibinfo{title}{{Caltech-256 Object Category Dataset}} ,
  \bibinfo{pages}{300}.
\bibitem[{Guo et~al.(2016)Guo, Zhang, Hu, He and Gao}]{Guo2016}
\bibinfo{author}{Guo, Y.}, \bibinfo{author}{Zhang, L.}, \bibinfo{author}{Hu,
  Y.}, \bibinfo{author}{He, X.}, \bibinfo{author}{Gao, J.},
  \bibinfo{year}{2016}.
\newblock \bibinfo{title}{{MS-Celeb-1M: A Dataset and Benchmark for Large-Scale
  Face Recognition}} .
\bibitem[{Gupta et~al.(2019)Gupta, Dollar and Girshick}]{Gupta2019}
\bibinfo{author}{Gupta, A.}, \bibinfo{author}{Dollar, P.},
  \bibinfo{author}{Girshick, R.}, \bibinfo{year}{2019}.
\newblock \bibinfo{title}{{Lvis: A dataset for large vocabulary instance
  segmentation}}.
\newblock \bibinfo{journal}{Proceedings of the IEEE Computer Society Conference
  on Computer Vision and Pattern Recognition} \bibinfo{volume}{2019-June},
  \bibinfo{pages}{5351--5359}.
\newblock \DOIprefix\doi{10.1109/CVPR.2019.00550}.
\bibitem[{Hariharan et~al.(2011)Hariharan, Arbel{\'{a}}ez, Bourdev, Maji and
  Malik}]{Hariharan2011}
\bibinfo{author}{Hariharan, B.}, \bibinfo{author}{Arbel{\'{a}}ez, P.},
  \bibinfo{author}{Bourdev, L.}, \bibinfo{author}{Maji, S.},
  \bibinfo{author}{Malik, J.}, \bibinfo{year}{2011}.
\newblock \bibinfo{title}{{Semantic Contours from Inverse Detectors * -
  Hariharan et al.pdf}}.
\newblock \bibinfo{journal}{International Conference on Computer Vision} ,
  \bibinfo{pages}{8}\URLprefix
  \url{http://home.bharathh.info/pubs/pdfs/BharathICCV2011.pdf}.
\bibitem[{He et~al.(2017)He, Gkioxari, Doll{\'{a}}r and Girshick}]{He2017}
\bibinfo{author}{He, K.}, \bibinfo{author}{Gkioxari, G.},
  \bibinfo{author}{Doll{\'{a}}r, P.}, \bibinfo{author}{Girshick, R.},
  \bibinfo{year}{2017}.
\newblock \bibinfo{title}{{Mask r-cnn}}, in: \bibinfo{booktitle}{Proceedings of
  the IEEE international conference on computer vision}, pp.
  \bibinfo{pages}{2961--2969}.
\bibitem[{He et~al.(2016)He, Zhang, Ren and Sun}]{He2016}
\bibinfo{author}{He, K.}, \bibinfo{author}{Zhang, X.}, \bibinfo{author}{Ren,
  S.}, \bibinfo{author}{Sun, J.}, \bibinfo{year}{2016}.
\newblock \bibinfo{title}{{Deep residual learning for image recognition}}, in:
  \bibinfo{booktitle}{Proceedings of the IEEE Computer Society Conference on
  Computer Vision and Pattern Recognition}, \bibinfo{publisher}{IEEE Computer
  Society}. pp. \bibinfo{pages}{770--778}.
\newblock \URLprefix \url{http://image-net.org/challenges/LSVRC/2015/},
  \DOIprefix\doi{10.1109/CVPR.2016.90}.
\bibitem[{Heath et~al.()Heath, Bowyer, Kopans, Morre, {Kegelmeyer, W. Philip
  Chang} and Munishkumaran}]{Heath}
\bibinfo{author}{Heath, M.}, \bibinfo{author}{Bowyer, K.},
  \bibinfo{author}{Kopans, D.}, \bibinfo{author}{Morre, R.},
  \bibinfo{author}{{Kegelmeyer, W. Philip Chang}, K.},
  \bibinfo{author}{Munishkumaran, S.}, .
\newblock \bibinfo{title}{{Current Status of the Digital Database for Screening
  Mammography}}.
\newblock \bibinfo{journal}{Digital Mammography} ,
  \bibinfo{pages}{457--460}\DOIprefix\doi{https://doi.org/10.1007/978-94-011-5318-8_75}.
\bibitem[{Heath et~al.(2001)Heath, Bowyer, Kopans, Morre, {Kegelmeyer, W.
  Philip Chang} and Munishkumaran}]{Heath2001}
\bibinfo{author}{Heath, M.}, \bibinfo{author}{Bowyer, K.},
  \bibinfo{author}{Kopans, D.}, \bibinfo{author}{Morre, R.},
  \bibinfo{author}{{Kegelmeyer, W. Philip Chang}, K.},
  \bibinfo{author}{Munishkumaran, S.}, \bibinfo{year}{2001}.
\newblock \bibinfo{title}{{THE DIGITAL DATABASE FOR SCREENING MAMMOGRAPHY.}}
\newblock \bibinfo{publisher}{Medical Physics Publishing}.
\bibitem[{Heitz and Koller()}]{Heitz}
\bibinfo{author}{Heitz, G.}, \bibinfo{author}{Koller, D.}, .
\newblock \bibinfo{title}{{Learning Spatial Context: Using Stuff to Find
  Things}}.
\newblock \bibinfo{type}{Technical Report}.
\bibitem[{Heller et~al.(2019)Heller, Sathianathen, Kalapara, Walczak, Moore,
  Kaluzniak, Rosenberg, Blake, Rengel, Oestreich, Dean, Tradewell, Shah,
  Tejpaul, Edgerton, Peterson, Raza, Regmi, Papanikolopoulos and
  Weight}]{Heller2019}
\bibinfo{author}{Heller, N.}, \bibinfo{author}{Sathianathen, N.},
  \bibinfo{author}{Kalapara, A.}, \bibinfo{author}{Walczak, E.},
  \bibinfo{author}{Moore, K.}, \bibinfo{author}{Kaluzniak, H.},
  \bibinfo{author}{Rosenberg, J.}, \bibinfo{author}{Blake, P.},
  \bibinfo{author}{Rengel, Z.}, \bibinfo{author}{Oestreich, M.},
  \bibinfo{author}{Dean, J.}, \bibinfo{author}{Tradewell, M.},
  \bibinfo{author}{Shah, A.}, \bibinfo{author}{Tejpaul, R.},
  \bibinfo{author}{Edgerton, Z.}, \bibinfo{author}{Peterson, M.},
  \bibinfo{author}{Raza, S.}, \bibinfo{author}{Regmi, S.},
  \bibinfo{author}{Papanikolopoulos, N.}, \bibinfo{author}{Weight, C.},
  \bibinfo{year}{2019}.
\newblock \bibinfo{title}{{The KiTS19 Challenge Data: 300 Kidney Tumor Cases
  with Clinical Context, CT Semantic Segmentations, and Surgical Outcomes}} ,
  \bibinfo{pages}{1--14}.
\bibitem[{Hinton et~al.(2006)Hinton, Osindero and Teh}]{Hinton2006}
\bibinfo{author}{Hinton, G.E.}, \bibinfo{author}{Osindero, S.},
  \bibinfo{author}{Teh, Y.W.}, \bibinfo{year}{2006}.
\newblock \bibinfo{title}{{A fast learning algorithm for deep belief nets}}.
\newblock \bibinfo{journal}{Neural computation} \bibinfo{volume}{18},
  \bibinfo{pages}{1527--1554}.
\bibitem[{Hinton and Salakhutdinov(2006)}]{Hinton2006a}
\bibinfo{author}{Hinton, G.E.}, \bibinfo{author}{Salakhutdinov, R.R.},
  \bibinfo{year}{2006}.
\newblock \bibinfo{title}{{Reducing the dimensionality of data with neural
  networks}}.
\newblock \bibinfo{journal}{science} \bibinfo{volume}{313},
  \bibinfo{pages}{504--507}.
\bibitem[{Hong-Wei and Stefan(2014)}]{Hong-Wei2014}
\bibinfo{author}{Hong-Wei, N.}, \bibinfo{author}{Stefan, W.},
  \bibinfo{year}{2014}.
\newblock \bibinfo{title}{{A DATA-DRIVEN APPROACH TO CLEANING LARGE FACE
  DATASETS}}.
\newblock \bibinfo{journal}{International Conference on Image Processing(ICIP)}
  , \bibinfo{pages}{343--347}.
\bibitem[{Horn et~al.(2018)Horn, Aodha, Song, Cui, Sun, Shepard, Adam, Perona
  and Belongie}]{Horn2018}
\bibinfo{author}{Horn, G.V.}, \bibinfo{author}{Aodha, O.M.},
  \bibinfo{author}{Song, Y.}, \bibinfo{author}{Cui, Y.}, \bibinfo{author}{Sun,
  C.}, \bibinfo{author}{Shepard, A.}, \bibinfo{author}{Adam, H.},
  \bibinfo{author}{Perona, P.}, \bibinfo{author}{Belongie, S.},
  \bibinfo{year}{2018}.
\newblock \bibinfo{title}{{The iNaturalist Species Classification and Detection
  Dataset}}, in: \bibinfo{booktitle}{Proceedings of the IEEE Computer Society
  Conference on Computer Vision and Pattern Recognition}, pp.
  \bibinfo{pages}{8769--8778}.
\newblock \DOIprefix\doi{10.1109/CVPR.2018.00914}.
\bibitem[{Horn et~al.()Horn, Branson, Farrell, Barry and Tech}]{Horn}
\bibinfo{author}{Horn, G.V.}, \bibinfo{author}{Branson, S.},
  \bibinfo{author}{Farrell, R.}, \bibinfo{author}{Barry, J.},
  \bibinfo{author}{Tech, C.}, .
\newblock \bibinfo{title}{{Building a bird recognition app and large scale
  dataset with citizen scientists : The fine print in fine-grained dataset
  collection}} .
\bibitem[{Hosseini et~al.(2019)Hosseini, Chan, Tse, Tang, Deng, Norouzi,
  Rowsell, Plataniotis and Damaskinos}]{Hosseini2019}
\bibinfo{author}{Hosseini, M.S.}, \bibinfo{author}{Chan, L.},
  \bibinfo{author}{Tse, G.}, \bibinfo{author}{Tang, M.}, \bibinfo{author}{Deng,
  J.}, \bibinfo{author}{Norouzi, S.}, \bibinfo{author}{Rowsell, C.},
  \bibinfo{author}{Plataniotis, K.N.}, \bibinfo{author}{Damaskinos, S.},
  \bibinfo{year}{2019}.
\newblock \bibinfo{title}{{Atlas of digital pathology: A generalized
  hierarchical histological tissue type-annotated database for deep learning}},
  in: \bibinfo{booktitle}{Proceedings of the IEEE Computer Society Conference
  on Computer Vision and Pattern Recognition}, \bibinfo{address}{Long Beach}.
  pp. \bibinfo{pages}{11739--11748}.
\newblock \DOIprefix\doi{10.1109/CVPR.2019.01202}.
\bibitem[{Huang et~al.(2008)Huang, Mattar, Berg, Learned-Miller and
  Learned-Miller}]{Huang2008}
\bibinfo{author}{Huang, G.B.}, \bibinfo{author}{Mattar, M.},
  \bibinfo{author}{Berg, T.}, \bibinfo{author}{Learned-Miller, E.},
  \bibinfo{author}{Learned-Miller, E.}, \bibinfo{year}{2008}.
\newblock \bibinfo{title}{{Labeled Faces in the Wild: A Database forStudying
  Face Recognition in Unconstrained Environments Labeled Faces in the Wild: A
  Database for Studying Face Recognition in Unconstrained Environments}}.
\newblock \bibinfo{type}{Technical Report}.
\newblock \URLprefix \url{https://hal.inria.fr/inria-00321923}.
\bibitem[{Huang et~al.(2015)Huang, Feris, Chen and Yan}]{Huang2015a}
\bibinfo{author}{Huang, J.}, \bibinfo{author}{Feris, R.},
  \bibinfo{author}{Chen, Q.}, \bibinfo{author}{Yan, S.}, \bibinfo{year}{2015}.
\newblock \bibinfo{title}{{Cross-domain image retrieval with a dual
  attribute-aware ranking network}}, in: \bibinfo{booktitle}{Proceedings of the
  IEEE International Conference on Computer Vision}, pp.
  \bibinfo{pages}{1062--1070}.
\newblock \DOIprefix\doi{10.1109/ICCV.2015.127}.
\bibitem[{Irvin et~al.(2019)Irvin, Rajpurkar, Ko, Yu, Ciurea-Ilcus, Chute,
  Marklund, Haghgoo, Ball, Shpanskaya, Seekins, Mong, Halabi, Sandberg, Jones,
  Larson, Langlotz, Patel, Lungren and Ng}]{Irvin2019}
\bibinfo{author}{Irvin, J.}, \bibinfo{author}{Rajpurkar, P.},
  \bibinfo{author}{Ko, M.}, \bibinfo{author}{Yu, Y.},
  \bibinfo{author}{Ciurea-Ilcus, S.}, \bibinfo{author}{Chute, C.},
  \bibinfo{author}{Marklund, H.}, \bibinfo{author}{Haghgoo, B.},
  \bibinfo{author}{Ball, R.}, \bibinfo{author}{Shpanskaya, K.},
  \bibinfo{author}{Seekins, J.}, \bibinfo{author}{Mong, D.A.},
  \bibinfo{author}{Halabi, S.S.}, \bibinfo{author}{Sandberg, J.K.},
  \bibinfo{author}{Jones, R.}, \bibinfo{author}{Larson, D.B.},
  \bibinfo{author}{Langlotz, C.P.}, \bibinfo{author}{Patel, B.N.},
  \bibinfo{author}{Lungren, M.P.}, \bibinfo{author}{Ng, A.Y.},
  \bibinfo{year}{2019}.
\newblock \bibinfo{title}{{CheXpert: A Large Chest Radiograph Dataset with
  Uncertainty Labels and Expert Comparison}}.
\newblock \bibinfo{journal}{Proceedings of the AAAI Conference on Artificial
  Intelligence} \bibinfo{volume}{33}, \bibinfo{pages}{590--597}.
\newblock \DOIprefix\doi{10.1609/aaai.v33i01.3301590}.
\bibitem[{Jacobs et~al.(2016)Jacobs, Setio, Traverso and Ginneken}]{Jacobs2016}
\bibinfo{author}{Jacobs, C.}, \bibinfo{author}{Setio, A.A.A.},
  \bibinfo{author}{Traverso, A.}, \bibinfo{author}{Ginneken, B.V.},
  \bibinfo{year}{2016}.
\newblock \bibinfo{title}{{LUNA 2016}}.
\bibitem[{Jain and Grauman(2014)}]{inproceedings}
\bibinfo{author}{Jain, S.}, \bibinfo{author}{Grauman, K.},
  \bibinfo{year}{2014}.
\newblock \bibinfo{title}{{Supervoxel-Consistent Foreground Propagation in
  Video}}, pp. \bibinfo{pages}{656--671}.
\newblock \DOIprefix\doi{10.1007/978-3-319-10593-2_43}.
\bibitem[{Jesorsky et~al.(2001)Jesorsky, Kirchberg and
  Frischholz}]{Jesorsky2001}
\bibinfo{author}{Jesorsky, O.}, \bibinfo{author}{Kirchberg, K.J.},
  \bibinfo{author}{Frischholz, R.W.}, \bibinfo{year}{2001}.
\newblock \bibinfo{title}{{Robust face detection using the Hausdorff
  distance}}.
\newblock \bibinfo{journal}{Lecture Notes in Computer Science (including
  subseries Lecture Notes in Artificial Intelligence and Lecture Notes in
  Bioinformatics)} \bibinfo{volume}{2091}, \bibinfo{pages}{90--95}.
\newblock \DOIprefix\doi{10.1007/3-540-45344-x_14}.
\bibitem[{{Jonathon Phillips} et~al.(2000){Jonathon Phillips}, Moon, Rizvi and
  Rauss}]{JonathonPhillips2000}
\bibinfo{author}{{Jonathon Phillips}, P.}, \bibinfo{author}{Moon, H.},
  \bibinfo{author}{Rizvi, S.A.}, \bibinfo{author}{Rauss, P.J.},
  \bibinfo{year}{2000}.
\newblock \bibinfo{title}{{The FERET evaluation methodology for
  face-recognition algorithms}}.
\newblock \bibinfo{journal}{IEEE Transactions on Pattern Analysis and Machine
  Intelligence} \bibinfo{volume}{22}, \bibinfo{pages}{1090--1104}.
\newblock \DOIprefix\doi{10.1109/34.879790}.
\bibitem[{Kaggle(2018)}]{Kaggle2018}
\bibinfo{author}{Kaggle}, \bibinfo{year}{2018}.
\newblock \bibinfo{title}{{CVPR 2018 WAD Video Segmentation Challenge}}.
\newblock
  \DOIprefix\doi{https://www.kaggle.com/c/cvpr-2018-autonomous-driving}.
\bibitem[{Kaggle.com(2017)}]{dstl}
\bibinfo{author}{Kaggle.com}, \bibinfo{year}{2017}.
\newblock \bibinfo{title}{Dstl satelite imagery feature detection}.
\newblock \URLprefix
  \url{https://www.kaggle.com/c/dstl-satellite-imagery-feature-detection}.
\bibitem[{K{\"{a}}rkk{\"{a}}inen and {Joo UCLA}()}]{Karkkainen}
\bibinfo{author}{K{\"{a}}rkk{\"{a}}inen, K.}, \bibinfo{author}{{Joo UCLA}, J.},
  .
\newblock \bibinfo{title}{{FairFace: Face Attribute Dataset for Balanced Race,
  Gender, and Age}}.
\newblock \bibinfo{type}{Technical Report}.
\newblock \URLprefix \url{https://github.com/joojs/fairface{\%}7D.}
\bibitem[{Kauppi et~al.(2007)Kauppi, Kalesnykiene, Kamarainen, Lensu, Sorri,
  Raninen, Voutilainen, Pietil{\"{a}}, K{\"{a}}lvi{\"{a}}inen and
  Uusitalo}]{Kauppi2007}
\bibinfo{author}{Kauppi, T.}, \bibinfo{author}{Kalesnykiene, V.},
  \bibinfo{author}{Kamarainen, J.K.}, \bibinfo{author}{Lensu, L.},
  \bibinfo{author}{Sorri, I.}, \bibinfo{author}{Raninen, A.},
  \bibinfo{author}{Voutilainen, R.}, \bibinfo{author}{Pietil{\"{a}}, J.},
  \bibinfo{author}{K{\"{a}}lvi{\"{a}}inen, H.}, \bibinfo{author}{Uusitalo, H.},
  \bibinfo{year}{2007}.
\newblock \bibinfo{title}{{The DIARETDB1 diabetic retinopathy database and
  evaluation protocol}}.
\newblock \bibinfo{journal}{BMVC 2007 - Proceedings of the British Machine
  Vision Conference 2007} ,
  \bibinfo{pages}{1--18}\DOIprefix\doi{10.5244/C.21.15}.
\bibitem[{Kemelmacher-Shlizerman et~al.(2016)Kemelmacher-Shlizerman, Seitz,
  Miller and Brossard}]{Kemelmacher-Shlizerman2016}
\bibinfo{author}{Kemelmacher-Shlizerman, I.}, \bibinfo{author}{Seitz, S.M.},
  \bibinfo{author}{Miller, D.}, \bibinfo{author}{Brossard, E.},
  \bibinfo{year}{2016}.
\newblock \bibinfo{title}{{The MegaFace benchmark: 1 million faces for
  recognition at scale}}.
\newblock \bibinfo{journal}{Proceedings of the IEEE Computer Society Conference
  on Computer Vision and Pattern Recognition} \bibinfo{volume}{2016-Decem},
  \bibinfo{pages}{4873--4882}.
\newblock \DOIprefix\doi{10.1109/CVPR.2016.527}.
\bibitem[{Kesten et~al.(2019)Kesten, Usman, Houston, Pandya, Nadhamuni,
  Ferreira, Yuan, Low, Jain, Ondruska, Omari, Shah, Kulkarni, Kazakova, Tao,
  Platinsky, Jiang and Shet.}]{Kesten2019}
\bibinfo{author}{Kesten, R.}, \bibinfo{author}{Usman, M.},
  \bibinfo{author}{Houston, J.}, \bibinfo{author}{Pandya, T.},
  \bibinfo{author}{Nadhamuni, K.}, \bibinfo{author}{Ferreira, A.},
  \bibinfo{author}{Yuan, M.}, \bibinfo{author}{Low, B.}, \bibinfo{author}{Jain,
  A.}, \bibinfo{author}{Ondruska, P.}, \bibinfo{author}{Omari, S.},
  \bibinfo{author}{Shah, S.}, \bibinfo{author}{Kulkarni, A.},
  \bibinfo{author}{Kazakova, A.}, \bibinfo{author}{Tao, C.},
  \bibinfo{author}{Platinsky, L.}, \bibinfo{author}{Jiang, W.},
  \bibinfo{author}{Shet., V.}, \bibinfo{year}{2019}.
\newblock \bibinfo{title}{{Lyft Level 5 AV Dataset}}.
\newblock \URLprefix \url{https://level5.lyft.com/dataset/}.
\bibitem[{Khan et~al.(2019)Khan, McDonagh, Khan, Shahabuddin, Arora, Khan, Shao
  and Tzimiropoulos}]{Khan2019}
\bibinfo{author}{Khan, M.H.}, \bibinfo{author}{McDonagh, J.},
  \bibinfo{author}{Khan, S.}, \bibinfo{author}{Shahabuddin, M.},
  \bibinfo{author}{Arora, A.}, \bibinfo{author}{Khan, F.S.},
  \bibinfo{author}{Shao, L.}, \bibinfo{author}{Tzimiropoulos, G.},
  \bibinfo{year}{2019}.
\newblock \bibinfo{title}{{AnimalWeb: A Large-Scale Hierarchical Dataset of
  Annotated Animal Faces}} , \bibinfo{pages}{1--15}\URLprefix
  \url{http://arxiv.org/abs/1909.04951}.
\bibitem[{Khosla et~al.(2011)Khosla, Jayadevaprakash, Yao and
  Fei-Fei}]{Khosla2011}
\bibinfo{author}{Khosla, A.}, \bibinfo{author}{Jayadevaprakash, N.},
  \bibinfo{author}{Yao, B.}, \bibinfo{author}{Fei-Fei, L.},
  \bibinfo{year}{2011}.
\newblock \bibinfo{title}{{Novel dataset for fine-grained image
  categorization}}.
\newblock \bibinfo{journal}{Proc. IEEE Conf. Comput. Vision and Pattern
  Recognition} .
\bibitem[{Kiapour et~al.(2015)Kiapour, Han, Lazebnik, Berg and
  Berg}]{Kiapour2015}
\bibinfo{author}{Kiapour, M.H.}, \bibinfo{author}{Han, X.},
  \bibinfo{author}{Lazebnik, S.}, \bibinfo{author}{Berg, A.C.},
  \bibinfo{author}{Berg, T.L.}, \bibinfo{year}{2015}.
\newblock \bibinfo{title}{{Where to buy it: Matching street clothing photos in
  online shops}}, in: \bibinfo{booktitle}{Proceedings of the IEEE International
  Conference on Computer Vision}, pp. \bibinfo{pages}{3343--3351}.
\newblock \DOIprefix\doi{10.1109/ICCV.2015.382}.
\bibitem[{Klare et~al.(2015)Klare, Klein, Taborsky, Blanton, Cheney, Allen,
  Grother, Mah, Burge and Jain}]{Klare2015}
\bibinfo{author}{Klare, B.F.}, \bibinfo{author}{Klein, B.},
  \bibinfo{author}{Taborsky, E.}, \bibinfo{author}{Blanton, A.},
  \bibinfo{author}{Cheney, J.}, \bibinfo{author}{Allen, K.},
  \bibinfo{author}{Grother, P.}, \bibinfo{author}{Mah, A.},
  \bibinfo{author}{Burge, M.}, \bibinfo{author}{Jain, A.K.},
  \bibinfo{year}{2015}.
\newblock \bibinfo{title}{{Pushing the frontiers of unconstrained face
  detection and recognition: IARPA Janus Benchmark A}}.
\newblock \bibinfo{journal}{Proceedings of the IEEE Computer Society Conference
  on Computer Vision and Pattern Recognition} \bibinfo{volume}{07-12-June},
  \bibinfo{pages}{1931--1939}.
\newblock \DOIprefix\doi{10.1109/CVPR.2015.7298803}.
\bibitem[{Krajewski et~al.(2018)Krajewski, Bock, Kloeker and
  Eckstein}]{Krajewski2018}
\bibinfo{author}{Krajewski, R.}, \bibinfo{author}{Bock, J.},
  \bibinfo{author}{Kloeker, L.}, \bibinfo{author}{Eckstein, L.},
  \bibinfo{year}{2018}.
\newblock \bibinfo{title}{{The highD Dataset: A Drone Dataset of Naturalistic
  Vehicle Trajectories on German Highways for Validation of Highly Automated
  Driving Systems}}.
\newblock \bibinfo{journal}{IEEE Conference on Intelligent Transportation
  Systems, Proceedings, ITSC} \bibinfo{volume}{2018-November},
  \bibinfo{pages}{2118--2125}.
\newblock \URLprefix \url{http://arxiv.org/abs/1810.05642}.
\bibitem[{Krishna et~al.(2017)Krishna, Zhu, Groth, Johnson, Hata, Kravitz,
  Chen, Kalantidis, Li, Shamma, Bernstein and Fei-Fei}]{Krishna2017}
\bibinfo{author}{Krishna, R.}, \bibinfo{author}{Zhu, Y.},
  \bibinfo{author}{Groth, O.}, \bibinfo{author}{Johnson, J.},
  \bibinfo{author}{Hata, K.}, \bibinfo{author}{Kravitz, J.},
  \bibinfo{author}{Chen, S.}, \bibinfo{author}{Kalantidis, Y.},
  \bibinfo{author}{Li, L.J.}, \bibinfo{author}{Shamma, D.A.},
  \bibinfo{author}{Bernstein, M.S.}, \bibinfo{author}{Fei-Fei, L.},
  \bibinfo{year}{2017}.
\newblock \bibinfo{title}{{Visual Genome: Connecting Language and Vision Using
  Crowdsourced Dense Image Annotations}}.
\newblock \bibinfo{journal}{International Journal of Computer Vision}
  \bibinfo{volume}{123}, \bibinfo{pages}{32--73}.
\newblock \DOIprefix\doi{10.1007/s11263-016-0981-7}.
\bibitem[{Krizhevsky(2012)}]{Krizhevsky2012}
\bibinfo{author}{Krizhevsky, A.}, \bibinfo{year}{2012}.
\newblock \bibinfo{title}{{Learning Multiple Layers of Features from Tiny
  Images}}.
\newblock \bibinfo{journal}{University of Toronto} .
\bibitem[{Krizhevsky et~al.(2012)Krizhevsky, Sutskever and
  Hinton.}]{Krizhevsky2012b}
\bibinfo{author}{Krizhevsky, A.}, \bibinfo{author}{Sutskever, I.},
  \bibinfo{author}{Hinton., G.E.}, \bibinfo{year}{2012}.
\newblock \bibinfo{title}{{Imagenet classification with deep convolutional
  neural networks.}}
\newblock \bibinfo{journal}{Advances in neural information processing systems}
  , \bibinfo{pages}{1097--1105}\URLprefix \url{http://arxiv.org/abs/1102.0183}.
\bibitem[{Kumar et~al.(2009)Kumar, Berg, Belhumeur and Nayar}]{Kumar2009}
\bibinfo{author}{Kumar, N.}, \bibinfo{author}{Berg, A.C.},
  \bibinfo{author}{Belhumeur, P.N.}, \bibinfo{author}{Nayar, S.K.},
  \bibinfo{year}{2009}.
\newblock \bibinfo{title}{{Attribute and simile classifiers for face
  verification}}.
\newblock \bibinfo{journal}{Proceedings of the IEEE International Conference on
  Computer Vision} ,
  \bibinfo{pages}{365--372}\DOIprefix\doi{10.1109/ICCV.2009.5459250}.
\bibitem[{Kuznetsova et~al.(2018)Kuznetsova, Rom, Alldrin, Uijlings, Krasin,
  Pont-Tuset, Kamali, Popov, Malloci, Kolesnikov, Duerig and
  Ferrari}]{Kuznetsova2018}
\bibinfo{author}{Kuznetsova, A.}, \bibinfo{author}{Rom, H.},
  \bibinfo{author}{Alldrin, N.}, \bibinfo{author}{Uijlings, J.},
  \bibinfo{author}{Krasin, I.}, \bibinfo{author}{Pont-Tuset, J.},
  \bibinfo{author}{Kamali, S.}, \bibinfo{author}{Popov, S.},
  \bibinfo{author}{Malloci, M.}, \bibinfo{author}{Kolesnikov, A.},
  \bibinfo{author}{Duerig, T.}, \bibinfo{author}{Ferrari, V.},
  \bibinfo{year}{2018}.
\newblock \bibinfo{title}{{The Open Images Dataset V4: Unified image
  classification, object detection, and visual relationship detection at
  scale}} , \bibinfo{pages}{1--20}\URLprefix
  \url{http://arxiv.org/abs/1811.00982}.
\bibitem[{Lam et~al.(2018)Lam, Kuzma, McGee, Dooley, Laielli, Klaric, Bulatov
  and McCord}]{Lam2018}
\bibinfo{author}{Lam, D.}, \bibinfo{author}{Kuzma, R.}, \bibinfo{author}{McGee,
  K.}, \bibinfo{author}{Dooley, S.}, \bibinfo{author}{Laielli, M.},
  \bibinfo{author}{Klaric, M.}, \bibinfo{author}{Bulatov, Y.},
  \bibinfo{author}{McCord, B.}, \bibinfo{year}{2018}.
\newblock \bibinfo{title}{{xView: Objects in Context in Overhead Imagery}}
  \URLprefix \url{http://arxiv.org/abs/1802.07856}.
\bibitem[{Lambert et~al.(2019)Lambert, Petitjean, Dubray and
  Ruan}]{Lambert2019}
\bibinfo{author}{Lambert, Z.}, \bibinfo{author}{Petitjean, C.},
  \bibinfo{author}{Dubray, B.}, \bibinfo{author}{Ruan, S.},
  \bibinfo{year}{2019}.
\newblock \bibinfo{title}{{SegTHOR: Segmentation of Thoracic Organs at Risk in
  CT images}} , \bibinfo{pages}{1--16}.
\bibitem[{LaMontagne et~al.(2019)LaMontagne, Benzinger, Morris, Keefe,
  Hornbeck, Xiong, Grant, Hassenstab, Moulder, Vlassenko, {Raichle, Marcus},
  Carlos and Marcus}]{LaMontagne2019}
\bibinfo{author}{LaMontagne, P.J.}, \bibinfo{author}{Benzinger, T.L.},
  \bibinfo{author}{Morris, J.C.}, \bibinfo{author}{Keefe, S.},
  \bibinfo{author}{Hornbeck, R.}, \bibinfo{author}{Xiong, C.},
  \bibinfo{author}{Grant, E.}, \bibinfo{author}{Hassenstab, J.},
  \bibinfo{author}{Moulder, K.}, \bibinfo{author}{Vlassenko, A.},
  \bibinfo{author}{{Raichle, Marcus}, E.}, \bibinfo{author}{Carlos, C.},
  \bibinfo{author}{Marcus, D.}, \bibinfo{year}{2019}.
\newblock \bibinfo{title}{{OASIS-3: Longitudinal Neuroimaging, Clinical, and
  Cognitive Dataset for Normal Aging and Alzheimer Disease}}.
\newblock \bibinfo{journal}{Journal of Chemical Information and Modeling}
  \bibinfo{volume}{53}, \bibinfo{pages}{1689--1699}.
\newblock \DOIprefix\doi{10.1017/CBO9781107415324.004}.
\bibitem[{Lateef and Ruichek(2019)}]{LATEEF2019321}
\bibinfo{author}{Lateef, F.}, \bibinfo{author}{Ruichek, Y.},
  \bibinfo{year}{2019}.
\newblock \bibinfo{title}{Survey on semantic segmentation using deep learning
  techniques}.
\newblock \bibinfo{journal}{Neurocomputing} \bibinfo{volume}{338},
  \bibinfo{pages}{321--348}.
\newblock \URLprefix
  \url{https://www.sciencedirect.com/science/article/pii/S092523121930181X},
  \DOIprefix\doi{https://doi.org/10.1016/j.neucom.2019.02.003}.
\bibitem[{Lazebnik et~al.(2006)Lazebnik, Schmid and Ponce}]{Lazebnik2006}
\bibinfo{author}{Lazebnik, S.}, \bibinfo{author}{Schmid, C.},
  \bibinfo{author}{Ponce, J.}, \bibinfo{year}{2006}.
\newblock \bibinfo{title}{{Beyond bags of features: Spatial pyramid matching
  for recognizing natural scene categories}}.
\newblock \bibinfo{journal}{Proceedings of the IEEE Computer Society Conference
  on Computer Vision and Pattern Recognition} \bibinfo{volume}{2},
  \bibinfo{pages}{2169--2178}.
\newblock \DOIprefix\doi{10.1109/CVPR.2006.68}.
\bibitem[{Le et~al.(2019)Le, Nguyen, Nie, Tran and Sugimoto}]{CAMO}
\bibinfo{author}{Le, T.N.}, \bibinfo{author}{Nguyen, T.}, \bibinfo{author}{Nie,
  Z.}, \bibinfo{author}{Tran, M.T.}, \bibinfo{author}{Sugimoto, A.},
  \bibinfo{year}{2019}.
\newblock \bibinfo{title}{Anabranch network for camouflaged object
  segmentation}.
\newblock \bibinfo{journal}{Computer Vision and Image Understanding}
  \bibinfo{volume}{184}.
\newblock \DOIprefix\doi{10.1016/j.cviu.2019.04.006}.
\bibitem[{Lecun et~al.(1989)Lecun, Boser, Denker, Henderson, Howard, Hubbard
  and Jackel}]{Lecun1989}
\bibinfo{author}{Lecun, Y.}, \bibinfo{author}{Boser, B.},
  \bibinfo{author}{Denker, J.S.}, \bibinfo{author}{Henderson, D.},
  \bibinfo{author}{Howard, R.E.}, \bibinfo{author}{Hubbard, W.},
  \bibinfo{author}{Jackel, L.D.}, \bibinfo{year}{1989}.
\newblock \bibinfo{title}{{Backpropagation applied to handwritten zip code
  recognition}}.
\newblock \bibinfo{journal}{Neural computation} \bibinfo{volume}{1},
  \bibinfo{pages}{541--551}.
\bibitem[{Lecun et~al.(1990)Lecun, Boser, Denker, Henderson, Howard, Hubbard
  and Jackel}]{Lecun1990}
\bibinfo{author}{Lecun, Y.}, \bibinfo{author}{Boser, B.E.},
  \bibinfo{author}{Denker, J.S.}, \bibinfo{author}{Henderson, D.},
  \bibinfo{author}{Howard, R.E.}, \bibinfo{author}{Hubbard, W.E.},
  \bibinfo{author}{Jackel, L.D.}, \bibinfo{year}{1990}.
\newblock \bibinfo{title}{{Handwritten digit recognition with a
  back-propagation network}}, in: \bibinfo{booktitle}{Advances in neural
  information processing systems}, pp. \bibinfo{pages}{396--404}.
\bibitem[{Lecun et~al.(1998)Lecun, Bottou, Bengio and Ha}]{Lecun1998}
\bibinfo{author}{Lecun, Y.}, \bibinfo{author}{Bottou, L.},
  \bibinfo{author}{Bengio, Y.}, \bibinfo{author}{Ha, P.}, \bibinfo{year}{1998}.
\newblock \bibinfo{title}{{LeNet}}.
\newblock \bibinfo{journal}{Proceedings of the IEEE} ,
  \bibinfo{pages}{1--46}\DOIprefix\doi{10.1109/5.726791}.
\bibitem[{Lecun and Others(1997)}]{Lecun1997}
\bibinfo{author}{Lecun, Y.}, \bibinfo{author}{Others}, \bibinfo{year}{1997}.
\newblock \bibinfo{title}{{Handwritten Digit Recognition with a
  Back-Propagation Network}}.
\newblock \bibinfo{journal}{Neural Information Processing Systems}
  \bibinfo{volume}{2}.
\bibitem[{{LERA}(2018)}]{StanfordUniversity}
\bibinfo{author}{{LERA}}, \bibinfo{year}{2018}.
\newblock \bibinfo{title}{{LERA- Lower Extremity RAdiographs}}.
\newblock \URLprefix
  \url{https://aimi.stanford.edu/lera-lower-extremity-radiographs-2}.
\bibitem[{Li et~al.(2013)Li, Kim, Humayun, Tsai and Rehg}]{6751383}
\bibinfo{author}{Li, F.}, \bibinfo{author}{Kim, T.}, \bibinfo{author}{Humayun,
  A.}, \bibinfo{author}{Tsai, D.}, \bibinfo{author}{Rehg, J.M.},
  \bibinfo{year}{2013}.
\newblock \bibinfo{title}{{Video Segmentation by Tracking Many Figure-Ground
  Segments}}, in: \bibinfo{booktitle}{2013 IEEE International Conference on
  Computer Vision}, pp. \bibinfo{pages}{2192--2199}.
\newblock \DOIprefix\doi{10.1109/ICCV.2013.273}.
\bibitem[{Li and Yu(2015)}]{HKU}
\bibinfo{author}{Li, G.}, \bibinfo{author}{Yu, Y.}, \bibinfo{year}{2015}.
\newblock \bibinfo{title}{Visual saliency based on multiscale deep features}
  \DOIprefix\doi{10.1109/CVPR.2015.7299184}.
\bibitem[{Li and Chen(2020)}]{Li2020}
\bibinfo{author}{Li, H.}, \bibinfo{author}{Chen, M.}, \bibinfo{year}{2020}.
\newblock \bibinfo{title}{{Automatic Structure Segmentation for Radiotherapy
  Planning Challenge 2020}}.
\newblock \DOIprefix\doi{10.5281/zenodo.3718885}.
\bibitem[{Li et~al.(2019)Li, Zhou, Deng, Chen, SenseTime, YINO and {Zhejiang
  Cancer Hospital}}]{Li2019a}
\bibinfo{author}{Li, H.}, \bibinfo{author}{Zhou, J.}, \bibinfo{author}{Deng,
  J.}, \bibinfo{author}{Chen, M.}, \bibinfo{author}{SenseTime},
  \bibinfo{author}{YINO}, \bibinfo{author}{{Zhejiang Cancer Hospital}},
  \bibinfo{year}{2019}.
\newblock \bibinfo{title}{{StructSeg 2019}}.
\bibitem[{Li et~al.(2014a)Li, Zang, Zhang, Li and Wu}]{MiaoRemote}
\bibinfo{author}{Li, M.}, \bibinfo{author}{Zang, S.}, \bibinfo{author}{Zhang,
  B.}, \bibinfo{author}{Li, S.}, \bibinfo{author}{Wu, C.},
  \bibinfo{year}{2014}a.
\newblock \bibinfo{title}{A review of remote sensing image classification
  techniques: the role of spatio-contextual information}.
\newblock \bibinfo{journal}{European Journal of Remote Sensing}
  \bibinfo{volume}{47}, \bibinfo{pages}{389--411}.
\newblock \URLprefix \url{https://doi.org/10.5721/EuJRS20144723},
  \DOIprefix\doi{10.5721/EuJRS20144723}.
\bibitem[{Li and Wang(2019)}]{Li2019}
\bibinfo{author}{Li, S.}, \bibinfo{author}{Wang}, \bibinfo{year}{2019}.
\newblock \bibinfo{title}{{AASCE}}.
\newblock \URLprefix \url{https://aasce19.grand-challenge.org/}.
\bibitem[{Li et~al.(2017)Li, Yang, Cheng, Chen, Guo and Chen}]{LiSalient2}
\bibinfo{author}{Li, X.}, \bibinfo{author}{Yang, F.}, \bibinfo{author}{Cheng,
  H.}, \bibinfo{author}{Chen, J.}, \bibinfo{author}{Guo, Y.},
  \bibinfo{author}{Chen, L.}, \bibinfo{year}{2017}.
\newblock \bibinfo{title}{Multi-scale cascade network for salient object
  detection} ,
  \bibinfo{pages}{439--447}\DOIprefix\doi{10.1145/3123266.3123290}.
\bibitem[{Li et~al.(2018)Li, Yang, Cheng, Liu and Shen}]{LiSalient}
\bibinfo{author}{Li, X.}, \bibinfo{author}{Yang, F.}, \bibinfo{author}{Cheng,
  H.}, \bibinfo{author}{Liu, W.}, \bibinfo{author}{Shen, D.},
  \bibinfo{year}{2018}.
\newblock \bibinfo{title}{Contour knowledge transfer for salient object
  detection: 15th european conference, munich, germany, september 8-14, 2018,
  proceedings, part xv} ,
  \bibinfo{pages}{370--385}\DOIprefix\doi{10.1007/978-3-030-01267-0_22}.
\bibitem[{Li et~al.(2014b)Li, Hou, Koch, Rehg and Yuille}]{PASCAL-S}
\bibinfo{author}{Li, Y.}, \bibinfo{author}{Hou, X.}, \bibinfo{author}{Koch,
  C.}, \bibinfo{author}{Rehg, J.}, \bibinfo{author}{Yuille, A.},
  \bibinfo{year}{2014}b.
\newblock \bibinfo{title}{The secrets of salient object segmentation}.
\newblock \bibinfo{journal}{Proceedings of the IEEE Computer Society Conference
  on Computer Vision and Pattern Recognition}
  \DOIprefix\doi{10.1109/CVPR.2014.43}.
\bibitem[{Lin et~al.(2016)Lin, Dollár, Girshick, He, Hariharan and
  Belongie}]{LinFeature}
\bibinfo{author}{Lin, T.Y.}, \bibinfo{author}{Dollár, P.},
  \bibinfo{author}{Girshick, R.}, \bibinfo{author}{He, K.},
  \bibinfo{author}{Hariharan, B.}, \bibinfo{author}{Belongie, S.},
  \bibinfo{year}{2016}.
\newblock \bibinfo{title}{Feature pyramid networks for object detection} .
\bibitem[{Lin et~al.(2014)Lin, Maire, Belongie, Hays, Perona, Ramanan,
  Doll{\'{a}}r and Zitnick}]{Lin2014}
\bibinfo{author}{Lin, T.Y.}, \bibinfo{author}{Maire, M.},
  \bibinfo{author}{Belongie, S.}, \bibinfo{author}{Hays, J.},
  \bibinfo{author}{Perona, P.}, \bibinfo{author}{Ramanan, D.},
  \bibinfo{author}{Doll{\'{a}}r, P.}, \bibinfo{author}{Zitnick, C.L.},
  \bibinfo{year}{2014}.
\newblock \bibinfo{title}{{Microsoft COCO: Common objects in context}}.
\newblock \bibinfo{journal}{Lecture Notes in Computer Science (including
  subseries Lecture Notes in Artificial Intelligence and Lecture Notes in
  Bioinformatics)} \bibinfo{volume}{8693 LNCS}, \bibinfo{pages}{740--755}.
\newblock \DOIprefix\doi{10.1007/978-3-319-10602-1_48}.
\bibitem[{Liu et~al.(2015a)Liu, Yuen and Torralba}]{Liu2015b}
\bibinfo{author}{Liu, C.}, \bibinfo{author}{Yuen, J.},
  \bibinfo{author}{Torralba, A.}, \bibinfo{year}{2015}a.
\newblock \bibinfo{title}{{Nonparametric scene parsing via label transfer}}.
\newblock \bibinfo{journal}{Dense Image Correspondences for Computer Vision}
  \bibinfo{volume}{33}, \bibinfo{pages}{207--236}.
\newblock \DOIprefix\doi{10.1007/978-3-319-23048-1_10}.
\bibitem[{Liu and Mattyus(2015)}]{Liu2015}
\bibinfo{author}{Liu, K.}, \bibinfo{author}{Mattyus, G.}, \bibinfo{year}{2015}.
\newblock \bibinfo{title}{{Fast Multiclass Vehicle Detection on Aerial
  Images}}.
\newblock \bibinfo{journal}{IEEE Geoscience and Remote Sensing Letters}
  \bibinfo{volume}{12}, \bibinfo{pages}{1938--1942}.
\newblock \DOIprefix\doi{10.1109/LGRS.2015.2439517}.
\bibitem[{Liu et~al.(2020a)Liu, Ouyang, Wang, Fieguth, Chen, Liu and
  Pietik{\"{a}}inen}]{Liu2020}
\bibinfo{author}{Liu, L.}, \bibinfo{author}{Ouyang, W.}, \bibinfo{author}{Wang,
  X.}, \bibinfo{author}{Fieguth, P.}, \bibinfo{author}{Chen, J.},
  \bibinfo{author}{Liu, X.}, \bibinfo{author}{Pietik{\"{a}}inen, M.},
  \bibinfo{year}{2020}a.
\newblock \bibinfo{title}{{Deep Learning for Generic Object Detection: A
  Survey}}.
\newblock \bibinfo{journal}{International Journal of Computer Vision}
  \bibinfo{volume}{128}, \bibinfo{pages}{261--318}.
\newblock \URLprefix \url{https://doi.org/10.1007/s11263-019-01247-4},
  \DOIprefix\doi{10.1007/s11263-019-01247-4}.
\bibitem[{Liu et~al.(2020b)Liu, Ouyang, Wang, Fieguth, Chen, Liu and
  Pietikinen}]{Liu20200}
\bibinfo{author}{Liu, L.}, \bibinfo{author}{Ouyang, W.}, \bibinfo{author}{Wang,
  X.}, \bibinfo{author}{Fieguth, P.}, \bibinfo{author}{Chen, J.},
  \bibinfo{author}{Liu, X.}, \bibinfo{author}{Pietikinen, M.},
  \bibinfo{year}{2020}b.
\newblock \bibinfo{title}{Deep learning for generic object detection: A
  survey}.
\newblock \bibinfo{journal}{International Journal of Computer Vision}
  \bibinfo{volume}{128}, \bibinfo{pages}{261--318}.
\newblock \URLprefix \url{https://doi.org/10.1007/s11263-019-01247-4},
  \DOIprefix\doi{10.1007/s11263-019-01247-4}.
\bibitem[{Liu et~al.(2007)Liu, Sun, Zheng, Tang and Shum}]{LiuSalient}
\bibinfo{author}{Liu, T.}, \bibinfo{author}{Sun, J.}, \bibinfo{author}{Zheng,
  N.N.}, \bibinfo{author}{Tang, X.}, \bibinfo{author}{Shum, H.Y.},
  \bibinfo{year}{2007}.
\newblock \bibinfo{title}{Learning to detect a salient object} ,
  \bibinfo{pages}{1--8}\DOIprefix\doi{10.1109/CVPR.2007.383047}.
\bibitem[{Liu et~al.(2015b)Liu, Anguelov, Erhan, Szegedy, Reed, Fu and
  Berg}]{Liu2015c}
\bibinfo{author}{Liu, W.}, \bibinfo{author}{Anguelov, D.},
  \bibinfo{author}{Erhan, D.}, \bibinfo{author}{Szegedy, C.},
  \bibinfo{author}{Reed, S.}, \bibinfo{author}{Fu, C.Y.},
  \bibinfo{author}{Berg, A.C.}, \bibinfo{year}{2015}b.
\newblock \bibinfo{title}{{SSD: Single Shot MultiBox Detector}}
  \DOIprefix\doi{10.1007/978-3-319-46448-0_2}.
\bibitem[{Liu et~al.(2016)Liu, Luo, Qiu, Wang and Tang}]{Liu2016}
\bibinfo{author}{Liu, Z.}, \bibinfo{author}{Luo, P.}, \bibinfo{author}{Qiu,
  S.}, \bibinfo{author}{Wang, X.}, \bibinfo{author}{Tang, X.},
  \bibinfo{year}{2016}.
\newblock \bibinfo{title}{{DeepFashion: Powering Robust Clothes Recognition and
  Retrieval with Rich Annotations}}, in: \bibinfo{booktitle}{Proceedings of the
  IEEE Computer Society Conference on Computer Vision and Pattern Recognition},
  pp. \bibinfo{pages}{1096--1104}.
\newblock \DOIprefix\doi{10.1109/CVPR.2016.124}.
\bibitem[{Liu et~al.(2015c)Liu, Luo, Wang and Tang}]{Liu2015a}
\bibinfo{author}{Liu, Z.}, \bibinfo{author}{Luo, P.}, \bibinfo{author}{Wang,
  X.}, \bibinfo{author}{Tang, X.}, \bibinfo{year}{2015}c.
\newblock \bibinfo{title}{{Deep learning face attributes in the wild}}.
\newblock \bibinfo{journal}{Proceedings of the IEEE International Conference on
  Computer Vision} \bibinfo{volume}{2015 Inter}, \bibinfo{pages}{3730--3738}.
\newblock \DOIprefix\doi{10.1109/ICCV.2015.425}.
\bibitem[{Lowe(1999)}]{Lowe1999}
\bibinfo{author}{Lowe, D.G.}, \bibinfo{year}{1999}.
\newblock \bibinfo{title}{{Object recognition from local scale-invariant
  features}}, in: \bibinfo{booktitle}{Proceedings of the IEEE International
  Conference on Computer Vision}, \bibinfo{publisher}{IEEE}. pp.
  \bibinfo{pages}{1150--1157}.
\newblock \DOIprefix\doi{10.1109/iccv.1999.790410}.
\bibitem[{Lyft(2019)}]{Lyft2019}
\bibinfo{author}{Lyft}, \bibinfo{year}{2019}.
\newblock \bibinfo{title}{{Lyft 3D Object Detection for Autonomous Vehicles}}.
\newblock \URLprefix
  \url{https://www.kaggle.com/c/3d-object-detection-for-autonomous-vehicles}.
\bibitem[{Maddern et~al.()Maddern, Pascoe, Linegar and Newman}]{Maddern}
\bibinfo{author}{Maddern, W.}, \bibinfo{author}{Pascoe, G.},
  \bibinfo{author}{Linegar, C.}, \bibinfo{author}{Newman, P.}, .
\newblock \bibinfo{title}{{1 Year , 1000km : The Oxford RobotCar Dataset}}
  \bibinfo{volume}{3}.
\bibitem[{Maier(2015)}]{Sicas}
\bibinfo{author}{Maier, O.}, \bibinfo{year}{2015}.
\newblock \bibinfo{title}{{SMIR Database}} \URLprefix
  \url{https://www.smir.ch}.
\bibitem[{Martin et~al.(2001)Martin, Fowlkes, Tal and Malik}]{Martin2001}
\bibinfo{author}{Martin, D.}, \bibinfo{author}{Fowlkes, C.},
  \bibinfo{author}{Tal, D.}, \bibinfo{author}{Malik, J.}, \bibinfo{year}{2001}.
\newblock \bibinfo{title}{{A database of human segmented natural images and its
  application to evaluating segmentation algorithms and measuring ecological
  statistics}}.
\newblock \bibinfo{journal}{Proceedings of the IEEE International Conference on
  Computer Vision} \bibinfo{volume}{2}, \bibinfo{pages}{416--423}.
\newblock \DOIprefix\doi{10.1109/ICCV.2001.937655}.
\bibitem[{Martinez(1998)}]{Martinez1998}
\bibinfo{author}{Martinez, A.M.}, \bibinfo{year}{1998}.
\newblock \bibinfo{title}{{The AR face database}}.
\newblock \bibinfo{journal}{CVC Technical Report24} .
\bibitem[{Masi et~al.(2019)Masi, Wu, Hassner and Natarajan}]{Masi2019}
\bibinfo{author}{Masi, I.}, \bibinfo{author}{Wu, Y.}, \bibinfo{author}{Hassner,
  T.}, \bibinfo{author}{Natarajan, P.}, \bibinfo{year}{2019}.
\newblock \bibinfo{title}{{Deep Face Recognition: A Survey}}.
\newblock \bibinfo{journal}{Proceedings - 31st Conference on Graphics, Patterns
  and Images, SIBGRAPI 2018} ,
  \bibinfo{pages}{471--478}\DOIprefix\doi{10.1109/SIBGRAPI.2018.00067}.
\bibitem[{Maze et~al.(2018)Maze, Adams, Duncan, Kalka, Miller, Otto, Jain,
  Niggel, Anderson, Cheney and Grother}]{Maze2018}
\bibinfo{author}{Maze, B.}, \bibinfo{author}{Adams, J.},
  \bibinfo{author}{Duncan, J.A.}, \bibinfo{author}{Kalka, N.},
  \bibinfo{author}{Miller, T.}, \bibinfo{author}{Otto, C.},
  \bibinfo{author}{Jain, A.K.}, \bibinfo{author}{Niggel, W.T.},
  \bibinfo{author}{Anderson, J.}, \bibinfo{author}{Cheney, J.},
  \bibinfo{author}{Grother, P.}, \bibinfo{year}{2018}.
\newblock \bibinfo{title}{{IARPA janus benchmark-C: Face dataset and
  protocol}}.
\newblock \bibinfo{journal}{Proceedings - 2018 International Conference on
  Biometrics, ICB 2018} ,
  \bibinfo{pages}{158--165}\DOIprefix\doi{10.1109/ICB2018.2018.00033}.
\bibitem[{Merler et~al.(2019)Merler, Ratha, Feris and Smith}]{Merler2019}
\bibinfo{author}{Merler, M.}, \bibinfo{author}{Ratha, N.},
  \bibinfo{author}{Feris, R.S.}, \bibinfo{author}{Smith, J.R.},
  \bibinfo{year}{2019}.
\newblock \bibinfo{title}{{Diversity in Faces}} ,
  \bibinfo{pages}{1--29}\URLprefix \url{http://arxiv.org/abs/1901.10436}.
\bibitem[{Meyer and Kuschk(2019)}]{Meyer2019}
\bibinfo{author}{Meyer, M.}, \bibinfo{author}{Kuschk, G.},
  \bibinfo{year}{2019}.
\newblock \bibinfo{title}{{Automotive radar dataset for deep learning based 3D
  object detection}}.
\newblock \bibinfo{journal}{EuRAD 2019 - 2019 16th European Radar Conference} ,
  \bibinfo{pages}{129--132}.
\bibitem[{Mottaghi et~al.(2014)Mottaghi, Chen, Liu, Cho, Lee, Fidler, Urtasun
  and Yuille}]{Mottaghi2014}
\bibinfo{author}{Mottaghi, R.}, \bibinfo{author}{Chen, X.},
  \bibinfo{author}{Liu, X.}, \bibinfo{author}{Cho, N.G.}, \bibinfo{author}{Lee,
  S.W.}, \bibinfo{author}{Fidler, S.}, \bibinfo{author}{Urtasun, R.},
  \bibinfo{author}{Yuille, A.}, \bibinfo{year}{2014}.
\newblock \bibinfo{title}{{The role of context for object detection and
  semantic segmentation in the wild}}.
\newblock \bibinfo{journal}{Proceedings of the IEEE Computer Society Conference
  on Computer Vision and Pattern Recognition} ,
  \bibinfo{pages}{891--898}\DOIprefix\doi{10.1109/CVPR.2014.119}.
\bibitem[{Mundhenk et~al.(2016)Mundhenk, Konjevod, Sakla and
  Boakye}]{Mundhenk2016}
\bibinfo{author}{Mundhenk, T.N.}, \bibinfo{author}{Konjevod, G.},
  \bibinfo{author}{Sakla, W.A.}, \bibinfo{author}{Boakye, K.},
  \bibinfo{year}{2016}.
\newblock \bibinfo{title}{{A Large Contextual Dataset for Classification,
  Detection and Counting of Cars with Deep Learning}}.
\newblock \bibinfo{journal}{Lecture Notes in Computer Science (including
  subseries Lecture Notes in Artificial Intelligence and Lecture Notes in
  Bioinformatics)} \bibinfo{volume}{9907 LNCS}, \bibinfo{pages}{785--800}.
\newblock \URLprefix \url{http://arxiv.org/abs/1609.04453}.
\bibitem[{{National Library of
  Medicine}(2006)}]{UniformedServicesUniversity2006}
\bibinfo{author}{{National Library of Medicine}}, \bibinfo{year}{2006}.
\newblock \bibinfo{title}{{MedPix}}.
\newblock \URLprefix \url{https://medpix.nlm.nih.gov/home}.
\bibitem[{Nech et~al.()Nech, Kemelmacher-Shlizerman and Allen}]{Nech}
\bibinfo{author}{Nech, A.}, \bibinfo{author}{Kemelmacher-Shlizerman, I.},
  \bibinfo{author}{Allen, P.G.}, .
\newblock \bibinfo{title}{{Level Playing Field for Million Scale Face
  Recognition}}.
\newblock \bibinfo{type}{Technical Report}.
\bibitem[{Nene et~al.(1996a)Nene, Nayar and Murase}]{Nene1996a}
\bibinfo{author}{Nene, S.}, \bibinfo{author}{Nayar, S.},
  \bibinfo{author}{Murase, H.}, \bibinfo{year}{1996}a.
\newblock \bibinfo{title}{{Columbia Object Image Library (COIL-100)}}.
\newblock \bibinfo{journal}{Technical Report} \bibinfo{volume}{95},
  \bibinfo{pages}{223--303}.
\newblock \URLprefix
  \url{http://citeseerx.ist.psu.edu/viewdoc/summary?doi=10.1.1.54.5914}.
\bibitem[{Nene et~al.(1996b)Nene, Nayar and Murase}]{Nene1996}
\bibinfo{author}{Nene, S.}, \bibinfo{author}{Nayar, S.},
  \bibinfo{author}{Murase, H.}, \bibinfo{year}{1996}b.
\newblock \bibinfo{title}{{Columbia Object Image Library (COIL-20)}}.
\newblock \bibinfo{journal}{Technical Report} \bibinfo{volume}{95},
  \bibinfo{pages}{223--303}.
\newblock \URLprefix
  \url{http://citeseerx.ist.psu.edu/viewdoc/summary?doi=10.1.1.54.5914}.
\bibitem[{Neuhold et~al.(2017)Neuhold, Ollmann, Bulo and
  Kontschieder}]{Neuhold2017}
\bibinfo{author}{Neuhold, G.}, \bibinfo{author}{Ollmann, T.},
  \bibinfo{author}{Bulo, S.R.}, \bibinfo{author}{Kontschieder, P.},
  \bibinfo{year}{2017}.
\newblock \bibinfo{title}{{The Mapillary Vistas Dataset for Semantic
  Understanding of Street Scenes}}.
\newblock \bibinfo{journal}{Proceedings of the IEEE International Conference on
  Computer Vision} \bibinfo{volume}{2017-Octob}, \bibinfo{pages}{5000--5009}.
\newblock \DOIprefix\doi{10.1109/ICCV.2017.534}.
\bibitem[{Neumann et~al.(2019)Neumann, Karg, Zhang, Scharfenberger, Piegert,
  Mistr, Prokofyeva, Thiel, Vedaldi, Zisserman and Schiele}]{Neumann2019}
\bibinfo{author}{Neumann, L.}, \bibinfo{author}{Karg, M.},
  \bibinfo{author}{Zhang, S.}, \bibinfo{author}{Scharfenberger, C.},
  \bibinfo{author}{Piegert, E.}, \bibinfo{author}{Mistr, S.},
  \bibinfo{author}{Prokofyeva, O.}, \bibinfo{author}{Thiel, R.},
  \bibinfo{author}{Vedaldi, A.}, \bibinfo{author}{Zisserman, A.},
  \bibinfo{author}{Schiele, B.}, \bibinfo{year}{2019}.
\newblock \bibinfo{title}{{NightOwls: A Pedestrians at Night Dataset}}, in:
  \bibinfo{booktitle}{Lecture Notes in Computer Science (including subseries
  Lecture Notes in Artificial Intelligence and Lecture Notes in
  Bioinformatics)}, \bibinfo{publisher}{Springer Verlag}. pp.
  \bibinfo{pages}{691--705}.
\newblock \URLprefix \url{http://www.nightowls-dataset.org/},
  \DOIprefix\doi{10.1007/978-3-030-20887-5_43}.
\bibitem[{Nilsback and Zisserman(2008)}]{Nilsback2008}
\bibinfo{author}{Nilsback, M.E.}, \bibinfo{author}{Zisserman, A.},
  \bibinfo{year}{2008}.
\newblock \bibinfo{title}{{Automated flower classification over a large number
  of classes}}.
\newblock \bibinfo{journal}{Proceedings - 6th Indian Conference on Computer
  Vision, Graphics and Image Processing, ICVGIP 2008} ,
  \bibinfo{pages}{722--729}\DOIprefix\doi{10.1109/ICVGIP.2008.47}.
\bibitem[{Ochs et~al.(2014)Ochs, Malik and Brox}]{6682905}
\bibinfo{author}{Ochs, P.}, \bibinfo{author}{Malik, J.}, \bibinfo{author}{Brox,
  T.}, \bibinfo{year}{2014}.
\newblock \bibinfo{title}{{Segmentation of Moving Objects by Long Term Video
  Analysis}}.
\newblock \bibinfo{journal}{IEEE Transactions on Pattern Analysis and Machine
  Intelligence} \bibinfo{volume}{36}, \bibinfo{pages}{1187--1200}.
\newblock \DOIprefix\doi{10.1109/TPAMI.2013.242}.
\bibitem[{{Odir}(2019)}]{PekingUniversity2019a}
\bibinfo{author}{{Odir}}, \bibinfo{year}{2019}.
\newblock \bibinfo{title}{{ODIR-5K}}.
\newblock \URLprefix
  \url{http://www.kaggle.com/andrewmvd/ocular-disease-recognition-odir5k}.
\bibitem[{Orlando et~al.(2019)Orlando, Fu, Breda, van Keer, Bathula,
  Diaz{-}Pinto, Fang, Heng, Kim, Lee, Lee, Li, Liu, Lu, Murugesan, Naranjo,
  Phaye, Shankaranarayana, Sikka, Son, van~den Hengel, Wang, Wu, Wu, Xu, Xu,
  Yin, Li, Zhang, Xu and Bogunovic}]{Fu}
\bibinfo{author}{Orlando, J.I.}, \bibinfo{author}{Fu, H.},
  \bibinfo{author}{Breda, J.B.}, \bibinfo{author}{van Keer, K.},
  \bibinfo{author}{Bathula, D.R.}, \bibinfo{author}{Diaz{-}Pinto, A.},
  \bibinfo{author}{Fang, R.}, \bibinfo{author}{Heng, P.}, \bibinfo{author}{Kim,
  J.}, \bibinfo{author}{Lee, J.}, \bibinfo{author}{Lee, J.},
  \bibinfo{author}{Li, X.}, \bibinfo{author}{Liu, P.}, \bibinfo{author}{Lu,
  S.}, \bibinfo{author}{Murugesan, B.}, \bibinfo{author}{Naranjo, V.},
  \bibinfo{author}{Phaye, S.S.R.}, \bibinfo{author}{Shankaranarayana, S.M.},
  \bibinfo{author}{Sikka, A.}, \bibinfo{author}{Son, J.},
  \bibinfo{author}{van~den Hengel, A.}, \bibinfo{author}{Wang, S.},
  \bibinfo{author}{Wu, J.}, \bibinfo{author}{Wu, Z.}, \bibinfo{author}{Xu, G.},
  \bibinfo{author}{Xu, Y.}, \bibinfo{author}{Yin, P.}, \bibinfo{author}{Li,
  F.}, \bibinfo{author}{Zhang, X.}, \bibinfo{author}{Xu, Y.},
  \bibinfo{author}{Bogunovic, H.}, \bibinfo{year}{2019}.
\newblock \bibinfo{title}{{REFUGE} challenge: {A} unified framework for
  evaluating automated methods for glaucoma assessment from fundus
  photographs}.
\newblock \bibinfo{journal}{CoRR} \bibinfo{volume}{abs/1910.03667}.
\bibitem[{Osuna et~al.(1997)Osuna, Freund and Girosi}]{Osuna1997}
\bibinfo{author}{Osuna, E.}, \bibinfo{author}{Freund, R.},
  \bibinfo{author}{Girosi, F.}, \bibinfo{year}{1997}.
\newblock \bibinfo{title}{{Training support vector machines: An application to
  face detection}}.
\newblock \bibinfo{journal}{Proceedings of the IEEE Computer Society Conference
  on Computer Vision and Pattern Recognition} ,
  \bibinfo{pages}{130--136}\DOIprefix\doi{10.1109/cvpr.1997.609310}.
\bibitem[{Papageorgiou and Poggio(2000)}]{Papageorgiou2000}
\bibinfo{author}{Papageorgiou, C.}, \bibinfo{author}{Poggio, T.},
  \bibinfo{year}{2000}.
\newblock \bibinfo{title}{{Trainable system for object detection}}.
\newblock \bibinfo{journal}{International Journal of Computer Vision}
  \bibinfo{volume}{38}, \bibinfo{pages}{15--33}.
\newblock \DOIprefix\doi{10.1023/A:1008162616689}.
\bibitem[{Parkhi et~al.(2015)Parkhi, Vedaldi and Zisserman}]{Parkhi2015}
\bibinfo{author}{Parkhi, O.M.}, \bibinfo{author}{Vedaldi, A.},
  \bibinfo{author}{Zisserman, A.}, \bibinfo{year}{2015}.
\newblock \bibinfo{title}{{Deep Face Recognition}} ,
  \bibinfo{pages}{41.1--41.12}\DOIprefix\doi{10.5244/c.29.41}.
\bibitem[{Patil et~al.(2019)Patil, Malla, Gang and Chen}]{Patil2019}
\bibinfo{author}{Patil, A.}, \bibinfo{author}{Malla, S.},
  \bibinfo{author}{Gang, H.}, \bibinfo{author}{Chen, Y.T.},
  \bibinfo{year}{2019}.
\newblock \bibinfo{title}{{The H3D dataset for full-surround 3D multi-object
  detection and tracking in crowded urban scenes}}.
\newblock \bibinfo{journal}{Proceedings - IEEE International Conference on
  Robotics and Automation} \bibinfo{volume}{2019-May},
  \bibinfo{pages}{9552--9557}.
\newblock \DOIprefix\doi{10.1109/ICRA.2019.8793925}.
\bibitem[{Patterson and Hays(2012)}]{Patterson2012}
\bibinfo{author}{Patterson, G.}, \bibinfo{author}{Hays, J.},
  \bibinfo{year}{2012}.
\newblock \bibinfo{title}{{SUN attribute database: Discovering, annotating, and
  recognizing scene attributes}}.
\newblock \bibinfo{journal}{Proceedings of the IEEE Computer Society Conference
  on Computer Vision and Pattern Recognition} ,
  \bibinfo{pages}{2751--2758}\DOIprefix\doi{10.1109/CVPR.2012.6247998}.
\bibitem[{Pham et~al.(2019)Pham, Sevestre, Pahwa, Zhan, Pang, Chen, Mustafa,
  Chandrasekhar and Lin}]{Pham2019}
\bibinfo{author}{Pham, Q.H.}, \bibinfo{author}{Sevestre, P.},
  \bibinfo{author}{Pahwa, R.S.}, \bibinfo{author}{Zhan, H.},
  \bibinfo{author}{Pang, C.H.}, \bibinfo{author}{Chen, Y.},
  \bibinfo{author}{Mustafa, A.}, \bibinfo{author}{Chandrasekhar, V.},
  \bibinfo{author}{Lin, J.}, \bibinfo{year}{2019}.
\newblock \bibinfo{title}{{A*3D Dataset: Towards Autonomous Driving in
  Challenging Environments}} .
\bibitem[{Phillips et~al.(1998)Phillips, Wechsler, Huang and
  Rauss}]{Phillips1998}
\bibinfo{author}{Phillips, P.J.}, \bibinfo{author}{Wechsler, H.},
  \bibinfo{author}{Huang, J.}, \bibinfo{author}{Rauss, P.J.},
  \bibinfo{year}{1998}.
\newblock \bibinfo{title}{{The FERET database and evaluation procedure for
  face-recognition algorithms}}.
\newblock \bibinfo{journal}{Image and Vision Computing} \bibinfo{volume}{16},
  \bibinfo{pages}{295--306}.
\newblock \DOIprefix\doi{10.1016/s0262-8856(97)00070-x}.
\bibitem[{Prest et~al.(2012)Prest, Leistner, Civera, Schmid and
  Ferrari}]{Prest2012}
\bibinfo{author}{Prest, A.}, \bibinfo{author}{Leistner, C.},
  \bibinfo{author}{Civera, J.}, \bibinfo{author}{Schmid, C.},
  \bibinfo{author}{Ferrari, V.}, \bibinfo{year}{2012}.
\newblock \bibinfo{title}{{Learning object class detectors from weakly
  annotated video}}.
\newblock \bibinfo{journal}{Proceedings of the IEEE Computer Society Conference
  on Computer Vision and Pattern Recognition} ,
  \bibinfo{pages}{3282--3289}\DOIprefix\doi{10.1109/CVPR.2012.6248065}.
\bibitem[{Quattoni and Torralba(2010)}]{Quattoni2010}
\bibinfo{author}{Quattoni, A.}, \bibinfo{author}{Torralba, A.},
  \bibinfo{year}{2010}.
\newblock \bibinfo{title}{{Recognizing indoor scenes}}.
\newblock \bibinfo{journal}{2009 IEEE Conference on Computer Vision and Pattern
  Recognition} ,
  \bibinfo{pages}{413--420}\DOIprefix\doi{10.1109/cvpr.2009.5206537}.
\bibitem[{{Radiological Society of North
  America}(2019)}]{RadiologicalSocietyofNorthAmerica2019}
\bibinfo{author}{{Radiological Society of North America}},
  \bibinfo{year}{2019}.
\newblock \bibinfo{title}{{RSNA Intracranial Hemorrhage Detection}}.
\bibitem[{Rajpurkar et~al.(2017)Rajpurkar, Irvin, Bagul, Ding, Duan, Mehta,
  Yang, Zhu, Laird, Ball, Langlotz, Shpanskaya, Lungren and Ng}]{Rajpurkar2017}
\bibinfo{author}{Rajpurkar, P.}, \bibinfo{author}{Irvin, J.},
  \bibinfo{author}{Bagul, A.}, \bibinfo{author}{Ding, D.},
  \bibinfo{author}{Duan, T.}, \bibinfo{author}{Mehta, H.},
  \bibinfo{author}{Yang, B.}, \bibinfo{author}{Zhu, K.},
  \bibinfo{author}{Laird, D.}, \bibinfo{author}{Ball, R.L.},
  \bibinfo{author}{Langlotz, C.}, \bibinfo{author}{Shpanskaya, K.},
  \bibinfo{author}{Lungren, M.P.}, \bibinfo{author}{Ng, A.Y.},
  \bibinfo{year}{2017}.
\newblock \bibinfo{title}{{MURA: Large Dataset for Abnormality Detection in
  Musculoskeletal Radiographs}} , \bibinfo{pages}{1--10}.
\bibitem[{Ranzato et~al.(2007)Ranzato, Huang, Boureau and LeCun}]{Ranzato2007}
\bibinfo{author}{Ranzato, M.}, \bibinfo{author}{Huang, F.J.},
  \bibinfo{author}{Boureau, Y.}, \bibinfo{author}{LeCun, Y.},
  \bibinfo{year}{2007}.
\newblock \bibinfo{title}{{Unsupervised Learning of Invariant Feature
  Hierarchies with Applications to Object Recognition}}, in:
  \bibinfo{booktitle}{2007 IEEE Conference on Computer Vision and Pattern
  Recognition}, pp. \bibinfo{pages}{1--8}.
\newblock \DOIprefix\doi{10.1109/CVPR.2007.383157}.
\bibitem[{Rawat and Wang(2017)}]{Rawat2017}
\bibinfo{author}{Rawat, W.}, \bibinfo{author}{Wang, Z.}, \bibinfo{year}{2017}.
\newblock \bibinfo{title}{{Deep Convolutional Neural Networks for Image
  Classification: A Comprehensive Review}}.
\newblock \bibinfo{journal}{Neural Computation} \bibinfo{volume}{29},
  \bibinfo{pages}{2352--2449}.
\newblock \DOIprefix\doi{10.1162/neco_a_00990}.
\bibitem[{Razakarivony and Jurie(2016)}]{Razakarivony2016}
\bibinfo{author}{Razakarivony, S.}, \bibinfo{author}{Jurie, F.},
  \bibinfo{year}{2016}.
\newblock \bibinfo{title}{{Vehicle detection in aerial imagery: A small target
  detection benchmark}}.
\newblock \bibinfo{journal}{Journal of Visual Communication and Image
  Representation} \bibinfo{volume}{34}, \bibinfo{pages}{187--203}.
\newblock \DOIprefix\doi{10.1016/j.jvcir.2015.11.002}.
\bibitem[{Real et~al.(2017)Real, Shlens, Mazzocchi, Pan and
  Vanhoucke}]{Real2017}
\bibinfo{author}{Real, E.}, \bibinfo{author}{Shlens, J.},
  \bibinfo{author}{Mazzocchi, S.}, \bibinfo{author}{Pan, X.},
  \bibinfo{author}{Vanhoucke, V.}, \bibinfo{year}{2017}.
\newblock \bibinfo{title}{{YouTube-BoundingBoxes: A large high-precision
  human-annotated data set for object detection in video}}, in:
  \bibinfo{booktitle}{Proceedings - 30th IEEE Conference on Computer Vision and
  Pattern Recognition, CVPR 2017}, pp. \bibinfo{pages}{7464--7473}.
\newblock \DOIprefix\doi{10.1109/CVPR.2017.789}.
\bibitem[{Redmon et~al.()Redmon, Divvala, Girshick and Farhadi}]{Redmon}
\bibinfo{author}{Redmon, J.}, \bibinfo{author}{Divvala, S.},
  \bibinfo{author}{Girshick, R.}, \bibinfo{author}{Farhadi, A.}, .
\newblock \bibinfo{title}{{You Only Look Once: Unified, Real-Time Object
  Detection}}.
\bibitem[{Redmon and Farhadi(2016)}]{Redmon2016}
\bibinfo{author}{Redmon, J.}, \bibinfo{author}{Farhadi, A.},
  \bibinfo{year}{2016}.
\newblock \bibinfo{title}{{YOLO9000: Better, Faster, Stronger}} .
\bibitem[{Redmon and Farhadi(2018)}]{Redmon2018}
\bibinfo{author}{Redmon, J.}, \bibinfo{author}{Farhadi, A.},
  \bibinfo{year}{2018}.
\newblock \bibinfo{title}{{YOLOv3: An Incremental Improvement}} .
\bibitem[{Reinertsen et~al.(2019)Reinertsen, Xiao, Rivaz and
  Chabanas}]{Reinertsen2019}
\bibinfo{author}{Reinertsen, I.}, \bibinfo{author}{Xiao, Y.},
  \bibinfo{author}{Rivaz, H.}, \bibinfo{author}{Chabanas, M.},
  \bibinfo{year}{2019}.
\newblock \bibinfo{title}{{CuRIOUS 2019}}.
\newblock \URLprefix \url{https://curious2019.grand-challenge.org/}.
\bibitem[{Ren et~al.(2015)Ren, He, Girshick and Sun}]{Ren2015}
\bibinfo{author}{Ren, S.}, \bibinfo{author}{He, K.}, \bibinfo{author}{Girshick,
  R.}, \bibinfo{author}{Sun, J.}, \bibinfo{year}{2015}.
\newblock \bibinfo{title}{{Faster r-cnn: Towards real-time object detection
  with region proposal networks}}, in: \bibinfo{booktitle}{Advances in neural
  information processing systems}, pp. \bibinfo{pages}{91--99}.
\bibitem[{Ronneberger et~al.(2015)Ronneberger, Fischer and
  Brox}]{Ronneberger2015}
\bibinfo{author}{Ronneberger, O.}, \bibinfo{author}{Fischer, P.},
  \bibinfo{author}{Brox, T.}, \bibinfo{year}{2015}.
\newblock \bibinfo{title}{{U-net: Convolutional networks for biomedical image
  segmentation}}, in: \bibinfo{booktitle}{International Conference on Medical
  image computing and computer-assisted intervention}, pp.
  \bibinfo{pages}{234--241}.
\bibitem[{Rothe et~al.()Rothe, Timofte and {Van Gool}}]{Rothe}
\bibinfo{author}{Rothe, R.}, \bibinfo{author}{Timofte, R.},
  \bibinfo{author}{{Van Gool}, L.}, .
\newblock \bibinfo{title}{{Deep expectation of real and apparent age from a
  single image without facial landmarks Real age 20 years DEX age predic3on}}.
\newblock \bibinfo{type}{Technical Report}.
\bibitem[{Rothe et~al.(2018)Rothe, Timofte and {Van Gool}}]{Rothe2018}
\bibinfo{author}{Rothe, R.}, \bibinfo{author}{Timofte, R.},
  \bibinfo{author}{{Van Gool}, L.}, \bibinfo{year}{2018}.
\newblock \bibinfo{title}{{Deep Expectation of Real and Apparent Age from a
  Single Image Without Facial Landmarks}}.
\newblock \bibinfo{journal}{International Journal of Computer Vision}
  \bibinfo{volume}{126}, \bibinfo{pages}{144--157}.
\newblock \DOIprefix\doi{10.1007/s11263-016-0940-3}.
\bibitem[{Russakovsky et~al.(2015)Russakovsky, Deng, Su, Krause, Satheesh, Ma,
  Huang, Karpathy, Khosla, Bernstein, Berg and Fei-Fei}]{Russakovsky2015}
\bibinfo{author}{Russakovsky, O.}, \bibinfo{author}{Deng, J.},
  \bibinfo{author}{Su, H.}, \bibinfo{author}{Krause, J.},
  \bibinfo{author}{Satheesh, S.}, \bibinfo{author}{Ma, S.},
  \bibinfo{author}{Huang, Z.}, \bibinfo{author}{Karpathy, A.},
  \bibinfo{author}{Khosla, A.}, \bibinfo{author}{Bernstein, M.},
  \bibinfo{author}{Berg, A.C.}, \bibinfo{author}{Fei-Fei, L.},
  \bibinfo{year}{2015}.
\newblock \bibinfo{title}{{ImageNet Large Scale Visual Recognition Challenge}}.
\newblock \bibinfo{journal}{International Journal of Computer Vision}
  \bibinfo{volume}{115}, \bibinfo{pages}{211--252}.
\newblock \DOIprefix\doi{10.1007/s11263-015-0816-y}.
\bibitem[{Russell et~al.(2008)Russell, Torralba, Murphy and
  Freeman}]{Russell2008}
\bibinfo{author}{Russell, B.C.}, \bibinfo{author}{Torralba, A.},
  \bibinfo{author}{Murphy, K.P.}, \bibinfo{author}{Freeman, W.T.},
  \bibinfo{year}{2008}.
\newblock \bibinfo{title}{{LabelMe: A database and web-based tool for image
  annotation}}.
\newblock \bibinfo{journal}{International Journal of Computer Vision}
  \bibinfo{volume}{77}, \bibinfo{pages}{157--173}.
\newblock \DOIprefix\doi{10.1007/s11263-007-0090-8}.
\bibitem[{Schroff and Philbin()}]{Schroff}
\bibinfo{author}{Schroff, F.}, \bibinfo{author}{Philbin, J.}, .
\newblock \bibinfo{title}{{FaceNet: A Unified Embedding for Face Recognition
  and Clustering}}.
\newblock \bibinfo{type}{Technical Report}.
\bibitem[{Sensing et~al.(2017)Sensing, Sciences, Hackel, Savinov, Ladicky,
  Wegner, Schindler and Pollefeys}]{Sensing2017}
\bibinfo{author}{Sensing, R.}, \bibinfo{author}{Sciences, S.I.},
  \bibinfo{author}{Hackel, T.}, \bibinfo{author}{Savinov, N.},
  \bibinfo{author}{Ladicky, L.}, \bibinfo{author}{Wegner, J.D.},
  \bibinfo{author}{Schindler, K.}, \bibinfo{author}{Pollefeys, M.},
  \bibinfo{year}{2017}.
\newblock \bibinfo{title}{{SEMANTIC3D . NET : A NEW LARGE-SCALE POINT CLOUD
  CLASSIFICATION}} \bibinfo{volume}{IV}, \bibinfo{pages}{6--9}.
\newblock \DOIprefix\doi{10.5194/isprs-annals-IV-1-W1-91-2017}.
\bibitem[{Shafiee et~al.(2017)Shafiee, Chywl, Li and Wong}]{Shafiee2017}
\bibinfo{author}{Shafiee, M.J.}, \bibinfo{author}{Chywl, B.},
  \bibinfo{author}{Li, F.}, \bibinfo{author}{Wong, A.}, \bibinfo{year}{2017}.
\newblock \bibinfo{title}{{Fast YOLO: A Fast You Only Look Once System for
  Real-time Embedded Object Detection in Video}} .
\bibitem[{Shao et~al.(2019)Shao, Li, Zhang, Peng, Yu, Zhang, Li, Sun and
  Technology}]{Shao2019}
\bibinfo{author}{Shao, S.}, \bibinfo{author}{Li, Z.}, \bibinfo{author}{Zhang,
  T.}, \bibinfo{author}{Peng, C.}, \bibinfo{author}{Yu, G.},
  \bibinfo{author}{Zhang, X.}, \bibinfo{author}{Li, J.}, \bibinfo{author}{Sun,
  J.}, \bibinfo{author}{Technology, M.}, \bibinfo{year}{2019}.
\newblock \bibinfo{title}{{Objects365: A Large-scale, High-quality Dataset for
  Object Detection}}.
\newblock \bibinfo{journal}{Proc. IEEE International Conference on Computer
  Vision (ICCV)} , \bibinfo{pages}{8430--8439}.
\bibitem[{Shotton et~al.(2006)Shotton, Winn, Rother and
  Criminisi}]{10.1007/11744023_1}
\bibinfo{author}{Shotton, J.}, \bibinfo{author}{Winn, J.},
  \bibinfo{author}{Rother, C.}, \bibinfo{author}{Criminisi, A.},
  \bibinfo{year}{2006}.
\newblock \bibinfo{title}{{TextonBoost: Joint Appearance, Shape and Context
  Modeling for Multi-class Object Recognition and Segmentation}}, in:
  \bibinfo{editor}{Leonardis, A.}, \bibinfo{editor}{Bischof, H.},
  \bibinfo{editor}{Pinz, A.} (Eds.), \bibinfo{booktitle}{Computer Vision --
  ECCV 2006}, \bibinfo{publisher}{Springer Berlin Heidelberg},
  \bibinfo{address}{Berlin, Heidelberg}. pp. \bibinfo{pages}{1--15}.
\bibitem[{Silberman et~al.(2012)Silberman, Hoiem, Kohli and
  Fergus}]{Silberman2012}
\bibinfo{author}{Silberman, N.}, \bibinfo{author}{Hoiem, D.},
  \bibinfo{author}{Kohli, P.}, \bibinfo{author}{Fergus, R.},
  \bibinfo{year}{2012}.
\newblock \bibinfo{title}{{Indoor segmentation and support inference from RGBD
  images}}.
\newblock \bibinfo{journal}{Lecture Notes in Computer Science (including
  subseries Lecture Notes in Artificial Intelligence and Lecture Notes in
  Bioinformatics)} \bibinfo{volume}{7576 LNCS}, \bibinfo{pages}{746--760}.
\newblock \DOIprefix\doi{10.1007/978-3-642-33715-4_54}.
\bibitem[{Sim et~al.(2002)Sim, Baker and Bsat}]{Sim2002}
\bibinfo{author}{Sim, T.}, \bibinfo{author}{Baker, S.}, \bibinfo{author}{Bsat,
  M.}, \bibinfo{year}{2002}.
\newblock \bibinfo{title}{{The CMU Pose, Illumination, and Expression (PIE)
  database}}.
\newblock \bibinfo{journal}{Proceedings - 5th IEEE International Conference on
  Automatic Face Gesture Recognition, FGR 2002} ,
  \bibinfo{pages}{53--58}\DOIprefix\doi{10.1109/AFGR.2002.1004130}.
\bibitem[{Sirinukunwattana et~al.(2016)Sirinukunwattana, Raza, Tsang, Snead,
  Cree and Rajpoot}]{Sirinukunwattana2016}
\bibinfo{author}{Sirinukunwattana, K.}, \bibinfo{author}{Raza, s.E.A.},
  \bibinfo{author}{Tsang, Y.}, \bibinfo{author}{Snead, D.R.},
  \bibinfo{author}{Cree, I.A.}, \bibinfo{author}{Rajpoot, N.M.},
  \bibinfo{year}{2016}.
\newblock \bibinfo{title}{{Locality Sensitive Deep Learning for Detection and
  Classification of Nuclei in Routine Colon Cancer Histology Images}}.
\newblock \bibinfo{journal}{IEEE Transactions on Medical Imaging}
  \bibinfo{volume}{35}, \bibinfo{pages}{1196--1206}.
\newblock \DOIprefix\doi{10.1109/TMI.2016.2525803}.
\bibitem[{Song et~al.(2015)Song, Lichtenberg and Xiao}]{Song2015}
\bibinfo{author}{Song, S.}, \bibinfo{author}{Lichtenberg, S.P.},
  \bibinfo{author}{Xiao, J.}, \bibinfo{year}{2015}.
\newblock \bibinfo{title}{{SUN RGB-D: A RGB-D scene understanding benchmark
  suite}}.
\newblock \bibinfo{journal}{Proceedings of the IEEE Computer Society Conference
  on Computer Vision and Pattern Recognition} \bibinfo{volume}{07-12-June},
  \bibinfo{pages}{567--576}.
\newblock \DOIprefix\doi{10.1109/CVPR.2015.7298655}.
\bibitem[{S{\o}rensen et~al.(2010)S{\o}rensen, Shaker and {De
  Bruijne}}]{Sorensen2010}
\bibinfo{author}{S{\o}rensen, L.}, \bibinfo{author}{Shaker, S.B.},
  \bibinfo{author}{{De Bruijne}, M.}, \bibinfo{year}{2010}.
\newblock \bibinfo{title}{{Quantitative analysis of pulmonary emphysema using
  local binary patterns}}.
\newblock \bibinfo{journal}{IEEE Transactions on Medical Imaging}
  \bibinfo{volume}{29}, \bibinfo{pages}{559--569}.
\newblock \DOIprefix\doi{10.1109/TMI.2009.2038575}.
\bibitem[{Souza et~al.(2018)Souza, Lucena, Garrafa, Gobbi, Saluzzi,
  Appenzeller, Rittner, Frayne and Lotofo}]{Souza2018}
\bibinfo{author}{Souza, R.}, \bibinfo{author}{Lucena, O.},
  \bibinfo{author}{Garrafa, J.}, \bibinfo{author}{Gobbi, D.},
  \bibinfo{author}{Saluzzi, M.}, \bibinfo{author}{Appenzeller, S.},
  \bibinfo{author}{Rittner, L.}, \bibinfo{author}{Frayne, R.},
  \bibinfo{author}{Lotofo, R.}, \bibinfo{year}{2018}.
\newblock \bibinfo{title}{{An open, multi-vendor, multi-field-strength brain MR
  dataset and analysis of publicly available skull stripping methods
  agreement.}}
\newblock \bibinfo{journal}{NeuroImage} \bibinfo{volume}{170},
  \bibinfo{pages}{482--494}.
\newblock \DOIprefix\doi{https://doi.org/10.1016/j.neuroimage.2017.08.021}.
\bibitem[{Staal et~al.(2013)Staal, Abr{\`{a}}moff, Niemeijer, Viergever and
  Ginneken}]{Staal2013}
\bibinfo{author}{Staal, J.}, \bibinfo{author}{Abr{\`{a}}moff, M.},
  \bibinfo{author}{Niemeijer, M.}, \bibinfo{author}{Viergever, M.},
  \bibinfo{author}{Ginneken, B.}, \bibinfo{year}{2013}.
\newblock \bibinfo{title}{{Digital Retinal Image for Vessel Extraction (DRIVE)
  Database}}.
\bibitem[{Sun et~al.(2018)Sun, Yuan, Zhou and Ding}]{Sun2018a}
\bibinfo{author}{Sun, M.}, \bibinfo{author}{Yuan, Y.}, \bibinfo{author}{Zhou,
  F.}, \bibinfo{author}{Ding, E.}, \bibinfo{year}{2018}.
\newblock \bibinfo{title}{{Multi-Attention Multi-Class Constraint for
  Fine-grained Image Recognition}}, in: \bibinfo{booktitle}{Lecture Notes in
  Computer Science (including subseries Lecture Notes in Artificial
  Intelligence and Lecture Notes in Bioinformatics)}, pp.
  \bibinfo{pages}{834--850}.
\newblock \DOIprefix\doi{10.1007/978-3-030-01270-0_49}.
\bibitem[{Sun et~al.(2019)Sun, Kretzschmar, Dotiwalla, Chouard, Patnaik, Tsui,
  Guo, Zhou, Chai, Caine, Vasudevan, Han, Ngiam, Zhao, Timofeev, Ettinger,
  Krivokon, Gao, Joshi, Zhang, Shlens, Chen and Anguelov}]{Sun2019}
\bibinfo{author}{Sun, P.}, \bibinfo{author}{Kretzschmar, H.},
  \bibinfo{author}{Dotiwalla, X.}, \bibinfo{author}{Chouard, A.},
  \bibinfo{author}{Patnaik, V.}, \bibinfo{author}{Tsui, P.},
  \bibinfo{author}{Guo, J.}, \bibinfo{author}{Zhou, Y.}, \bibinfo{author}{Chai,
  Y.}, \bibinfo{author}{Caine, B.}, \bibinfo{author}{Vasudevan, V.},
  \bibinfo{author}{Han, W.}, \bibinfo{author}{Ngiam, J.},
  \bibinfo{author}{Zhao, H.}, \bibinfo{author}{Timofeev, A.},
  \bibinfo{author}{Ettinger, S.}, \bibinfo{author}{Krivokon, M.},
  \bibinfo{author}{Gao, A.}, \bibinfo{author}{Joshi, A.},
  \bibinfo{author}{Zhang, Y.}, \bibinfo{author}{Shlens, J.},
  \bibinfo{author}{Chen, Z.}, \bibinfo{author}{Anguelov, D.},
  \bibinfo{year}{2019}.
\newblock \bibinfo{title}{{Scalability in Perception for Autonomous Driving:
  Waymo Open Dataset}} .
\bibitem[{Sun et~al.(2015)Sun, Liang, Wang and Tang}]{Sun2015}
\bibinfo{author}{Sun, Y.}, \bibinfo{author}{Liang, D.}, \bibinfo{author}{Wang,
  X.}, \bibinfo{author}{Tang, X.}, \bibinfo{year}{2015}.
\newblock \bibinfo{title}{{DeepID3: Face Recognition with Very Deep Neural
  Networks}} \URLprefix \url{http://arxiv.org/abs/1502.00873}.
\bibitem[{Sun et~al.()Sun, Wang and Tang}]{Sun}
\bibinfo{author}{Sun, Y.}, \bibinfo{author}{Wang, X.}, \bibinfo{author}{Tang,
  X.}, .
\newblock \bibinfo{title}{{Deep Learning Face Representation by Joint
  Identification-Verification}}.
\newblock \bibinfo{type}{Technical Report}.
\bibitem[{Sun et~al.(2014)Sun, Wang and Tang}]{Sun2014}
\bibinfo{author}{Sun, Y.}, \bibinfo{author}{Wang, X.}, \bibinfo{author}{Tang,
  X.}, \bibinfo{year}{2014}.
\newblock \bibinfo{title}{{Deep learning face representation from predicting
  10,000 classes}}, in: \bibinfo{booktitle}{Proceedings of the IEEE Computer
  Society Conference on Computer Vision and Pattern Recognition},
  \bibinfo{publisher}{IEEE Computer Society}. pp. \bibinfo{pages}{1891--1898}.
\newblock \DOIprefix\doi{10.1109/CVPR.2014.244}.
\bibitem[{Sung(1996)}]{Sung1996}
\bibinfo{author}{Sung, K.k.}, \bibinfo{year}{1996}.
\newblock \bibinfo{title}{{Learning and Example Selection for Object and
  Pattern Detection}}.
\newblock \bibinfo{journal}{PhD thesis} ,
  \bibinfo{pages}{195}\DOIprefix\doi{https://doi.org/10.1016/j.comnet.2014.12.002}.
\bibitem[{Swanson et~al.(2015)Swanson, Kosmala, Lintott, Simpson, Smith and
  Packer}]{Swanson2015}
\bibinfo{author}{Swanson, A.}, \bibinfo{author}{Kosmala, M.},
  \bibinfo{author}{Lintott, C.}, \bibinfo{author}{Simpson, R.},
  \bibinfo{author}{Smith, A.}, \bibinfo{author}{Packer, C.},
  \bibinfo{year}{2015}.
\newblock \bibinfo{title}{{Snapshot Serengeti, high-frequency annotated camera
  trap images of 40 mammalian species in an African savanna}}.
\newblock \bibinfo{journal}{Scientific Data} \bibinfo{volume}{2},
  \bibinfo{pages}{1--14}.
\newblock \DOIprefix\doi{10.1038/sdata.2015.26}.
\bibitem[{Taghanaki et~al.(2020)Taghanaki, Abhishek, Cohen, Cohen-Adad and
  Hamarneh}]{taghanaki2020deep}
\bibinfo{author}{Taghanaki, S.A.}, \bibinfo{author}{Abhishek, K.},
  \bibinfo{author}{Cohen, J.P.}, \bibinfo{author}{Cohen-Adad, J.},
  \bibinfo{author}{Hamarneh, G.}, \bibinfo{year}{2020}.
\newblock \bibinfo{title}{Deep semantic segmentation of natural and medical
  images: A review} .
\bibitem[{Taigman et~al.()Taigman, Marc', Ranzato and Wolf}]{Taigman}
\bibinfo{author}{Taigman, Y.}, \bibinfo{author}{Marc', M.Y.},
  \bibinfo{author}{Ranzato, A.}, \bibinfo{author}{Wolf, L.}, .
\newblock \bibinfo{title}{{DeepFace: Closing the Gap to Human-Level Performance
  in Face Verification}}.
\newblock \bibinfo{type}{Technical Report}.
\bibitem[{Taskiran et~al.(2020)Taskiran, Kahraman and
  Erdem}]{TASKIRAN2020102809}
\bibinfo{author}{Taskiran, M.}, \bibinfo{author}{Kahraman, N.},
  \bibinfo{author}{Erdem, C.E.}, \bibinfo{year}{2020}.
\newblock \bibinfo{title}{Face recognition: Past, present and future (a
  review)}.
\newblock \bibinfo{journal}{Digital Signal Processing} \bibinfo{volume}{106},
  \bibinfo{pages}{102809}.
\newblock \URLprefix
  \url{https://www.sciencedirect.com/science/article/pii/S1051200420301548},
  \DOIprefix\doi{https://doi.org/10.1016/j.dsp.2020.102809}.
\bibitem[{Thomee et~al.(2016)Thomee, Elizalde, Shamma, Ni, Friedland, Poland,
  Borth and Li}]{Thomee2016}
\bibinfo{author}{Thomee, B.}, \bibinfo{author}{Elizalde, B.},
  \bibinfo{author}{Shamma, D.A.}, \bibinfo{author}{Ni, K.},
  \bibinfo{author}{Friedland, G.}, \bibinfo{author}{Poland, D.},
  \bibinfo{author}{Borth, D.}, \bibinfo{author}{Li, L.J.},
  \bibinfo{year}{2016}.
\newblock \bibinfo{title}{{YFCC100M: The new data in multimedia research}}.
\newblock \bibinfo{journal}{Communications of the ACM} \bibinfo{volume}{59},
  \bibinfo{pages}{64--73}.
\newblock \DOIprefix\doi{10.1145/2812802}.
\bibitem[{Tighe and Lazebnik(2010)}]{10.1007/978-3-642-15555-0_26}
\bibinfo{author}{Tighe, J.}, \bibinfo{author}{Lazebnik, S.},
  \bibinfo{year}{2010}.
\newblock \bibinfo{title}{{SuperParsing: Scalable Nonparametric Image Parsing
  with Superpixels}}, in: \bibinfo{editor}{Daniilidis, K.},
  \bibinfo{editor}{Maragos, P.}, \bibinfo{editor}{Paragios, N.} (Eds.),
  \bibinfo{booktitle}{Computer Vision -- ECCV 2010},
  \bibinfo{publisher}{Springer Berlin Heidelberg}, \bibinfo{address}{Berlin,
  Heidelberg}. pp. \bibinfo{pages}{352--365}.
\bibitem[{Tighe and Lazebnik(2013)}]{Tighe2013}
\bibinfo{author}{Tighe, J.}, \bibinfo{author}{Lazebnik, S.},
  \bibinfo{year}{2013}.
\newblock \bibinfo{title}{{Superparsing: Scalable nonparametric image parsing
  with superpixels}}.
\newblock \bibinfo{journal}{International Journal of Computer Vision}
  \bibinfo{volume}{101}, \bibinfo{pages}{329--349}.
\newblock \DOIprefix\doi{10.1007/s11263-012-0574-z}.
\bibitem[{Torralba et~al.(2008)Torralba, Fergus and Freeman}]{Torralba2008}
\bibinfo{author}{Torralba, A.}, \bibinfo{author}{Fergus, R.},
  \bibinfo{author}{Freeman, W.T.}, \bibinfo{year}{2008}.
\newblock \bibinfo{title}{{80 million tiny images: A large data set for
  nonparametric object and scene recognition}}.
\newblock \bibinfo{journal}{IEEE Transactions on Pattern Analysis and Machine
  Intelligence} \bibinfo{volume}{30}, \bibinfo{pages}{1958--1970}.
\newblock \DOIprefix\doi{10.1109/TPAMI.2008.128}.
\bibitem[{Torralba et~al.(2004)Torralba, Murphy and Freeman}]{Torralba2004}
\bibinfo{author}{Torralba, A.}, \bibinfo{author}{Murphy, K.P.},
  \bibinfo{author}{Freeman, W.T.}, \bibinfo{year}{2004}.
\newblock \bibinfo{title}{{Sharing features: Efficient boosting procedures for
  multiclass object detection}}.
\newblock \bibinfo{journal}{Proceedings of the IEEE Computer Society Conference
  on Computer Vision and Pattern Recognition} \bibinfo{volume}{2}.
\newblock \DOIprefix\doi{10.1109/cvpr.2004.1315241}.
\bibitem[{Tschandl et~al.(2018)Tschandl, Rosendahl and Kittler}]{Tschandl2018}
\bibinfo{author}{Tschandl, P.}, \bibinfo{author}{Rosendahl, C.},
  \bibinfo{author}{Kittler, H.}, \bibinfo{year}{2018}.
\newblock \bibinfo{title}{{Data descriptor: The HAM10000 dataset, a large
  collection of multi-source dermatoscopic images of common pigmented skin
  lesions}}.
\newblock \bibinfo{journal}{Scientific Data} \bibinfo{volume}{5},
  \bibinfo{pages}{1--9}.
\newblock \DOIprefix\doi{10.1038/sdata.2018.161}.
\bibitem[{Twinanda et~al.(2017)Twinanda, Shehata, Mutter, Marescaux, {De
  Mathelin} and Padoy}]{Twinanda2017}
\bibinfo{author}{Twinanda, A.P.}, \bibinfo{author}{Shehata, S.},
  \bibinfo{author}{Mutter, D.}, \bibinfo{author}{Marescaux, J.},
  \bibinfo{author}{{De Mathelin}, M.}, \bibinfo{author}{Padoy, N.},
  \bibinfo{year}{2017}.
\newblock \bibinfo{title}{{EndoNet: A Deep Architecture for Recognition Tasks
  on Laparoscopic Videos}}.
\newblock \bibinfo{journal}{IEEE Transactions on Medical Imaging}
  \bibinfo{volume}{36}, \bibinfo{pages}{86--97}.
\newblock \DOIprefix\doi{10.1109/TMI.2016.2593957}.
\bibitem[{{University of Minnesota} and {University of
  Melbourne}(2019)}]{UniversityofMinnesota2019}
\bibinfo{author}{{University of Minnesota}}, \bibinfo{author}{{University of
  Melbourne}}, \bibinfo{year}{2019}.
\newblock \bibinfo{title}{{KiTS19 Challenge}}.
\newblock \URLprefix \url{https://kits19.grand-challenge.org/}.
\bibitem[{{Van Brummelen} et~al.(2018){Van Brummelen}, O'Brien, Gruyer and
  Najjaran}]{VanBrummelen2018}
\bibinfo{author}{{Van Brummelen}, J.}, \bibinfo{author}{O'Brien, M.},
  \bibinfo{author}{Gruyer, D.}, \bibinfo{author}{Najjaran, H.},
  \bibinfo{year}{2018}.
\newblock \bibinfo{title}{{Autonomous vehicle perception: The technology of
  today and tomorrow}}.
\newblock \bibinfo{journal}{Transportation Research Part C: Emerging
  Technologies} \bibinfo{volume}{89}, \bibinfo{pages}{384--406}.
\newblock \URLprefix \url{https://doi.org/10.1016/j.trc.2018.02.012},
  \DOIprefix\doi{10.1016/j.trc.2018.02.012}.
\bibitem[{Viola et~al.(2001a)Viola, Viola and Jones}]{Viola2001a}
\bibinfo{author}{Viola, P.}, \bibinfo{author}{Viola, P.},
  \bibinfo{author}{Jones, M.}, \bibinfo{year}{2001}a.
\newblock \bibinfo{title}{{Rapid object detection using a boosted cascade of
  simple features}}.
\newblock \bibinfo{journal}{ACCEPTED CONFERENCE ON COMPUTER VISION AND PATTERN
  RECOGNITION 2001} .
\bibitem[{Viola et~al.(2001b)Viola, Viola and Jones}]{Viola2001}
\bibinfo{author}{Viola, P.}, \bibinfo{author}{Viola, P.},
  \bibinfo{author}{Jones, M.}, \bibinfo{year}{2001}b.
\newblock \bibinfo{title}{{Robust Real-time Object Detection}}.
\newblock \bibinfo{journal}{INTERNATIONAL JOURNAL OF COMPUTER VISION} .
\bibitem[{Wah et~al.(2011)Wah, Branson, Welinder, Perona and
  Belongie}]{wah2011caltech}
\bibinfo{author}{Wah, C.}, \bibinfo{author}{Branson, S.},
  \bibinfo{author}{Welinder, P.}, \bibinfo{author}{Perona, P.},
  \bibinfo{author}{Belongie, S.}, \bibinfo{year}{2011}.
\newblock \bibinfo{title}{{The Caltech-ucsd Birds-200-2011 Dataset}} .
\bibitem[{Wang et~al.(2017a)Wang, Lu, Wang, Feng, Wang, Yin and Ruan}]{DUTS}
\bibinfo{author}{Wang, L.}, \bibinfo{author}{Lu, H.}, \bibinfo{author}{Wang,
  Y.}, \bibinfo{author}{Feng, M.}, \bibinfo{author}{Wang, D.},
  \bibinfo{author}{Yin, B.}, \bibinfo{author}{Ruan, X.}, \bibinfo{year}{2017}a.
\newblock \bibinfo{title}{Learning to detect salient objects with image-level
  supervision} ,
  \bibinfo{pages}{3796--3805}\DOIprefix\doi{10.1109/CVPR.2017.404}.
\bibitem[{Wang et~al.(2019)Wang, Huang, Cheng, Zhou, Geng and Yang}]{Wang2019}
\bibinfo{author}{Wang, P.}, \bibinfo{author}{Huang, X.},
  \bibinfo{author}{Cheng, X.}, \bibinfo{author}{Zhou, D.},
  \bibinfo{author}{Geng, Q.}, \bibinfo{author}{Yang, R.}, \bibinfo{year}{2019}.
\newblock \bibinfo{title}{{The ApolloScape Open Dataset for Autonomous Driving
  and its Application}}.
\newblock \bibinfo{journal}{IEEE Transactions on Pattern Analysis and Machine
  Intelligence} ,
  \bibinfo{pages}{1--1}\DOIprefix\doi{10.1109/tpami.2019.2926463}.
\bibitem[{Wang et~al.()Wang, Bai, Mattyus, Chu, Luo, Yang, Liang, Cheverie,
  Fidler and Urtasun}]{Wang}
\bibinfo{author}{Wang, S.}, \bibinfo{author}{Bai, M.},
  \bibinfo{author}{Mattyus, G.}, \bibinfo{author}{Chu, H.},
  \bibinfo{author}{Luo, W.}, \bibinfo{author}{Yang, B.},
  \bibinfo{author}{Liang, J.}, \bibinfo{author}{Cheverie, J.},
  \bibinfo{author}{Fidler, S.}, \bibinfo{author}{Urtasun, R.}, .
\newblock \bibinfo{title}{{TorontoCity: Seeing the World with a Million Eyes}}.
\newblock \bibinfo{type}{Technical Report}.
\bibitem[{Wang et~al.(2016)Wang, Bai, Mattyus, Chu, Luo, Yang, Liang, Cheverie,
  Fidler and Urtasun}]{Wang2016}
\bibinfo{author}{Wang, S.}, \bibinfo{author}{Bai, M.},
  \bibinfo{author}{Mattyus, G.}, \bibinfo{author}{Chu, H.},
  \bibinfo{author}{Luo, W.}, \bibinfo{author}{Yang, B.},
  \bibinfo{author}{Liang, J.}, \bibinfo{author}{Cheverie, J.},
  \bibinfo{author}{Fidler, S.}, \bibinfo{author}{Urtasun, R.},
  \bibinfo{year}{2016}.
\newblock \bibinfo{title}{{TorontoCity : Seeing the World with a Million Eyes}}
  .
\bibitem[{Wang et~al.(2017b)Wang, Peng, Lu, Lu, Bagheri and
  Summers}]{Wang2017a}
\bibinfo{author}{Wang, X.}, \bibinfo{author}{Peng, Y.}, \bibinfo{author}{Lu,
  L.}, \bibinfo{author}{Lu, Z.}, \bibinfo{author}{Bagheri, M.},
  \bibinfo{author}{Summers, R.M.}, \bibinfo{year}{2017}b.
\newblock \bibinfo{title}{{ChestX-ray8: Hospital-scale chest X-ray database and
  benchmarks on weakly-supervised classification and localization of common
  thorax diseases}}.
\newblock \bibinfo{journal}{Proceedings - 30th IEEE Conference on Computer
  Vision and Pattern Recognition, CVPR 2017} \bibinfo{volume}{2017-Janua},
  \bibinfo{pages}{3462--3471}.
\newblock \DOIprefix\doi{10.1109/CVPR.2017.369}.
\bibitem[{Whitelam et~al.(2017)Whitelam, Taborsky, Blanton, Maze, Adams,
  Miller, Kalka, Jain, Duncan, Allen, Cheney and Grother}]{Whitelam2017}
\bibinfo{author}{Whitelam, C.}, \bibinfo{author}{Taborsky, E.},
  \bibinfo{author}{Blanton, A.}, \bibinfo{author}{Maze, B.},
  \bibinfo{author}{Adams, J.}, \bibinfo{author}{Miller, T.},
  \bibinfo{author}{Kalka, N.}, \bibinfo{author}{Jain, A.K.},
  \bibinfo{author}{Duncan, J.A.}, \bibinfo{author}{Allen, K.},
  \bibinfo{author}{Cheney, J.}, \bibinfo{author}{Grother, P.},
  \bibinfo{year}{2017}.
\newblock \bibinfo{title}{{IARPA Janus Benchmark-B Face Dataset}}.
\newblock \bibinfo{journal}{IEEE Computer Society Conference on Computer Vision
  and Pattern Recognition Workshops} \bibinfo{volume}{2017-July},
  \bibinfo{pages}{592--600}.
\newblock \DOIprefix\doi{10.1109/CVPRW.2017.87}.
\bibitem[{{Winship Cancer Institute}()}]{WinshipCancerInstitute}
\bibinfo{author}{{Winship Cancer Institute}}, .
\newblock \bibinfo{title}{{Cancer Digital Slide Archive}}.
\newblock \URLprefix \url{https://cancer.digitalslidearchive.org/}.
\bibitem[{Wolf et~al.(2011)Wolf, Hassner and Maoz}]{Wolf2011}
\bibinfo{author}{Wolf, L.}, \bibinfo{author}{Hassner, T.},
  \bibinfo{author}{Maoz, I.}, \bibinfo{year}{2011}.
\newblock \bibinfo{title}{{Face recognition in unconstrained videos with
  matched background similarity}}.
\newblock \bibinfo{journal}{Proceedings of the IEEE Computer Society Conference
  on Computer Vision and Pattern Recognition} ,
  \bibinfo{pages}{529--534}\DOIprefix\doi{10.1109/CVPR.2011.5995566}.
\bibitem[{Wrenninge and Unger(2018)}]{Wrenninge2018}
\bibinfo{author}{Wrenninge, M.}, \bibinfo{author}{Unger, J.},
  \bibinfo{year}{2018}.
\newblock \bibinfo{title}{{Synscapes: A Photorealistic Synthetic Dataset for
  Street Scene Parsing}} \URLprefix \url{http://arxiv.org/abs/1810.08705}.
\bibitem[{Wu et~al.(2017)Wu, Bailey, Rasoulinejad and Li}]{Wu2017}
\bibinfo{author}{Wu, H.}, \bibinfo{author}{Bailey, C.},
  \bibinfo{author}{Rasoulinejad, P.}, \bibinfo{author}{Li, S.},
  \bibinfo{year}{2017}.
\newblock \bibinfo{title}{{Automatic Landmark Estimation for Adolescent
  Idiopathic Scoliosis Assessment Using BoostNet}}, in:
  \bibinfo{booktitle}{Medical Image Computing and Computer Assisted
  Intervention - MICCAI}, pp. \bibinfo{pages}{127--135}.
\bibitem[{Wu et~al.(2019)Wu, Zhan, Lai, Cheng and Yang}]{Wu2019}
\bibinfo{author}{Wu, X.}, \bibinfo{author}{Zhan, C.}, \bibinfo{author}{Lai,
  Y.K.}, \bibinfo{author}{Cheng, M.M.}, \bibinfo{author}{Yang, J.},
  \bibinfo{year}{2019}.
\newblock \bibinfo{title}{{IP102: A large-scale benchmark dataset for insect
  pest recognition}}, in: \bibinfo{booktitle}{Proceedings of the IEEE Computer
  Society Conference on Computer Vision and Pattern Recognition}, pp.
  \bibinfo{pages}{8779--8788}.
\newblock \DOIprefix\doi{10.1109/CVPR.2019.00899}.
\bibitem[{Xia et~al.(2017a)Xia, Li, Chen, Zheng and Zhang}]{XPIE}
\bibinfo{author}{Xia, C.}, \bibinfo{author}{Li, J.}, \bibinfo{author}{Chen,
  X.}, \bibinfo{author}{Zheng, A.}, \bibinfo{author}{Zhang, Y.},
  \bibinfo{year}{2017}a.
\newblock \bibinfo{title}{What is and what is not a salient object? learning
  salient object detector by ensembling linear exemplar regressors} ,
  \bibinfo{pages}{4399--4407}\DOIprefix\doi{10.1109/CVPR.2017.468}.
\bibitem[{Xia et~al.(2017b)Xia, Bai, Ding, Zhu, Belongie, Luo, Datcu, Pelillo
  and Zhang}]{Xia2017}
\bibinfo{author}{Xia, G.S.}, \bibinfo{author}{Bai, X.}, \bibinfo{author}{Ding,
  J.}, \bibinfo{author}{Zhu, Z.}, \bibinfo{author}{Belongie, S.},
  \bibinfo{author}{Luo, J.}, \bibinfo{author}{Datcu, M.},
  \bibinfo{author}{Pelillo, M.}, \bibinfo{author}{Zhang, L.},
  \bibinfo{year}{2017}b.
\newblock \bibinfo{title}{{DOTA: A Large-scale Dataset for Object Detection in
  Aerial Images}}.
\newblock \bibinfo{journal}{Proceedings of the IEEE Computer Society Conference
  on Computer Vision and Pattern Recognition} ,
  \bibinfo{pages}{3974--3983}\URLprefix \url{http://arxiv.org/abs/1711.10398}.
\bibitem[{Xia et~al.(2016)Xia, Hu, Hu, Shi, Bai, Zhong and Zhang}]{Xia2016}
\bibinfo{author}{Xia, G.S.}, \bibinfo{author}{Hu, J.}, \bibinfo{author}{Hu,
  F.}, \bibinfo{author}{Shi, B.}, \bibinfo{author}{Bai, X.},
  \bibinfo{author}{Zhong, Y.}, \bibinfo{author}{Zhang, L.},
  \bibinfo{year}{2016}.
\newblock \bibinfo{title}{{AID: A Benchmark Dataset for Performance Evaluation
  of Aerial Scene Classification}}.
\newblock \bibinfo{journal}{IEEE Transactions on Geoscience and Remote Sensing}
  \bibinfo{volume}{55}, \bibinfo{pages}{3965--3981}.
\newblock \URLprefix \url{http://arxiv.org/abs/1608.05167
  http://dx.doi.org/10.1109/TGRS.2017.2685945},
  \DOIprefix\doi{10.1109/TGRS.2017.2685945}.
\bibitem[{Xiao et~al.(2010)Xiao, Hays, Ehinger, Oliva and Torralba}]{Xiao2010}
\bibinfo{author}{Xiao, J.}, \bibinfo{author}{Hays, J.},
  \bibinfo{author}{Ehinger, K.A.}, \bibinfo{author}{Oliva, A.},
  \bibinfo{author}{Torralba, A.}, \bibinfo{year}{2010}.
\newblock \bibinfo{title}{{SUN database: Large-scale scene recognition from
  abbey to zoo}}.
\newblock \bibinfo{journal}{Proceedings of the IEEE Computer Society Conference
  on Computer Vision and Pattern Recognition} ,
  \bibinfo{pages}{3485--3492}\DOIprefix\doi{10.1109/CVPR.2010.5539970}.
\bibitem[{Xiao et~al.(2017)Xiao, Fortin, Unsg{\"{a}}rd, Rivaz and
  Reinertsen}]{Xiao2017}
\bibinfo{author}{Xiao, Y.}, \bibinfo{author}{Fortin, M.},
  \bibinfo{author}{Unsg{\"{a}}rd, G.}, \bibinfo{author}{Rivaz, H.},
  \bibinfo{author}{Reinertsen, I.}, \bibinfo{year}{2017}.
\newblock \bibinfo{title}{{REtroSpective Evaluation of Cerebral Tumors
  (RESECT): A clinical database of pre-operative MRI and intra-operative
  ultrasound in low-grade glioma surgeries: A}}.
\newblock \bibinfo{journal}{Medical Physics} \bibinfo{volume}{44},
  \bibinfo{pages}{3875--3882}.
\newblock \DOIprefix\doi{10.1002/mp.12268}.
\bibitem[{Xu et~al.(2018a)Xu, Anguelov and Jain}]{Xu2018}
\bibinfo{author}{Xu, D.}, \bibinfo{author}{Anguelov, D.},
  \bibinfo{author}{Jain, A.}, \bibinfo{year}{2018}a.
\newblock \bibinfo{title}{{PointFusion: Deep Sensor Fusion for 3D Bounding Box
  Estimation}}.
\newblock \bibinfo{type}{Technical Report}.
\bibitem[{Xu et~al.(2019)Xu, Song, Sun, Ku, Yang, Liu, Wang, Ma and
  Xu}]{Xu2019}
\bibinfo{author}{Xu, G.}, \bibinfo{author}{Song, Z.}, \bibinfo{author}{Sun,
  Z.}, \bibinfo{author}{Ku, C.}, \bibinfo{author}{Yang, Z.},
  \bibinfo{author}{Liu, C.}, \bibinfo{author}{Wang, S.}, \bibinfo{author}{Ma,
  J.}, \bibinfo{author}{Xu, W.}, \bibinfo{year}{2019}.
\newblock \bibinfo{title}{{CAMEL: A weakly supervised learning framework for
  histopathology image segmentation}}.
\newblock \bibinfo{journal}{Proceedings of the IEEE International Conference on
  Computer Vision} \bibinfo{volume}{2019-Octob}, \bibinfo{pages}{10681--10690}.
\newblock \DOIprefix\doi{10.1109/ICCV.2019.01078}.
\bibitem[{Xu et~al.(2018b)Xu, Yang, Fan, Yue, Liang, Yang and Huang}]{Xu2018a}
\bibinfo{author}{Xu, N.}, \bibinfo{author}{Yang, L.}, \bibinfo{author}{Fan,
  Y.}, \bibinfo{author}{Yue, D.}, \bibinfo{author}{Liang, Y.},
  \bibinfo{author}{Yang, J.}, \bibinfo{author}{Huang, T.},
  \bibinfo{year}{2018}b.
\newblock \bibinfo{title}{{YouTube-VOS: A Large-Scale Video Object Segmentation
  Benchmark}} , \bibinfo{pages}{1--10}\URLprefix
  \url{http://arxiv.org/abs/1809.03327}.
\bibitem[{Yan et~al.(2014)Yan, Shi, Xu and Jia}]{ECSSD}
\bibinfo{author}{Yan, Q.}, \bibinfo{author}{Shi, J.}, \bibinfo{author}{Xu, L.},
  \bibinfo{author}{Jia, J.}, \bibinfo{year}{2014}.
\newblock \bibinfo{title}{Hierarchical saliency detection on extended cssd}.
\newblock \bibinfo{journal}{IEEE Transactions on Pattern Analysis and Machine
  Intelligence} \bibinfo{volume}{38}.
\newblock \DOIprefix\doi{10.1109/TPAMI.2015.2465960}.
\bibitem[{Yang et~al.(2013)Yang, Zhang, Lu, Ruan and Yang}]{DUT-OMRON}
\bibinfo{author}{Yang, C.}, \bibinfo{author}{Zhang, L.}, \bibinfo{author}{Lu,
  H.}, \bibinfo{author}{Ruan, X.}, \bibinfo{author}{Yang, M.H.},
  \bibinfo{year}{2013}.
\newblock \bibinfo{title}{Saliency detection via graph-based manifold ranking}.
\newblock \bibinfo{journal}{Proceedings / CVPR, IEEE Computer Society
  Conference on Computer Vision and Pattern Recognition. IEEE Computer Society
  Conference on Computer Vision and Pattern Recognition} ,
  \bibinfo{pages}{3166--3173}\DOIprefix\doi{10.1109/CVPR.2013.407}.
\bibitem[{Yao et~al.(2016)Yao, Burns, Forsberg, Seitel, Rasoulian, Abolmaesumi,
  Hammernik, Urschler, Ibragimov, Korez, Vrtovec, Castro-Mateos, Pozo, Frangi,
  Summers and Li}]{Yao2016}
\bibinfo{author}{Yao, J.}, \bibinfo{author}{Burns, J.E.},
  \bibinfo{author}{Forsberg, D.}, \bibinfo{author}{Seitel, A.},
  \bibinfo{author}{Rasoulian, A.}, \bibinfo{author}{Abolmaesumi, P.},
  \bibinfo{author}{Hammernik, K.}, \bibinfo{author}{Urschler, M.},
  \bibinfo{author}{Ibragimov, B.}, \bibinfo{author}{Korez, R.},
  \bibinfo{author}{Vrtovec, T.}, \bibinfo{author}{Castro-Mateos, I.},
  \bibinfo{author}{Pozo, J.M.}, \bibinfo{author}{Frangi, A.F.},
  \bibinfo{author}{Summers, R.M.}, \bibinfo{author}{Li, S.},
  \bibinfo{year}{2016}.
\newblock \bibinfo{title}{{A multi-center milestone study of clinical vertebral
  CT segmentation}}.
\newblock \bibinfo{journal}{Computerized Medical Imaging and Graphics}
  \bibinfo{volume}{49}, \bibinfo{pages}{16--28}.
\newblock \DOIprefix\doi{10.1016/j.compmedimag.2015.12.006}.
\bibitem[{Yi et~al.(2014)Yi, Lei, Liao and Li}]{Yi2014}
\bibinfo{author}{Yi, D.}, \bibinfo{author}{Lei, Z.}, \bibinfo{author}{Liao,
  S.}, \bibinfo{author}{Li, S.Z.}, \bibinfo{year}{2014}.
\newblock \bibinfo{title}{{Learning Face Representation from Scratch}}
  \URLprefix \url{http://arxiv.org/abs/1411.7923}.
\bibitem[{Yu et~al.(2018)Yu, Xian, Chen, Liu, Liao, Madhavan and
  Darrell}]{Yu2018}
\bibinfo{author}{Yu, F.}, \bibinfo{author}{Xian, W.}, \bibinfo{author}{Chen,
  Y.}, \bibinfo{author}{Liu, F.}, \bibinfo{author}{Liao, M.},
  \bibinfo{author}{Madhavan, V.}, \bibinfo{author}{Darrell, T.},
  \bibinfo{year}{2018}.
\newblock \bibinfo{title}{{BDD100K: A Diverse Driving Video Database with
  Scalable Annotation Tooling}} , \bibinfo{pages}{1--16}.
\bibitem[{Zhai et~al.(2021)Zhai, Li, Yang, Chen, Cheng and Fan}]{ZhaiCam}
\bibinfo{author}{Zhai, Q.}, \bibinfo{author}{Li, X.}, \bibinfo{author}{Yang,
  F.}, \bibinfo{author}{Chen, C.}, \bibinfo{author}{Cheng, H.},
  \bibinfo{author}{Fan, D.P.}, \bibinfo{year}{2021}.
\newblock \bibinfo{title}{Mutual graph learning for camouflaged object
  detection} .
\bibitem[{Zhan et~al.(2019)Zhan, Sun, Wang, Shi, Clausse, Naumann, Kummerle,
  Konigshof, Stiller, {de La Fortelle} and Tomizuka}]{Zhan2019}
\bibinfo{author}{Zhan, W.}, \bibinfo{author}{Sun, L.}, \bibinfo{author}{Wang,
  D.}, \bibinfo{author}{Shi, H.}, \bibinfo{author}{Clausse, A.},
  \bibinfo{author}{Naumann, M.}, \bibinfo{author}{Kummerle, J.},
  \bibinfo{author}{Konigshof, H.}, \bibinfo{author}{Stiller, C.},
  \bibinfo{author}{{de La Fortelle}, A.}, \bibinfo{author}{Tomizuka, M.},
  \bibinfo{year}{2019}.
\newblock \bibinfo{title}{{INTERACTION Dataset: An INTERnational, Adversarial
  and Cooperative moTION Dataset in Interactive Driving Scenarios with Semantic
  Maps}} .
\bibitem[{Zhang et~al.(2015)Zhang, Ma, Sameki, Sclaroff, Betke, Lin, Shen,
  Price and Mech}]{SOS}
\bibinfo{author}{Zhang, J.}, \bibinfo{author}{Ma, S.}, \bibinfo{author}{Sameki,
  M.}, \bibinfo{author}{Sclaroff, S.}, \bibinfo{author}{Betke, M.},
  \bibinfo{author}{Lin, Z.}, \bibinfo{author}{Shen, X.},
  \bibinfo{author}{Price, B.}, \bibinfo{author}{Mech, R.},
  \bibinfo{year}{2015}.
\newblock \bibinfo{title}{Salient object subitizing} ,
  \bibinfo{pages}{4045--4054}\DOIprefix\doi{10.1109/CVPR.2015.7299031}.
\bibitem[{Zhang et~al.(2019)Zhang, Zhang, Lin, Lu and He}]{CapSal}
\bibinfo{author}{Zhang, L.}, \bibinfo{author}{Zhang, J.}, \bibinfo{author}{Lin,
  Z.}, \bibinfo{author}{Lu, H.}, \bibinfo{author}{He, Y.},
  \bibinfo{year}{2019}.
\newblock \bibinfo{title}{Capsal: Leveraging captioning to boost semantics for
  salient object detection} ,
  \bibinfo{pages}{6017--6026}\DOIprefix\doi{10.1109/CVPR.2019.00618}.
\bibitem[{Zhang et~al.(2017)Zhang, Benenson and Schiele}]{Zhang2017}
\bibinfo{author}{Zhang, S.}, \bibinfo{author}{Benenson, R.},
  \bibinfo{author}{Schiele, B.}, \bibinfo{year}{2017}.
\newblock \bibinfo{title}{{CityPersons: A Diverse Dataset for Pedestrian
  Detection}}.
\newblock \bibinfo{journal}{Proceedings - 30th IEEE Conference on Computer
  Vision and Pattern Recognition, CVPR 2017} \bibinfo{volume}{2017-January},
  \bibinfo{pages}{4457--4465}.
\newblock \URLprefix \url{http://arxiv.org/abs/1702.05693}.
\bibitem[{Zhao et~al.(2019)Zhao, Zheng, Xu and Wu}]{ZhaoObject}
\bibinfo{author}{Zhao, Z.Q.}, \bibinfo{author}{Zheng, P.}, \bibinfo{author}{Xu,
  S.T.}, \bibinfo{author}{Wu, X.}, \bibinfo{year}{2019}.
\newblock \bibinfo{title}{Object detection with deep learning: A review}.
\newblock \bibinfo{journal}{IEEE Transactions on Neural Networks and Learning
  Systems} \bibinfo{volume}{30}, \bibinfo{pages}{3212--3232}.
\newblock \DOIprefix\doi{10.1109/TNNLS.2018.2876865}.
\bibitem[{Zheng et~al.(2018)Zheng, {Hadi Kiapour}, Yang and
  Piramuthu}]{Zheng2018}
\bibinfo{author}{Zheng, S.}, \bibinfo{author}{{Hadi Kiapour}, M.},
  \bibinfo{author}{Yang, F.}, \bibinfo{author}{Piramuthu, R.},
  \bibinfo{year}{2018}.
\newblock \bibinfo{title}{{ModaNet: A large-scale street fashion dataset with
  polygon annotations}}.
\newblock \bibinfo{journal}{MM 2018 - Proceedings of the 2018 ACM Multimedia
  Conference} ,
  \bibinfo{pages}{1670--1678}\DOIprefix\doi{10.1145/3240508.3240652}.
\bibitem[{Zhou et~al.(2017a)Zhou, Lapedriza, Torralba and Oliva}]{Zhou2017a}
\bibinfo{author}{Zhou, B.}, \bibinfo{author}{Lapedriza, A.},
  \bibinfo{author}{Torralba, A.}, \bibinfo{author}{Oliva, A.},
  \bibinfo{year}{2017}a.
\newblock \bibinfo{title}{{Places: An Image Database for Deep Scene
  Understanding}}.
\newblock \bibinfo{journal}{Journal of Vision} \bibinfo{volume}{17},
  \bibinfo{pages}{296}.
\newblock \DOIprefix\doi{10.1167/17.10.296}.
\bibitem[{Zhou et~al.(2017b)Zhou, Zhao, Puig, Fidler, Barriuso and
  Torralba}]{Zhou2017}
\bibinfo{author}{Zhou, B.}, \bibinfo{author}{Zhao, H.}, \bibinfo{author}{Puig,
  X.}, \bibinfo{author}{Fidler, S.}, \bibinfo{author}{Barriuso, A.},
  \bibinfo{author}{Torralba, A.}, \bibinfo{year}{2017}b.
\newblock \bibinfo{title}{{Scene parsing through ADE20K dataset}}.
\newblock \bibinfo{journal}{Proceedings - 30th IEEE Conference on Computer
  Vision and Pattern Recognition, CVPR 2017} \bibinfo{volume}{2017-Janua},
  \bibinfo{pages}{5122--5130}.
\newblock \DOIprefix\doi{10.1109/CVPR.2017.544}.
\bibitem[{Zhou et~al.(2015)Zhou, Cao and Yin}]{Zhou2015}
\bibinfo{author}{Zhou, E.}, \bibinfo{author}{Cao, Z.}, \bibinfo{author}{Yin,
  Q.}, \bibinfo{year}{2015}.
\newblock \bibinfo{title}{{Naive-Deep Face Recognition: Touching the Limit of
  LFW Benchmark or Not?}} \URLprefix \url{http://arxiv.org/abs/1501.04690}.
\bibitem[{Zhou and Yin()}]{zhou}
\bibinfo{author}{Zhou, E.}, \bibinfo{author}{Yin, Q.}, .
\newblock \bibinfo{title}{{Naive-Deep Face Recognition: Touching the Limit of
  LFW Benchmark or Not?}}
\newblock \bibinfo{type}{Technical Report}.
\bibitem[{Zhu et~al.(2015)Zhu, Chen, Dai, Fu, Ye and Jiao}]{Zhu2015}
\bibinfo{author}{Zhu, H.}, \bibinfo{author}{Chen, X.}, \bibinfo{author}{Dai,
  W.}, \bibinfo{author}{Fu, K.}, \bibinfo{author}{Ye, Q.},
  \bibinfo{author}{Jiao, J.}, \bibinfo{year}{2015}.
\newblock \bibinfo{title}{{Orientation robust object detection in aerial images
  using deep convolutional neural network}}, in:
  \bibinfo{booktitle}{Proceedings - International Conference on Image
  Processing, ICIP}, \bibinfo{publisher}{IEEE Computer Society}. pp.
  \bibinfo{pages}{3735--3739}.
\newblock \DOIprefix\doi{10.1109/ICIP.2015.7351502}.
\bibitem[{Zou et~al.(2019a)Zou, Kong, Wong, Wang, Liu and Cao}]{Zou2019}
\bibinfo{author}{Zou, X.}, \bibinfo{author}{Kong, X.}, \bibinfo{author}{Wong,
  W.}, \bibinfo{author}{Wang, C.}, \bibinfo{author}{Liu, Y.},
  \bibinfo{author}{Cao, Y.}, \bibinfo{year}{2019}a.
\newblock \bibinfo{title}{{FashionAI: A Hierarchical Dataset for Fashion
  Understanding}}.
\newblock \bibinfo{journal}{Proceedings of the IEEE Conference on Computer
  Vision and Pattern Recognition Workshops} .
\bibitem[{Zou and Shi(2018)}]{Zou2018}
\bibinfo{author}{Zou, Z.}, \bibinfo{author}{Shi, Z.}, \bibinfo{year}{2018}.
\newblock \bibinfo{title}{{Random access memories: A new paradigm for target
  detection in high resolution aerial remote sensing images}}.
\newblock \bibinfo{journal}{IEEE Transactions on Image Processing}
  \bibinfo{volume}{27}, \bibinfo{pages}{1100--1111}.
\newblock \DOIprefix\doi{10.1109/TIP.2017.2773199}.
\bibitem[{Zou et~al.(2019b)Zou, Shi, Guo and Ye}]{Zou2019a}
\bibinfo{author}{Zou, Z.}, \bibinfo{author}{Shi, Z.}, \bibinfo{author}{Guo,
  Y.}, \bibinfo{author}{Ye, J.}, \bibinfo{year}{2019}b.
\newblock \bibinfo{title}{{Object Detection in 20 Years: A Survey}} ,
  \bibinfo{pages}{1--39}\URLprefix \url{http://arxiv.org/abs/1905.05055}.

\end{thebibliography}

\end{document}